\documentclass{article}

\usepackage{microtype}
\usepackage{graphicx}
\usepackage{subfigure}
\usepackage{booktabs} %

\usepackage{hyperref}

\PassOptionsToPackage{numbers}{natbib}

\usepackage[accepted]{icml2025}

\usepackage{amsmath}
\usepackage{amssymb}
\usepackage{mathtools}
\usepackage{amsthm}

\usepackage[capitalize,noabbrev]{cleveref}

\theoremstyle{plain}

\theoremstyle{definition}

\theoremstyle{remark}

\usepackage[textsize=tiny]{todonotes}

\usepackage[T1]{fontenc}
\usepackage{url}
\usepackage{multirow}
\usepackage{caption}
\usepackage{float}
\usepackage{array}
\usepackage{comment}
\usepackage{xcolor}

\usepackage{amsmath,amsfonts,bm}

\def\eqref#1{equation~\ref{#1}}

\def\1{\bm{1}}

\DeclareMathAlphabet{\mathsfit}{\encodingdefault}{\sfdefault}{m}{sl}
\SetMathAlphabet{\mathsfit}{bold}{\encodingdefault}{\sfdefault}{bx}{n}

\newcommand{\archon}{\textsc{Archon}}

\icmltitlerunning{An Architecture Search Framework for Inference-Time Techniques}

\begin{document}

\twocolumn[
\icmltitle{An Architecture Search Framework for Inference-Time Techniques}

\begin{icmlauthorlist}
\icmlauthor{Jon Saad-Falcon}{stanford}
\icmlauthor{Adrian Gamarra Lafuente}{stanford}
\icmlauthor{Shlok Natarajan}{stanford}
\icmlauthor{Nahum Maru}{stanford}
\icmlauthor{Hristo Todorov}{stanford}
\icmlauthor{Etash Guha}{uw}
\icmlauthor{E. Kelly Buchanan}{stanford}
\icmlauthor{Mayee Chen}{stanford}
\icmlauthor{Neel Guha}{stanford}
\icmlauthor{Christopher Ré}{stanford}
\icmlauthor{Azalia Mirhoseini}{stanford}
\end{icmlauthorlist}

\icmlaffiliation{stanford}{Stanford University, Stanford, CA, USA}
\icmlaffiliation{uw}{University of Washington, Seattle, WA, USA}

\icmlcorrespondingauthor{Jon Saad-Falcon}{jonsaadfalcon@stanford.edu}

\vskip 0.3in
]

\printAffiliationsAndNotice{}  %

\begin{abstract}

Inference-time techniques, such as repeated sampling or iterative revisions, are emerging as powerful ways to enhance large-language models (LLMs) at test time. 
However, best practices for developing systems that combine these techniques remain underdeveloped due to our limited understanding of the utility of each technique across models and tasks, the interactions between them, and the massive search space for combining them. 
To address these challenges, we introduce \archon{}, a modular and automated framework for optimizing the process of selecting and combining inference-time techniques and LLMs. 
Given a compute budget and a set of available LLMs, \archon{} explores a large design space to discover optimized configurations tailored to target benchmarks. 
It can design custom or general-purpose architectures that advance the Pareto frontier of accuracy vs. maximum token budget compared to top-performing baselines. 
Across instruction-following, reasoning, and coding tasks, we show that \archon{} can leverage additional inference compute budget to design systems that outperform frontier models such as OpenAI's o1, GPT-4o, and Claude 3.5 Sonnet by an average of 15.1$\%$. 
\end{abstract}

\section{Introduction}

\begin{figure}[!htbp]
   \centering
   \includegraphics[width=0.95\linewidth]{figures/Archon_First_Page_Results_v2.pdf}
   \caption{
   \textbf{\archon{}'s Performance Effectively Scales with Increasing Inference Budget.} Individual dataset analysis included in \autoref{fig:Archon_Budget_Analysis}.
   }
   \label{fig:Archon_First_Page_Results}
\end{figure}

Inference-time techniques---strategies that use additional compute during model inference---are gaining traction as effective methods for improving model capabilities.
LLMs, such as OpenAI's o1 \citep{openai2024o1}, QwQ \citep{qwq-32b-preview}, and Sky-T1 \citep{sky_t1_2025}, utilize such techniques to translate additional inference compute into better performance across a broad set of tasks.
Example techniques include generation ensembling, ranking, and fusion, where models in the ensemble are queried in parallel, their responses are ranked, and the best ones are fused into a single, higher quality output, respectively \citep{llm-blender-2023,wang2024mixtureofagentsenhanceslargelanguage}.
Other types of inference-time techniques are based on querying a single LLM successively (via repeated sampling) and using a voting strategy or unit tests to select the top generation \citep{brown2024largelanguagemonkeysscaling, chen2024llmcallsneedscaling, li2024agentsneed}.

Recent work has made progress towards building robust \textit{inference-time architectures}: systems composed of one or more large language models (LLMs) leveraging inference-time techniques. 
Examples include Mixture-of-Agents (MoA)  \citep{wang2024mixtureofagentsenhanceslargelanguage} and LLM-Blender \citep{llm-blender-2023}, as well as single-model systems like ADAS \citep{hu2024ADAS} and AFlow \citep{zhang2024aflowautomatingagenticworkflow}.
However, our experiments show that these top-performing baselines have limitations in compute utilization and task generalization. (see Section \ref{sec:archon_results}). 
We argue that designing effective and generalizable inference-time architectures requires the following:

\begin{figure*}[t]
   \centering
   \includegraphics[width=0.9\linewidth]{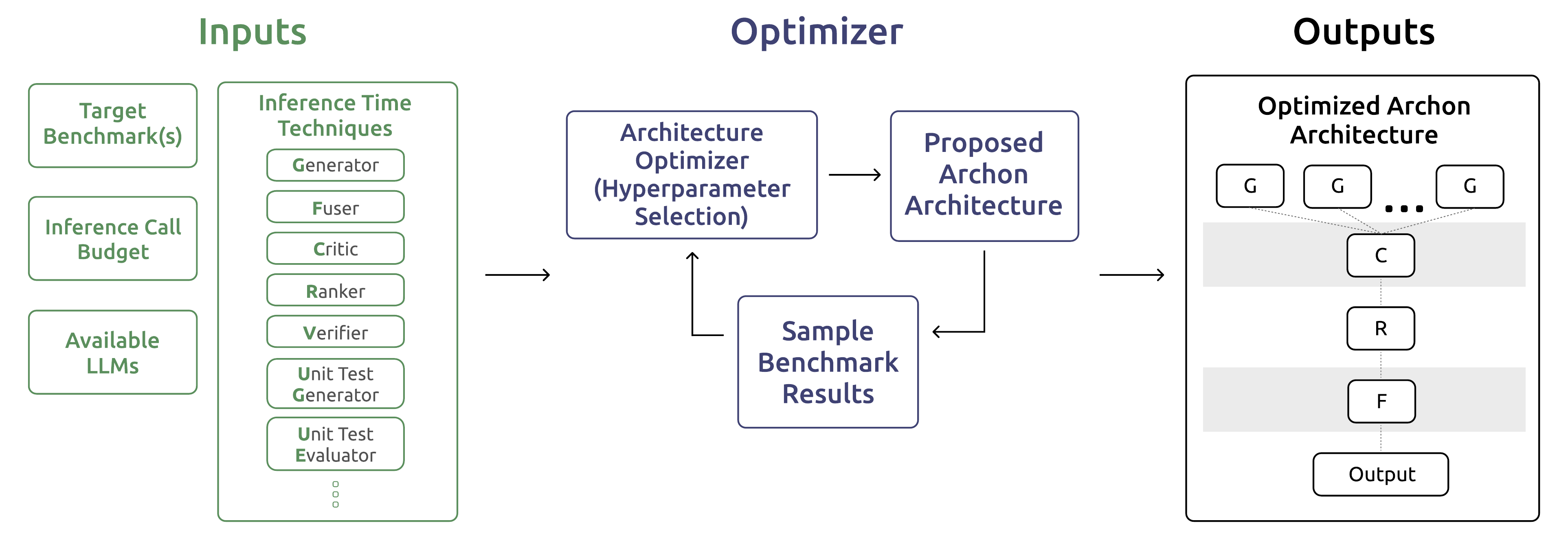}
   \caption{
   \textbf{Overview of \archon{} Framework}: \archon{}'s search algorithm requires the following inputs: target benchmarks, inference call budget, available LLMs, and available inference-time techniques (\textbf{left}).
   The search algorithm uses Bayesian optimization \citep{snoek2012practicalbayesianoptimizationmachine} to construct and evaluate different \archon{} configurations (\textbf{middle}) before returning the optimized \archon{} architecture (\textbf{right}) for the target benchmarks (Section \ref{sec:search_algorithms}).
   }
   \label{fig:main_figure}
\end{figure*}

\begin{itemize}
\item \textbf{Understanding the Utilities of Inference-Time Techniques}:
Inference-time architectures typically delegate their additional inference budget towards more model sampling calls \citep{chen2024llmcallsneedscaling, brown2024largelanguagemonkeysscaling}, which can be effective for math and coding tasks.
Other tasks, such as following instructions and reasoning, have been shown to benefit from additional techniques, including ranking and fusion \citep{wang2024mixtureofagentsenhanceslargelanguage, llm-blender-2023}.
While all of these methods are valuable, \textit{it is essential to identify which inference-time techniques are most effective for different task categories.}
\item \textbf{Understanding the Interactions Between Inference-Time Techniques}: 
While previous studies analyzed these techniques individually (e.g., generation sampling in \citet{chen2024llmcallsneedscaling}), \textit{we need a more comprehensive understanding of the relationships between different inference-time techniques} across different tasks (e.g., is it better to use more models or generate more samples per model?).

\item \textbf{Efficiently and Automatically Searching the Large Design Space of Inference-Time Architectures}: Given a set of available LLMs and target tasks, there is currently no single prevailing inference-time architecture for maximizing downstream accuracy across all tasks (\autoref{tab:archon_main_table_results}). The search space for inference-time architectures is expansive, requiring practitioners to make several key configuration decisions, such as \textit{which LLMs to use, how many times to sample them, and how to combine and filter the candidate generations}. These motivate the need for automated and adaptive architecture search approaches.  

\end{itemize}

In our work, we address each of these challenges.
First, we \textbf{evaluate the utilities of
a comprehensive set of existing and proposed inference-time techniques} across instruction-following, reasoning, and coding tasks. 
Using both open-source and closed-source models, we examine a range of techniques such as \textit{ensembling, fusion, ranking, critiquing, verification, and model-based unit test generation/evaluation} (Sections \ref{sec:compound_lm_modules} and \ref{sec:combining_compound_lm_components}).
We find that no single technique completely dominates across all tasks, with different approaches being more effective for different tasks.

Second, we \textbf{analyze the interactions between inference-time techniques} and explore the benefits of adding new models and new techniques individually.
We find that generation ensembling combined with critique, verification, and fusion improves the final response quality beyond the oracle best candidate from individual (non-fused) responses, particularly for instruction-following and reasoning tasks (\autoref{fig:compound_LM_ablations}; \autoref{fig:ensembling_graphs}; \autoref{tab:archon_component_ablations}).
We also demonstrate increased performance as we scale up the layers of inference-time techniques and combine multiple approaches together, allowing us to discover effective new combinations of inference-time techniques (Sections \ref{sec:combining_compound_lm_components}, \ref{sec:archon_results}, \ref{sec:tradeoffs}).
Combining multiple strategies significantly improves task performance, but determining the specific combination remains challenging. This requires manually testing models, inference-time techniques, architecture designs, inference budgets, and more.

Third, drawing upon our analysis of inference-time techniques, we present \textbf{\archon{}}, an open-source modular framework for automatically designing LLM systems composed of existing inference-time techniques (or new ones), allowing practitioners to optimize for their desired objective functions: accuracy, latency, and cost (Sections \ref{sec:compound_lm_modules}, \ref{sec:search_algorithms}).
Unlike alternative LM systems that perform prompt engineering and tool use over a single LM \citep{khattab2023dspy, yuksekgonul2024textgrad, hu2024ADAS, zhang2024aflowautomatingagenticworkflow}, our approach integrates multiple LMs in a single architecture and reduces prompt selection to a set of core components.
The \archon{} framework utilizes automatic architecture search algorithms to maximize generation quality for the given tasks(s), leveraging Bayesian optimization \citep{snoek2012practicalbayesianoptimizationmachine, 9124618} techniques inspired by (NAS) \citep{zoph2017neuralarchitecturesearchreinforcement, ren2021comprehensive} to rapidly traverse the space of potential inference architectures (Section \ref{sec:search_algorithms}).

We evaluate \archon{} architectures across a diverse set of instruction-following, reasoning, and coding benchmarks (\autoref{tab:archon_main_table_results}): MT-Bench, Arena-Hard-Auto, Alpaca-2.0 Eval, MixEval, MATH, and CodeContests \citep{zheng2023judging, arenahard2024, alpaca_eval, ni2024mixevalderivingwisdomcrowd, hendrycks2021measuring, li2022competition}.
Our best \archon{} architectures surpass both frontier models 
(e.g. OpenAI's O1, GPT-4o and Claude-3.5 Sonnet)
and prior top-performing inference-time architectures 
(e.g. ADAS, AFlow, and MoA), 
\textit{boosting state-of-the-art (SOTA) performance by 15.1$\%$}, on average.
Furthermore, \archon{} achieves SOTA performances while using \textit{20.0\% less inference calls, 15.1\% less input tokens, and 13.5\% less output tokens} than alternative inference-time architectures (\autoref{fig:Archon_First_Page_Results}; \autoref{tab:archon_main_table_results}; \autoref{fig:Archon_Budget_Analysis}).
Even when solely using open-source LLMs, \archon{} architectures, on average, surpass SOTA LLMs by 11.2$\%$.

Overall, we present \archon{} as an open-source inference-time framework, readily extensible to new inference-time techniques, models, and tasks via user-friendly interfaces.

\vspace{-2mm}
\section{Related Work}
\vspace{-2mm}

Despite advancements in inference-time architectures,
many architectures focus on additional generations \citep{llm-blender-2023, chen2024llmcallsneedscaling, davis2024networksnetworkscomplexityclass}, which is effective for reasoning tasks \citep{brown2024largelanguagemonkeysscaling}.
However, for tasks like instruction-following and reasoning, techniques such as fusion and ranking are effective for bolstering task performances \citep{wang2024mixtureofagentsenhanceslargelanguage,llm-blender-2023}.
Prior studies have explored limited aspects of configurations, often focusing on specific benchmarks \citep{llm-blender-2023, wang2024mixtureofagentsenhanceslargelanguage, chen2024llmcallsneedscaling, li2024agentsneed}.
It's crucial to efficiently develop inference-time architectures, as optimal configurations vary based on benchmarks, available models, and inference compute limits (Section \ref{sec:archon_results}).
Furthermore, LM orchestration frameworks, such as DSPy \citep{khattab2023dspy}, only optimize a single prompt for a single LM, better equipping it for tool use by utilizing supervised data but still unable to leverage multiple inference-time techniques in parallel or sequentially.
While each of these approaches manually selects a subset of existing techniques, \archon{} unifies available inference-time techniques and automates architecture construction with search algorithms, simplifying the model and component selection process for each set of tasks (Sections \ref{sec:compound_lm_modules} and \ref{sec:search_algorithms}).

\vspace{-2mm}
\section{Inference-Time Techniques for \archon{}}
\vspace{-1mm}

With the proliferation of inference-time techniques, \archon{} introduces a systematic framework for understanding and unifying these methods into inference-time architectures.
Below, we elaborate on the structure, inputs, and outputs of each of the inference-time techniques (\autoref{tab:archon_layers_outputs_figure}).
Then, we discuss how to combine the different techniques into an inference-time architecture (Section \ref{sec:combining_compound_lm_components}) 
before finally exploring automatic approaches for constructing inference-time architectures (Section \ref{sec:search_algorithms}).

\vspace{-2mm}
\subsection{LLM Components of \archon}
\label{sec:compound_lm_modules}
\vspace{-1mm}

In this section, we discuss the \textit{LLM components} of \archon{}, which are LLMs that perform a specific inference-time technique. 
{We test an array of different components inspired by recent work, incorporating approaches for generating, ranking, and fusing candidates \citep{wang2024mixtureofagentsenhanceslargelanguage, llm-blender-2023} as well as approaches for improving candidate response quality through critiquing, verifying, and unit testing \citep{bai2022constitutionalaiharmlessnessai, zheng2023judging}.}
The components and their prompts are summarized in \autoref{tab:archon_layers_outputs_figure} and Appendix \ref{sec:lm_construction_and_prompts}.
{We also perform an extensive ablation study of the given \archon{} components across instruction-following, reasoning, and coding benchmarks to better understand their individual utilities and their optimal combinations for different tasks (Appendix \ref{sec:tradeoffs}).}

\noindent \textbf{Generator} is an LLM that takes in the instruction prompt and outputs candidate responses.  
Generators can be called in parallel to perform \textit{generation ensembling} (i.e. calling multiple LLMs in parallel) \citep{wang2024mixtureofagentsenhanceslargelanguage}, or sampled multiple times \citep{brown2024largelanguagemonkeysscaling}.
The number of models, samples, and generation temperature can be adjusted.

\noindent {We find additional model sampling to significantly boost performance (\autoref{fig:sampling_graphs}), particularly for coding tasks (\autoref{tab:archon_main_table_results}).
We see a similar pattern for model ensembling, where sampling from additional models leads to continual performance increases (assuming the models are ordered from best to worst for the given task) (\autoref{fig:ensembling_graphs}).}

\noindent \textbf{Fuser} is an LLM that, given an instruction prompt and a set of proposed responses as input, combines these responses to generate one or more higher-quality fused responses.

\noindent {For every benchmark explored, we found that the Fuser module substantially improved performance (8.9\% on average) (\autoref{fig:sampling_graphs}; \autoref{fig:ensembling_graphs}; \autoref{fig:compound_LM_ablations}).
Additionally, we observed similar benefits in the \archon{} framework when adding multiple layers of Fusers (\autoref{fig:compound_LM_ablations}).
The number of Fuser layers needed to improve performance varied by task (\autoref{fig:fusion_layer_analysis}), with some tasks receiving limited benefits from added layers (1-2 point increase in accuracy for MixEval) while others experienced significant benefits with 3-4 fusion layers and more (10 to 15 point increase in win rate for MT Bench and Alpaca Eval 2.0).
}

\noindent \textbf{Ranker} is an LLM that, given an instruction prompt and a set of proposed responses as input, ranks the candidate generations based on their quality, producing a ranked list of responses as output. This ranking is then used to filter the set of responses to the top-$K$, as specified.

\noindent {From our results in \autoref{tab:archon_component_ablations}, \autoref{fig:sampling_graphs}, and \autoref{fig:ensembling_graphs}, our results show the Ranker was most effective for instruction-following and reasoning tasks by using pair-wise comparisons that focus on style and prompt adherence. 
We found that on MT Bench and Arena-Hard-Auto benchmarks, the Ranker improved output quality by 10.8\% over random selection while performing within 2.7\% of oracle selection.}

\noindent \textbf{Critic} is an LLM that, given an instruction prompt and a set of proposed responses as input, produces a list of strengths and weaknesses for each response, which is then used to improve the quality of the final response (Section \ref{sec:combining_compound_lm_components}; \autoref{fig:compound_LM_ablations}).

\noindent {The Critic module proved effective for every task we explored in \autoref{fig:compound_LM_ablations} and \autoref{tab:archon_component_ablations}.
With our 10-model 70B+ Generator ensemble and Fuser configuration of \archon{}, the added Critic improved performance on average by 11.5 percentage points across the benchmarks explored.}

\noindent \textbf{Verifier} is an LLM that verifies whether a provided candidate response has appropriate reasoning for a given instruction prompt.
It proceeds in two stages: \textbf{Stage \#1} takes in the instruction prompt and a candidate response as input and outputs reasoning for why the candidate response is correct; \textbf{Stage \#2} takes in the instruction prompt, candidate response, and produced reasoning before outputting reasoning and a verdict (i.e., binary [Correct] or [Incorrect]) for whether or not the candidate response is correct according to the provided instruction prompt and reasoning.
Only verified responses are passed to the next \archon{} layer.

\noindent {The Verifier was most effective for the reasoning benchmarks explored in \autoref{tab:archon_component_ablations}, improving performance by 8.4\% for MixEval, MixEval Hard, and MATH.
When just using a 70B+ Generator ensemble with Verifier module after generation, the \archon{} configuration lagged behind the \archon{} ensemble and fuser configuration by 1.5\%, on average, across all benchmarks explored, suggesting verification is most effective when combined with other inference-time techniques.}

\noindent \textbf{Unit Test Generator} and \textbf{Unit Test Evaluator} are complementary LLM components in our system: the Unit Test Generator takes an instruction prompt and produces 5-10 concise test statements (Section \ref{sec:archon_results}; examples in \autoref{tab:unit_test_examples}) for assessing response accuracy and relevance, while the Evaluator takes the instruction prompt, candidate response(s), and these tests as input to rank responses by test passage. The Evaluator justifies and aggregates test verdicts across candidates, scoring each response for reasoning and coding tasks, and only responses passing all tests proceed to the next \archon{} layer. This approach extends evaluation beyond coding to various task types through configurable test quantities.

\noindent {The Unit Test Generator and Evaluator were most effective on reasoning and coding tasks, improving performance on benchmarks that required more verification steps (7.4\% boost) (\autoref{tab:archon_component_ablations}). 
When the 70B+ ensemble of Generators was only combined with unit tests, it was less effective for reasoning tasks like Arena-Hard-Auto and MixEval, lagging behind the ensemble and fuser configuration by 3.1\%. 
However, when we increased generation sampling and added unit test generation/evaluation for CodeContests, we observed a 56\% boost in Pass@1 performance (\autoref{tab:archon_main_table_results}), increasing from 17.9 to 29.3\% Pass@1.
}

\subsection{Combining the LLM Components}
\label{sec:combining_compound_lm_components}

\noindent \textbf{Performance Gains from Scaling Inference-Time Techniques}: 
We explore the utilities of individual \archon{} components and evaluate whether combinations of inference-time techniques enable us to \textit{build LM systems greater than the sum of their parts}.
{For our analysis, we look at seven datasets spanning instruction-following, reasoning, mathematics, and coding: MT-Bench \citep{zheng2023judging}, AlpacaEval 2.0 \citep{alpaca_eval}, Arena Hard Auto \citep{arenahard2024}, MixEval \citep{ni2024mixevalderivingwisdomcrowd}, MixEval-Hard, MATH \citep{hendrycks2021measuring}, and CodeContests \citep{li2022competition}. 
We also test across the current SOTA open-source and closed-source LMs (\autoref{tab:models_overview}; \autoref{tab:generator_and_fusion_rankings}).
For the analysis of each inference-time technique, we focus on \textbf{1)} testing it across different benchmarks, \textbf{2)} scaling its usage individually, \textbf{3)} scaling it while randomly choosing another technique and holding that technique constant, and \textbf{4)} varying its position among different components. 
We include these ablation experiments in Section \ref{sec:tradeoffs}, where we include the \archon{} component combinations in \autoref{tab:archon_component_ablations} and the model type used in the combinations in \autoref{tab:archon_component_ablations_with_gpt4o} and \autoref{tab:archon_component_ablations_with_haiku}.
}

From our analysis, we find several trends (designated with \textbf{T}s) across the combinations of inference-time architectures:

\begin{itemize}

    \item \textbf{T1}: Repeated model sampling and additional ensemble models leads to substantial gains, leading to 9.3\% and 18.5\% increases, respectively (\autoref{fig:combined_sampling_and_ensembling_graphs}; \autoref{fig:ensembling_graphs}).

    \item \textbf{T2}: Scaling the layers of inference-time techniques significantly improves performance across instruction-following, reasoning, and coding tasks, such as always adding a single fuser as the last layer (\autoref{fig:compound_LM_ablations}). 

    \item \textbf{T3}: Scaling the diversity of inference-time techniques included also bolsters task performance across the explored tasks, with critics and rankers before fusers being particularly effective (\autoref{fig:compound_LM_ablations}; \autoref{fig:combined_sampling_and_ensembling_graphs}).

    \item \textbf{T4}:  In reasoning tasks, incorporating the Verifier and Unit Test Generator/Evaluator modules alongside the Fuser improves performance by filtering out flawed responses, contributing to significant performance gains in tasks like MixEval and CodeContests (\autoref{tab:archon_component_ablations}; Section \ref{sec:archon_architectures}).

\end{itemize}

\noindent \textbf{Framework Overview}: 
{
Drawing upon our analysis of the inference-time components, we propose \textbf{\archon{}}, a framework for automatically designing LLM systems composed of existing inference-time techniques (or new ones).
}
Inspired by the structure of neural networks \citep{hinton1992neural}, \archon{} consists of layers of LLM components (\autoref{fig:main_figure}; Section \ref{sec:compound_lm_modules}).
Each layer is composed of sets of LLM components called in parallel. These components perform a text-to-text operation on the initial instruction prompt and the candidate responses from the previous layer.
Furthermore, like a neural network, some layers perform \textit{transformations} of the provided list of strings (e.g., Generator and Fuser), converting a list of strings into a different list of strings (the numbers of candidates can vary from the original number of candidates).
Other components introduce non-linearities into the \archon{} structure, performing filtering of the list of strings (e.g., Ranker and Verifier).
Ultimately, the inputs and outputs for each layer is always a list of strings, whether that is the instruction prompt (i.e., a single string) or a list of candidate responses.
If a list of strings is outputted at the last layer of the \archon{} structure, the first string in the list is returned.

\begin{figure}[]
   \centering
   \includegraphics[width=0.70\linewidth]{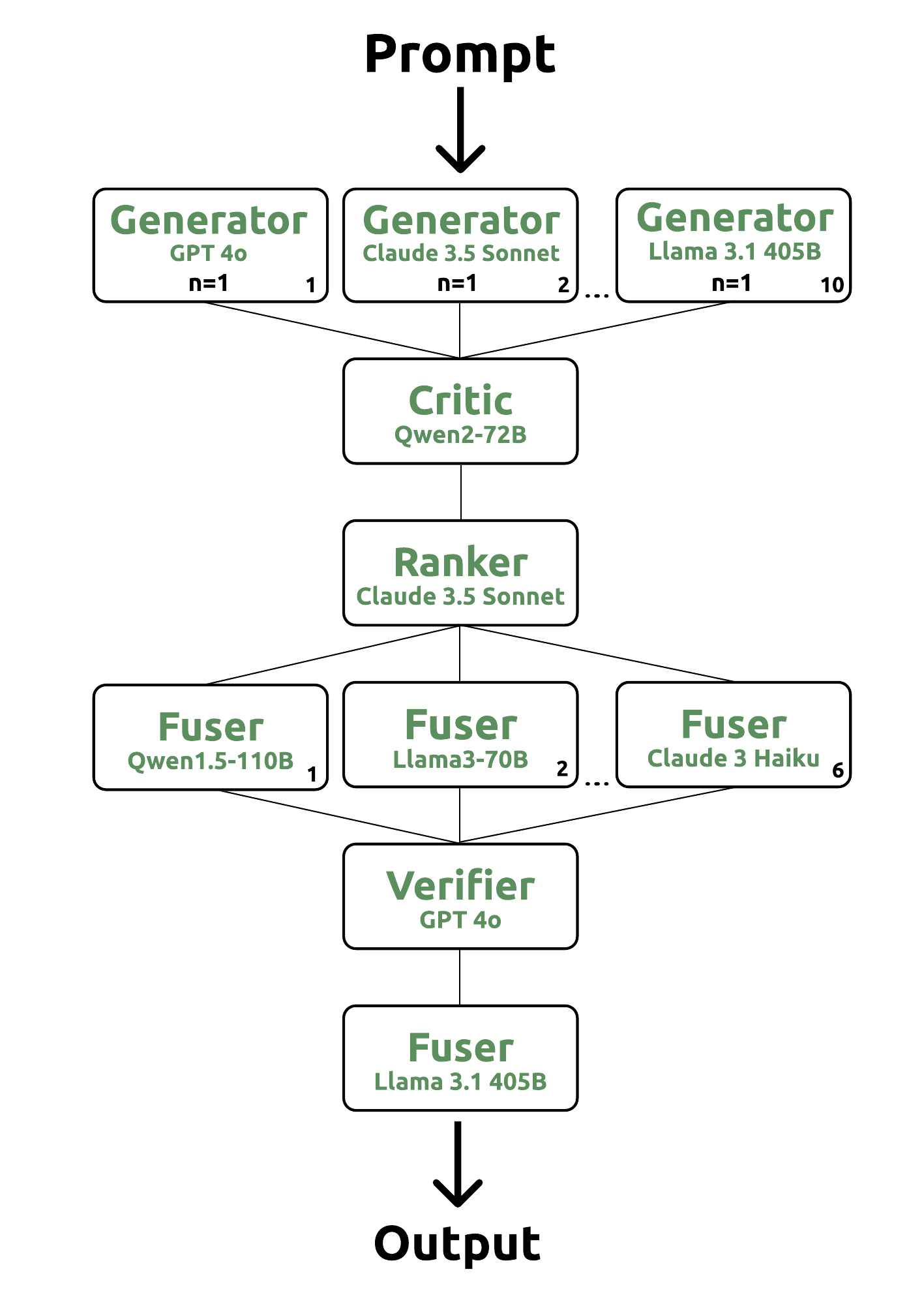}
   \caption{\textbf{Example \archon{} Architecture}: This architecture starts with ten generator models (each sampled once), followed by a critic model, a ranker model, one layer of six fuser models, a verifier model, and finishes with a fuser model.
   }
   \label{fig:example_archon_architecture}
\end{figure}

Unlike a classical neural network, no weights are learned between the LLM components and the layers; in turn, the \archon{} architecture can be deployed off-the-shelf without any tuning.
Additionally, a single state is transformed sequentially from the input layer to the final output; this single state is the initial instruction prompt and the current candidate responses (example architecture in \autoref{fig:example_archon_architecture}).

\noindent \textbf{Rules for Construction}: The LLM components in Section \ref{sec:compound_lm_modules} can only be placed in specific orders (\ref{tab:rules_of_construction}). 
{While alternative combinations and orderings of \archon{} components are technically viable, we found these orderings to be optimal after conducting an ablation study of \archon{} components across seven benchmarks and two model classes (open-source and closed-source) (Appendix \ref{sec:tradeoffs}).}

\begin{enumerate}
\item {Only one type of component is allowed in any given layer.}
\item {Generator components can only be placed in the first layer of \archon{}; you can place one or more Generators.}
\item {The Critic must come before a Ranker or a Fuser. Otherwise, the generated strengths and weaknesses cannot be incorporated into generation ranking or fusion.}
\item {Ranker, Critic, Verifier, and Unit Test Generator/Evaluator layers can go anywhere in \archon{} except the first layer. For each of these components, it must be the only module in its layer.}
\item {Fuser components can go anywhere in \archon{} except the first layer. Multiple Fusers can be used in a layer.}
\item {Unit Test Generators and Evaluators are placed in consecutive layers, with the Unit Test Generator always first.}
\end{enumerate}

\begin{figure*}[]
   \centering
   \includegraphics[width=0.85\linewidth]{figures/Archon_Ablations.pdf}
   \caption{\textbf{Performance Improves by Scaling \textit{Layers} of Inference-Time Techniques}: 
   When controlling for inference budget, generation ensembling and fusion across 8 different 70B LLMs is generally more effective than repeated sampling with only the top performing model.
   Furthermore, adding layers of critique and fusion led to a 18.8\% boost in task performance, on average.
   However, the best inference-time architecture differed by task, such as MixEval and CodeContests (Section \ref{sec:archon_by_task}), which inspired us to develop architecture search techniques for \archon{} (Section \ref{sec:search_algorithms}). 
   }
   \label{fig:compound_LM_ablations}
\end{figure*}

\subsection{Architecture Search Algorithms}
\label{sec:search_algorithms}

\noindent \textbf{Search Hyperparameters}: In this section, we explore how to automatically design inference-time architectures for target tasks via  \archon{}'s architecture search algorithms. 
Guided by the trends found in our analysis in Section \ref{sec:combining_compound_lm_components}, we establish six axes of hyperparameters for the search space:

\begin{enumerate}
    \item \textbf{Top-$K$ Generators for Ensemble}: The top-$K$ models for the initial Generator ensemble, ranging from 1 to 10 (\textbf{T1}). 
    The top-$K$ models are selected greedily based on their individual performances on target task(s)(\autoref{tab:generator_and_fusion_rankings}). 
    \item \textbf{Top-$K$ Generator Samples}: The number of samples gathered from each ensemble generator (same for all the models), ranging from 1 to 5 (\textbf{T1}).
    For CodeContests, we explore high-sample settings: [1, 10, 100, 500, 1000]. 
    \item \textbf{Number of Fusion Layers}: Ranges from 1 to 4. 
    The last fusion layer will always have a single Fuser (\textbf{T2}).
    \item \textbf{Top-$K$ Fusers}: Number of models used for each fusion layer, ranges from 2 to 10 in increments of 2 (\textbf{T2,3}).
    \item \textbf{Critic and Ranker Layers}: We add critic and ranker layers before each fuser layer since we find they provide added benefits across the benchmarks explored (\textbf{T3}) (Section \ref{sec:combining_compound_lm_components}; \autoref{fig:compound_LM_ablations}; \autoref{fig:ensembling_graphs}).
    \item \textbf{Evaluation Layer}: Option to add Verifier, Unit Test Gen./Eval., or neither before the last Fuser layer (\textbf{T4}).
\end{enumerate}

While it is possible to further expand the search space of potential \archon{} architectures (e.g., different temperatures for generative LLM components, alternative prompts for each LLM component, additional LLM components for \archon{}, etc.), 
the trends we identify from Section \ref{sec:combining_compound_lm_components} reasonably constrain the search space of configurations to focus on the most influential hyperparameters.
In total, our search space contains 9,576 configurations, which we obtain by combining all possible hyperparameters and removing invalid configurations (for example, we discard configurations where the number of initial generations exceeds the context window of the fusers).

\noindent \textbf{Search Method}: 
The \archon{} search method takes in four inputs: the target benchmark(s), the inference call budget, the set of available LLMs, and the inference-time techniques for construction (\autoref{fig:main_figure}).
As output, the search method outputs a single optimized \archon{} architecture.
We use 20\% of each target dataset as a development set for guiding architecture search.
We explore three approaches for \archon{}'s architecture search: \textit{random search} (randomly test potential architectures in the search space), \textit{greedy search} (greedily optimize individual hyperparameters one at a time, starting from a random initial architecture), and \textit{Bayesian Optimization} \citep{snoek2012practicalbayesianoptimizationmachine} (global hyperparameter optimization with Gaussian processes).
As inputs, Bayesian optimization takes in a vector specifying the configuration choices for the generators (i.e., number of models and samples), layers of fusers, numbers of fusers per layer, and final verifier / unit tester (Section \ref{sec:combining_compound_lm_components}). 
Bayesian optimization begins by sampling a specified number of random \archon{} architectures to calibrate its surrogate model. The task performance of these sampled architectures is used to guide more informed architecture suggestions during the configuration search.
The algorithm repeats the following cycle---evaluating each suggested architecture and using its performance to refine future suggestions---until it discovers the optimal \archon{} configuration, or until the inference call budget is exhausted.
For more details on our open-source Bayesian optimization approach, please see Appendix \ref{sec:bayesian_optimization}, where we further discuss implementation and how to utilize alternative optimization functions, such as latency.

Bayesian optimization found the best architectures in 96.0\% of searches and required 88.5\% fewer architecture evaluations than greedy search and 90.4\% fewer than random search (\autoref{fig:search_algorithms_comparison}).
The effectiveness of Bayesian optimization increases with the number of initial randomly sampled architectures, up to around 230-240 samples, after which further testing is better focused on configuration search (\autoref{tab:bayesian_optimization_comparisons}).
For limited inference call budgets (<20 calls), Bayesian optimization is less effective, and traditional methods like greedy search may perform comparably (\autoref{tab:search_algorithms_by_inference_call_budget}).

\noindent {
\textbf{Adding Search Restrictions}: 
To impose compute constraints during architecture search, we exclude any \archon{} architecture that would exceed the inference call, input token, or output token budgets from the search space.
Multiple restrictions can be added. 
For example, you can filter out architectures with more than 20 inference calls or more than 20,000 input tokens.
This prevents our Bayesian optimization algorithm from even considering these invalid architectures in our architecture search, allowing us to compute-match \archon{} against alternate inference-time frameworks such as ADAS and AFlow (\autoref{fig:Archon_Budget_Analysis}; \autoref{fig:PFLOPs_vs_Performance}).
}

\vspace{-2mm}
\section{Experiments}
\label{sec:experiments}
\vspace{-2mm}

Our experiments focus on answering the following questions: 
\textbf{(1)} how does \archon{} compare to existing SOTA LLMs and inference-time architectures in terms of accuracy and compute efficiency (Section \ref{sec:archon_results})?
\textbf{(2)} how does \archon{} performance compare across the tasks explored (Section \ref{sec:archon_by_task})?
\textbf{(3)} what are the considerations for model size, latency, and cost surrounding \archon{} (Section \ref{sec:limitations_and_future_work})?
We outline the benchmarks, models, and techniques for constructing \archon{} architectures in Section \ref{sec:benchmarks_and_models}.

\vspace{-2mm}
\subsection{Benchmarks and Models}
\vspace{-1mm}

\label{sec:benchmarks_and_models}

\begin{table*}[t]
   \centering
       \scriptsize
       \setlength{\tabcolsep}{1.0pt}
       \begin{tabular}{ccccccccccccccc}
       \toprule
       & & & & & & & & \textbf{\begin{tabular}[c]{@{}c@{}}MT\\ Bench\end{tabular}} & \textbf{\begin{tabular}[c]{@{}c@{}}Alpaca\\Eval 2.0\end{tabular}}  & \textbf{\begin{tabular}[c]{@{}c@{}}Arena\\Hard Auto\end{tabular}} & \textbf{\begin{tabular}[c]{@{}c@{}}MixEval\\ Hard\end{tabular}} & \textbf{MixEval} & \textbf{MATH$^*$} & \textbf{\begin{tabular}[c]{@{}c@{}}Code\\Contests$^*$\end{tabular}}  \\
       \midrule
       & & \textbf{Approaches} & \begin{tabular}[c]{@{}c@{}} \textbf{Average}\\ \textbf{Infer.}\\ \textbf{Calls}\end{tabular} & \begin{tabular}[c]{@{}c@{}} \textbf{Average}\\ \textbf{Input}\\ \textbf{Tokens}\end{tabular} & \begin{tabular}[c]{@{}c@{}} \textbf{Average}\\ \textbf{Output}\\ \textbf{Tokens}\end{tabular} & {\begin{tabular}[c]{@{}c@{}} \textbf{Avg. PFLOPs}\\ \textbf{per Query}\end{tabular}} & {\begin{tabular}[c]{@{}c@{}} \textbf{Dollars}\\ \textbf{per Query}\end{tabular}} &  W.R. & \begin{tabular}[c]{@{}c@{}}L.C.\\ W.R.\end{tabular} & W.R & Acc. & Acc. & \begin{tabular}[c]{@{}c@{}}Pass\\ @1\end{tabular} & \begin{tabular}[c]{@{}c@{}}Pass\\ @1\end{tabular} \\
       \midrule
       {\multirow{8}{*}{\rotatebox{90}{\begin{tabular}[c]{@{}c@{}}\textbf{Baselines}\\ \end{tabular}}}} & \multirow{3}{*}{\rotatebox{0}{\begin{tabular}[c]{@{}c@{}} \textbf{LM} \end{tabular}}} & GPT-4o  &  1    & {95} & {549} & {0.6 ± 0.1} & {0.01 ± 0.01} & 44.2\% ±0.5 & 57.8\% ±0.6 & 80.6\% ±0.6 & 63.4\% ±0.2 & 87.5\% ±0.3 & 83.5\% ±0.4 & 18.1\% ±0.2 \\ 
       & & Claude 3.5 Sonnet  & 1 & {105} & {602}  & {1.4 ± 0.2} & {0.01 ± 0.01} & N/A & 52.7\% ±0.4 & 81.4\% ±0.4 & 68.7\% ±0.2 & 89.1\% ±0.2 & 82.5\% ±0.7 & 12.3\% ±0.4  \\ 
       & & Llama 3.1 405B  & 1 & {118} & {631} & {1.5 ± 0.1} & {0.01 ± 0.01} & 44.1\% ±0.3 & 40.7\% ±0.5 & 64.5\% ±0.7 & 66.0\% ±0.3 & 88.2\% ±0.2 & 85.0\% ±0.5 & 20.4\% ±0.5 \\ 
       \cmidrule{2-15}
       & {\multirow{4}{*}{\rotatebox{0}{\begin{tabular}[c]{@{}c@{}} \textbf{LM} \\ \textbf{Systems} \end{tabular}}}} & MoA  & 19 & {25,109} & {17,422} & {15.3 ± 0.3} & {0.06 ± 0.01} & 51.6\% ±0.6 & 65.0\% ±0.3 & 85.3\% ±0.3 & 62.3\% ±0.4 & 86.9\% ±0.2 & 82.9\% ±0.6 & 15.1\% ±0.5 \\ 
       & & {ADAS}                   & {52} & {72,804} & {44,872} & {58.8 ± 0.3} & {0.63 ± 0.04} & {66.3\%} ±{0.7} & {60.1\%} ±{0.5} & {85.4\%} ±{0.4} & {64.2\%} ±{0.2} & {87.0\%} ±{0.2} & {86.0\%} ±{0.8} & 23.7\% ±0.3 \\ 
        & & {AFlow}                   & {48} & {68,596} & {41,748} & {55.2 ± 0.4} & {0.59 ± 0.05} & {62.4\%} ±{0.2} & {57.8\%} ±{0.6} & {83.2\%} ±{0.6} & {63.5\%} ±{0.3} & {87.2\%} ±{0.4} & {84.5\%} ±{0.2} & 21.1\% ±0.6 \\ 
        & & {o1}               & {Unk.} & {112} & {Unk.} & {Unk.} & {0.52 ±0.05} & {56.3\%} ±{0.5} & {59.3\%} ±{0.5} & {81.7\%} ±{0.3} & {72.0\%} ±{0.4} & {87.5\%} ±{0.2} & \underline{92.7\% ±{0.5}} & \underline{31.5\% ±0.8} \\ 
       \midrule
       {\multirow{7}{*}{\rotatebox{90}{\begin{tabular}[c]{@{}c@{}}\textbf{Archon}\\ \end{tabular}}}} & \multirow{2}{*}{\rotatebox{0}{\begin{tabular}[c]{@{}c@{}} \textbf{Open} \\ \textbf{Source}  \end{tabular}}} & \begin{tabular}[c]{@{}c@{}}  General Purpose\end{tabular}     & 35 & {51,113} & {31,508} & {3.1 ± 0.3} & {0.12 ± 0.02} & 67.2\% ±0.4 & 63.3\% ±0.6 & 85.6\% ±0.5 & 65.3\% ±0.3 & 86.2\% ±0.2 & 87.5\% ±0.6 & 18.2\% ±0.4 \\ 
       & & \begin{tabular}[c]{@{}c@{}} Task Specific\end{tabular}        & 44 & {63,157} & {39,949} & {3.7 ± 0.3} & {0.15 ± 0.02} & 71.1\% ±0.6 & 68.1\% ±0.4 & {89.6\% ±0.4} & 67.5\% ±0.2 & 88.8\% ±0.3 & 89.5\% ±0.3 & 28.9\% ±0.9 \\ 
       \cmidrule{2-15}
       & \multirow{2}{*}{\rotatebox{0}{\begin{tabular}[c]{@{}c@{}} \textbf{Closed} \\ \textbf{Source}  \end{tabular}}} & \begin{tabular}[c]{@{}c@{}} General Purpose \end{tabular}      & 32 & {52,747} & {27,894} & {40.3 ± 0.5} & {0.44 ± 0.04} & 72.7\% ±0.3 & 63.9\% ±0.7 & 86.2\% ±0.7 & 67.5\% ±0.4 & 87.2\% ±0.2 & 87.9\% ±0.7 & 20.2\% ±0.6 \\ 
       & & \begin{tabular}[c]{@{}c@{}} Task Specific\end{tabular}        & 40 & {59,085} & {37,271} & {48.2 ± 0.4} & {0.49 ± 0.05} & \underline{77.0\% ±0.5} & \underline{68.9\% ±0.5} & 90.5\% ±0.3 & \underline{72.3\% ±0.3} & \underline{89.5\% ±0.3} & 92.1\% ±0.4 & 25.1\% ±0.6 \\ 
       \cmidrule{2-15}
       & \multirow{2}{*}{\rotatebox{0}{\begin{tabular}[c]{@{}c@{}} \textbf{All} \\ \textbf{Source}  \end{tabular}}} & \begin{tabular}[c]{@{}c@{}} General Purpose \end{tabular}      & 35 & {50,427} & {30,461} & {27.8 ± 0.4} & {0.32 ± 0.04} & 76.2\% ±0.7 & 66.4\% ±0.3 & \underline{89.8\% ±0.6} & 69.8\% ±0.2 & 87.3\% ±0.4 & 89.3\% ±0.5 & 23.4\% ±0.9  \\ 
       & & \begin{tabular}[c]{@{}c@{}} Task Specific\end{tabular}         & 39 & {58,250}  & {36,114} & {33.7 ± 0.6} & {0.37 ± 0.04} & \textbf{79.5\% ±0.4} & \textbf{69.0\% ±0.6} & \textbf{92.5\% ±0.5} & \textbf{72.7\% ±0.3} & \textbf{89.7\% ±0.2} & \textbf{93.5\% ±0.6} & \textbf{41.4\% ±0.7} \\ 
       \bottomrule
       \end{tabular}
   \caption{\textbf{\archon{}'s Strong Performance with Open Source, Closed Source, and All Source Models}: Consistent outperformance over SOTA LLMs and LM Systems across explored benchmarks.
   {The standard error numbers were calculated from 10 independent evaluation runs.} 
   $^*$MATH and CodeContests use a subset of their test sets for evaluation (Section \ref{sec:benchmarks_and_models}).
   }
   \label{tab:archon_main_table_results}
\end{table*}

\noindent \textbf{Benchmarks}: We evaluate our models with several benchmarks for instruction-following, reasoning, and coding: MT-Bench \citep{zheng2023judging}, AlpacaEval 2.0 \citep{alpaca_eval}, Arena Hard Auto \citep{arenahard2024}, MixEval \citep{ni2024mixevalderivingwisdomcrowd}, MixEval-Hard, MATH \citep{hendrycks2021measuring}, and CodeContests \citep{li2022competition}. 
We provide an overview of each dataset in \autoref{tab:benchmarks_overview}.
Since we perform automatic architecture search on a randomly sampled 20\% subset of each benchmark, we evaluate on the remaining held-out 80\% subset of the benchmark (\autoref{tab:archon_main_table_results}) (for \archon{} performances on the entire benchmarks, please see \autoref{tab:archon_full_dataset_results}).
The delta between the \archon{} performance on the entire benchmark vs. 80\% held-out subset is relatively small: only 0.44\%, on average, across these datasets with an S.D. of 0.20\%. 
For MATH, we evaluate a random sample of 200 problems from the dataset's test set.
For CodeContests, we evaluate on the 140 test set questions that do not include image tags in the problem description.

\noindent \textbf{Models}: {We test the efficacy of the \archon{} framework by creating different \archon{} architectures} across three model categories: 8B or less parameter models, 70B or more parameter models, and closed-source model APIs.
For our 8B and 70B+ models, we selected the top-10 performing chat models for each parameter range on the Chatbot Arena Leaderboard \citep{chiang2024chatbot} as of July 2024. 
For our \archon{} architectures, we explore multiple model types: open-source, closed-source, and \textit{all-source} (i.e. both open-source and closed-source available).
For our closed-source model APIs, we include GPT-4o, GPT-4-Turbo, Claude Opus 3.0, Claude Haiku 3.0, and Claude Sonnet 3.5.
We list and compare all of the models tested in the \archon{} framework in \autoref{tab:models_overview} and \autoref{tab:generator_and_fusion_rankings}.
For all the LLMs utilized and every \archon{} component, we set the generation temperature to $0.7$.
{As baselines, we compare \archon{} against both SOTA single-call LLMs (GPT-4o \citep{openai2024gpt4technicalreport}, Claude 3.5 Sonnet \citep{claude3}, and Llama 3.1 405B Instruct \citep{llama3modelcard}) as well as SOTA inference-time approaches (OpenAI's o1 \citep{openai2024o1}, MoA \citep{wang2024mixtureofagentsenhanceslargelanguage}, ADAS \citep{hu2024ADAS}, and AFlow \citep{zhang2024aflowautomatingagenticworkflow}).}

\noindent \textbf{Task-Specific and General-Purpose \archon{} Architectures}: We compare custom \archon{} architectures, specifically configured to a single evaluation dataset ("Task-specific \archon{} Architectures"), and a generalized \archon{} architecture configured to handle all the evaluation datasets ("General-purpose \archon{} Architectures") (\autoref{tab:archon_main_table_results}).
For our three model selection settings for \archon{} (i.e. open-source, closed-source, and all-source), we utilize automatic architecture search to find targeted \archon{} architectures for each task (7 architectures total) and find a single generalized \archon{} architecture for maximizing performance over all the tasks (\autoref{tab:archon_main_table_results}).
The benchmarks are concatenated together and shuffled for generalized \archon{} architecture search.
Importantly, all the \archon{} architectures utilized in Section \ref{sec:experiments} are automatically generated by our Bayesian architecture search technique, which searches over the hyperparameter search space for \archon{} as covered in Section \ref{sec:search_algorithms}.
For examples of targeted and generalized \archon{} architectures, please see Figure \ref{fig:example_archon_architecture} and Appendix \ref{sec:archon_architectures}.
For our architectures, we outline the average number of input tokens (i.e. combined total of tokens inputted over the entire architecture) and output tokens (i.e. combined total of tokens outputted over the entire architecture) for each category in Table \ref{tab:archon_main_table_results}.

\vspace{-2mm}
\subsection{\archon{} vs. Closed-Source LLMs and Other Inference-Time Architectures}
\label{sec:archon_results}
\vspace{-1mm}

\noindent \textbf{Task Performances}: We start by comparing \archon{} architectures to existing SOTA closed-source LLMs and inference-time architectures across a set of instruction-following, reasoning, and coding tasks. 
Based on our results in \autoref{tab:archon_main_table_results}, we find that \archon{} architectures consistently match or surpass existing approaches across all the benchmarks explored.
\archon{} architectures with open-source models demonstrate a 11.2\% average improvement over SOTA open-source approaches;
for its worst performance, our open-source \archon{} architectures are still 3.1\% above SOTA open-source approaches on AlpacaEval 2.0.
\archon{} architectures with closed-source models achieve SOTA performance across MT Bench, Arena-Hard-Auto, MixEval, and MixEval-Hard, leading to a 15.1\% average improvement over {closed-source LMs and a 8.4\% average improvement over open-source inference-time frameworks (i.e. MoA, ADAS, and AFlow).
Compared to o1 and o1-mini, \archon{}'s best targeted architectures beat them by 8.1\% and 9.7\%, on average, on MT Bench, AlpacaEval 2.0, Arena Hard Auto, MixEval, MixEval Hard, MATH, and CodeContests.
For approaches that use all models available, both open and closed-source, \archon{} achieves an average 10.9\% improvement over existing SOTA single-call LLMs {and an average 8.6\% improvement over existing inference-time frameworks.} 

\noindent \textbf{Compute Efficiency}: Compared to open-source inference-time frameworks (i.e. AFlow, ADAS, MoA), \archon{} is 20.0\% more inference call efficient while having higher performances on all benchmarks tested (\autoref{tab:archon_main_table_results}).
We also find that our best \archon{} architectures use 15.1\% less input tokens and 13.5\% less output tokens compared to the best alternative open-source inference-time frameworks.
When we utilize \archon{}'s architecture search technique with different token budgets (\autoref{fig:Archon_Budget_Analysis}), we find that the generated \archon{} architectures achieve 12.4\% higher performance than alternate baselines when given the same budget. 
Overall, the generalized all-source \archon{} architecture achieves 6.4\% better performance across all the tasks while being 31\% more token efficient than the best LM system baselines (\autoref{tab:archon_main_table_results}).
Furthermore, compared to the generalized all-source \archon{} architecture, the targeted all-source \archon{} architectures use 15.5\% and 18.6\% more input tokens and output tokens, respectively, but they achieve 8.4\% higher accuracies, on average. 
The targeted architectures are more compute intensive since they can further leverage additional LM operations towards a single set of specific task constraints (Appendix \ref{sec:archon_architectures}).

\noindent \textbf{Discovered Architectures}: We include the targeted and generalized \archon{} architectures in Appendix \ref{sec:archon_architectures} (\autoref{fig:funneling_architecture}).
The best performing all-source, general-purpose \archon{} architecture starts with a broad initial layer of our 10 best generators before four successive layers of critique and fusion with Qwen2 72B and Claude 3.5 Sonnet, respectively.
Each subsequent layer has fewer fuser models (i.e. 8, 6, and 4), leading to a "funneling" effect on the generations before the final output.
The best targeted architectures can vary by task.
For instruction-following and reasoning tasks, the targeted architectures tend to be multiple layers of critiquing and fusing with a diverse mix of LMs (\autoref{fig:funneling_architecture_open-source}).
For math tasks, the targeted architectures tend to consist of an initial broad set of generations before being reduced quickly to a chosen answer (\autoref{fig:claude_only}).
For coding tasks, the targeted architectures tend to focus on multiple iterations of generation, critique, and fusion over a single response before outputting an answer (\autoref{fig:llama_only}).
Besides the \archon{} architectures included in Appendix \ref{sec:archon_architectures}, we include all the generalized and targeted \archon{} architectures in our supplementary files.

{
To explore the efficacy of our general purpose \archon{} architectures, we evaluate them on three previously unseen tasks: GPQA \citep{rein2024gpqa}, MMLU \citep{hendrycks2021measuring}, and MMLU Pro \citep{wang2024mmlu}.
We find that our all-source general purpose \archon{} architecture captures 91 to 94\% of the task-specific \archon{} architectures performances on these benchmarks, suggesting that our architectures are more broadly applicable to out-of-domain tasks (\autoref{tab:generalization_test_on_GPQA_MMLU_MMLU_Pro}). 
The generalized ADAS and AFlow architectures only achieve 66\% and 74\% of their specialized architecture performance, respectively.
}

\begin{table}
   \centering
       \scriptsize
       \setlength{\tabcolsep}{1.8pt}
       \begin{tabular}{ccccc}
       \toprule
       & {\textbf{\begin{tabular}[c]{@{}c@{}}GPQA\\ Diamond\end{tabular}}} & {\textbf{MMLU$^*
$}} & {\textbf{\begin{tabular}[c]{@{}c@{}}MMLU\\Pro$^*
$\end{tabular}}} \\
       \midrule
       {\begin{tabular}[c]{@{}c@{}} All-Source Generalized AFlow 
       \end{tabular}}  & {37.1\%±0.2} & {53.0\%±0.4} & {43.4\%±0.4} \\
       {\begin{tabular}[c]{@{}c@{}} Task-Specific AFlow\end{tabular}}  & {52.4\%±0.1} & {71.8\%±0.5} & {62.9\%±0.5} \\
       {\begin{tabular}[c]{@{}c@{}} AFlow Performance Preservation\end{tabular}} & {70.8\%} & {73.8\%} & {67.0\%} \\
       \midrule
       {\begin{tabular}[c]{@{}c@{}} All-Source Generalized ADAS 
       \end{tabular}}  & {39.8\%±0.3} & {53.5\%±0.3} & {44.1\%±0.7} \\
       {\begin{tabular}[c]{@{}c@{}} Task-Specific ADAS\end{tabular}}  & {54.4\%±0.5} & {73.0\%±0.4} & {66.0\%±0.4} \\
       {\begin{tabular}[c]{@{}c@{}} ADAS Performance Preservation\end{tabular}} & {73.2\%} & {73.3\%} & {66.8\%} \\
       \midrule
       {\begin{tabular}[c]{@{}c@{}} All-Source Generalized \archon{} 
       \end{tabular}}  & {56.1\%±0.4} & {76.5\%±0.3} & {71.0\%±0.1} \\
       {\begin{tabular}[c]{@{}c@{}} Task-Specific \archon{}\end{tabular}}  & {61.2\%±0.5} & {81.5\%±0.3} & {75.4\%±0.4} \\
       {\begin{tabular}[c]{@{}c@{}} Archon Performance Preservation\end{tabular}} & {91.7\%} & {93.9\%} & {94.2\%} \\
       \bottomrule
       \end{tabular}
   \caption{{\textbf{Generalized \archon{} Architecture Strong Performance on Out-of-Domain Tasks}:
   The generalized \archon{} architectures achieved 91 to 95\% the performance of the specialized \archon{} architectures on GPQA, MMLU, and MMLU Pro, despite not being trained for these tasks.
   Standard error calculated from 10 independent evaluation runs.
   \textbf{$^*$}For MMLU and MMLU Pro, we use a randomly selected 500 query sample of the test set for evaluation.
   }}
   \label{tab:generalization_test_on_GPQA_MMLU_MMLU_Pro}
\end{table}

\begin{figure*}[]
   \centering
   \includegraphics[width=\linewidth]{figures/PFLOPs_vs_Performance_v2.pdf}
   \caption{{\textbf{\archon{}'s Performance Exceeds Baselines across FLOP Budgets}: 
   Across different FLOP budgets (Section \ref{sec:search_algorithms}), we compare \archon{} architectures against top-performing inference-time system baselines.
   The MoA architecture and OpenAI's o1 are static so they use the same number of tokens across budgets.
   The results were averaged over 10 independent evaluation runs. $^*$MATH and CodeContests use a subset of their test sets for evaluation (Section \ref{sec:benchmarks_and_models}).
   }
   }
   \label{fig:PFLOPs_vs_Performance}
\end{figure*}

\vspace{-.1in} 
\subsection{\archon{} by Task} 
\label{sec:archon_by_task}

\noindent \textbf{Instruction-Following and Reasoning}: 
On MT Bench, AlpacaEval 2.0, and Arena-Hard-Auto, open-source \archon{} architectures outperform current open-source baselines by 10.5\%, on average, while closed-source \archon{} outperforms current closed-source baselines by 14.6\% (\autoref{tab:archon_main_table_results}).
With \archon{}, multiple models used for Generators and the depth of fusion layers lead to performance boosts on instruction-following tasks, increasing the richness of responses and allowing multiple iterations for step-by-step instruction-following (\autoref{tab:jaccard_similarities}).
For reasoning, while the performance boost from \archon{} is smaller when we consider the \textit{aggregate} scores for MixEval and MixEval-Hard, we do see meaningful increases in performance when we create inference-time architectures for each individual task under MixEval and MixEval-Hard (\autoref{tab:mixeval_subdataset_results}; \autoref{tab:mixeval_hard_subdataset_results}).
When we create individual \archon{} architectures for each subtask, we see 3.7 and 8.9 percentage point increases in accuracy, on average, for MixEval and MixEval-Hard, respectively.
This finding suggests that reasoning tasks (e.g. math, sciences, logic) require more individualized inference-time architectures.

\noindent \textbf{Coding}: 
We have observed that ensembling, fusion, and ranking techniques have limited impact on CodeContests (\autoref{fig:compound_LM_ablations}).
For example, when we apply the general all-source architecture from \autoref{tab:benchmarks_overview} to CodeContests problems, we achieve small gains from \archon{} (see \autoref{tab:archon_main_table_results}). 
One contributing factor is that, unlike the distribution of instruction-following/reasoning tasks, coding tasks tend to have one or two LLMs that perform substantially better than the rest of models (\autoref{tab:generator_and_fusion_rankings}).
However, when we add unit test generation/evaluation, and scale the number of samples,  
\archon{}'s performance on CodeContests improves significantly (\autoref{tab:archon_main_table_results}), allowing us to boost GPT-4o Pass@1 performance by 44.3\% for Pass@1 (from 40 to 58 out of 140 questions).
For model-based unit test generation/evaluation, we generate 5 unit tests and use the LM to evaluate each candidate response against the generated unit tests, allowing us to rank the different candidate responses (details are provided in Section \ref{sec:lm_construction_and_prompts})

\vspace{-.1in}
\subsection{Discussion} 
\label{sec:limitations_and_future_work}

\noindent \textbf{Impact of Model Size}: 
The \archon{} framework is most effective when utilizing LLMs with 70B+ parameters.
When we build \archon{} architectures with 7B open-source models, we can boost task performance over the best individual 7B LM by 7.5\%, on average, compared to the best individual 7B model (\autoref{tab:archon_with_7b}).
Across tasks, 7B models work well for ranking but are less effective for critique and fusion.

\noindent \textbf{Latency and Costs}: 
Since \archon{} architectures make multiple LLM API calls successively for different operations 
it can take 5x more time and money than a single LLM API call (\autoref{tab:model_costs}; \autoref{tab:archon_costs}). 
Note that these increases in compute costs and latency translate to higher quality responses, and can be justified in many application domains, such as science, programming, and complex agentic tasks \citep{rein2023gpqagraduatelevelgoogleproofqa, mialon2023gaiabenchmarkgeneralai}. 
Furthermore, LLM vendors are rapidly decreasing their inference costs (\autoref{tab:model_costs}). 
For tasks in which speed is most preferred, future work should explore how distillation strategies \citep{sreenivas2024llmpruningdistillationpractice, deepseekai2025deepseekr1incentivizingreasoningcapability} could be used to pack the aggregate knowledge of \archon{} architectures into a smaller LM.

\section*{Acknowledgments}

We thank Simran Arora, Daniel Biderman, Bradley Brown, Ryan Ehrlich, Sabri Eyuboglu, Jordan Juravsky, Jerry Liu, Avanika Narayan, Benjamin Spector, Alyssa Unell, Benjamin Viggiano, and Michael Zhang for their constructive feedback during the composition of the paper.
We would also like to thank our collaborators at the Stanford Artificial Intelligence Laboratory
(SAIL) and TogetherAI.

We gratefully acknowledge the support of NIH under No. U54EB020405 (Mobilize); NSF under Nos. CCF2247015 (Hardware-Aware), CCF1763315 (Beyond Sparsity), CCF1563078 (Volume to Velocity), and 1937301 (RTML); US DEVCOM ARL under Nos. W911NF-23-2-0184 (Long-context) and W911NF-21-2-0251 (Interactive Human-AI Teaming); ONR under No. N000142312633 (Deep Signal Processing); Stanford HAI under No. 247183; Google DeepMind; Google Research; Google Cloud; NXP; Xilinx; LETI-CEA; Intel; IBM; Microsoft; NEC; Toshiba; TSMC; ARM; Hitachi; BASF; Accenture; Ericsson; Qualcomm; Analog Devices; Salesforce; Total; the HAI-GCP Cloud Credits for Research program; the Stanford Data Science Initiative (SDSI); members of the Stanford DAWN project: Meta, Google, and VMWare; and members of the Stanford SEAMS project: IBM and Felicis.
The U.S. Government is authorized to reproduce and distribute reprints for Governmental
purposes notwithstanding any copyright notation thereon. Any opinions, findings, and conclusions
or recommendations expressed in this material are those of the authors and do not necessarily
reflect the views, policies, or endorsements, either expressed or implied, of NIH, ONR, or the U.S.
Government.

\bibliography{iclr2025_conference}

\begin{thebibliography}{48}
\providecommand{\natexlab}[1]{#1}
\providecommand{\url}[1]{\texttt{#1}}
\expandafter\ifx\csname urlstyle\endcsname\relax
  \providecommand{\doi}[1]{doi: #1}\else
  \providecommand{\doi}{doi: \begingroup \urlstyle{rm}\Url}\fi

\bibitem[Abdin et~al.(2024)Abdin, Jacobs, Awan, Aneja, Awadallah, Awadalla, Bach, Bahree, Bakhtiari, Behl, et~al.]{abdin2024phi}
Abdin, M., Jacobs, S.~A., Awan, A.~A., Aneja, J., Awadallah, A., Awadalla, H., Bach, N., Bahree, A., Bakhtiari, A., Behl, H., et~al.
\newblock Phi-3 technical report: A highly capable language model locally on your phone.
\newblock \emph{arXiv preprint arXiv:2404.14219}, 2024.

\bibitem[Anthropic(2024)]{claude3}
Anthropic.
\newblock The claude 3 model family: Opus, sonnet, haiku.
\newblock \emph{ArXiv}, 2024.

\bibitem[at~Meta(2024)]{llama3modelcard}
at~Meta, A.
\newblock Llama 3 model card.
\newblock \emph{ArXiv}, 2024.
\newblock URL \url{https://github.com/meta-llama/llama3/blob/main/MODEL_CARD.md}.

\bibitem[Bai et~al.(2023)Bai, Bai, Chu, Cui, Dang, Deng, Fan, Ge, Han, Huang, Hui, Ji, Li, Lin, Lin, Liu, Liu, Lu, Lu, Ma, Men, Ren, Ren, Tan, Tan, Tu, Wang, Wang, Wang, Wu, Xu, Xu, Yang, Yang, Yang, Yang, Yao, Yu, Yuan, Yuan, Zhang, Zhang, Zhang, Zhang, Zhou, Zhou, Zhou, and Zhu]{qwen}
Bai, J., Bai, S., Chu, Y., Cui, Z., Dang, K., Deng, X., Fan, Y., Ge, W., Han, Y., Huang, F., Hui, B., Ji, L., Li, M., Lin, J., Lin, R., Liu, D., Liu, G., Lu, C., Lu, K., Ma, J., Men, R., Ren, X., Ren, X., Tan, C., Tan, S., Tu, J., Wang, P., Wang, S., Wang, W., Wu, S., Xu, B., Xu, J., Yang, A., Yang, H., Yang, J., Yang, S., Yao, Y., Yu, B., Yuan, H., Yuan, Z., Zhang, J., Zhang, X., Zhang, Y., Zhang, Z., Zhou, C., Zhou, J., Zhou, X., and Zhu, T.
\newblock Qwen technical report.
\newblock \emph{arXiv preprint arXiv:2309.16609}, 2023.

\bibitem[Bai et~al.(2022)Bai, Kadavath, Kundu, Askell, Kernion, Jones, Chen, Goldie, Mirhoseini, McKinnon, Chen, Olsson, Olah, Hernandez, Drain, Ganguli, Li, Tran-Johnson, Perez, Kerr, Mueller, Ladish, Landau, Ndousse, Lukosuite, Lovitt, Sellitto, Elhage, Schiefer, Mercado, DasSarma, Lasenby, Larson, Ringer, Johnston, Kravec, Showk, Fort, Lanham, Telleen-Lawton, Conerly, Henighan, Hume, Bowman, Hatfield-Dodds, Mann, Amodei, Joseph, McCandlish, Brown, and Kaplan]{bai2022constitutionalaiharmlessnessai}
Bai, Y., Kadavath, S., Kundu, S., Askell, A., Kernion, J., Jones, A., Chen, A., Goldie, A., Mirhoseini, A., McKinnon, C., Chen, C., Olsson, C., Olah, C., Hernandez, D., Drain, D., Ganguli, D., Li, D., Tran-Johnson, E., Perez, E., Kerr, J., Mueller, J., Ladish, J., Landau, J., Ndousse, K., Lukosuite, K., Lovitt, L., Sellitto, M., Elhage, N., Schiefer, N., Mercado, N., DasSarma, N., Lasenby, R., Larson, R., Ringer, S., Johnston, S., Kravec, S., Showk, S.~E., Fort, S., Lanham, T., Telleen-Lawton, T., Conerly, T., Henighan, T., Hume, T., Bowman, S.~R., Hatfield-Dodds, Z., Mann, B., Amodei, D., Joseph, N., McCandlish, S., Brown, T., and Kaplan, J.
\newblock Constitutional ai: Harmlessness from ai feedback, 2022.
\newblock URL \url{https://arxiv.org/abs/2212.08073}.

\bibitem[Brown et~al.(2024)Brown, Juravsky, Ehrlich, Clark, Le, Ré, and Mirhoseini]{brown2024largelanguagemonkeysscaling}
Brown, B., Juravsky, J., Ehrlich, R., Clark, R., Le, Q.~V., Ré, C., and Mirhoseini, A.
\newblock Large language monkeys: Scaling inference compute with repeated sampling, 2024.
\newblock URL \url{https://arxiv.org/abs/2407.21787}.

\bibitem[Chen et~al.(2024)Chen, Davis, Hanin, Bailis, Stoica, Zaharia, and Zou]{chen2024llmcallsneedscaling}
Chen, L., Davis, J.~Q., Hanin, B., Bailis, P., Stoica, I., Zaharia, M., and Zou, J.
\newblock Are more llm calls all you need? towards scaling laws of compound inference systems, 2024.
\newblock URL \url{https://arxiv.org/abs/2403.02419}.

\bibitem[Chiang et~al.(2024)Chiang, Zheng, Sheng, Angelopoulos, Li, Li, Zhang, Zhu, Jordan, Gonzalez, and Stoica]{chiang2024chatbot}
Chiang, W.-L., Zheng, L., Sheng, Y., Angelopoulos, A.~N., Li, T., Li, D., Zhang, H., Zhu, B., Jordan, M., Gonzalez, J.~E., and Stoica, I.
\newblock Chatbot arena: An open platform for evaluating llms by human preference, 2024.

\bibitem[Databricks(2024)]{dbrx}
Databricks.
\newblock Dbrx technical report.
\newblock 2024.

\bibitem[Davis et~al.(2024)Davis, Hanin, Chen, Bailis, Stoica, and Zaharia]{davis2024networksnetworkscomplexityclass}
Davis, J.~Q., Hanin, B., Chen, L., Bailis, P., Stoica, I., and Zaharia, M.
\newblock Networks of networks: Complexity class principles applied to compound ai systems design, 2024.
\newblock URL \url{https://arxiv.org/abs/2407.16831}.

\bibitem[DeepSeek-AI et~al.(2025)DeepSeek-AI, Guo, Yang, Zhang, Song, Zhang, Xu, Zhu, Ma, Wang, Bi, Zhang, Yu, Wu, Wu, Gou, Shao, Li, Gao, Liu, Xue, Wang, Wu, Feng, Lu, Zhao, Deng, Zhang, Ruan, Dai, Chen, Ji, Li, Lin, Dai, Luo, Hao, Chen, Li, Zhang, Bao, Xu, Wang, Ding, Xin, Gao, Qu, Li, Guo, Li, Wang, Chen, Yuan, Qiu, Li, Cai, Ni, Liang, Chen, Dong, Hu, Gao, Guan, Huang, Yu, Wang, Zhang, Zhao, Wang, Zhang, Xu, Xia, Zhang, Zhang, Tang, Li, Wang, Li, Tian, Huang, Zhang, Wang, Chen, Du, Ge, Zhang, Pan, Wang, Chen, Jin, Chen, Lu, Zhou, Chen, Ye, Wang, Yu, Zhou, Pan, Li, Zhou, Wu, Ye, Yun, Pei, Sun, Wang, Zeng, Zhao, Liu, Liang, Gao, Yu, Zhang, Xiao, An, Liu, Wang, Chen, Nie, Cheng, Liu, Xie, Liu, Yang, Li, Su, Lin, Li, Jin, Shen, Chen, Sun, Wang, Song, Zhou, Wang, Shan, Li, Wang, Wei, Zhang, Xu, Li, Zhao, Sun, Wang, Yu, Zhang, Shi, Xiong, He, Piao, Wang, Tan, Ma, Liu, Guo, Ou, Wang, Gong, Zou, He, Xiong, Luo, You, Liu, Zhou, Zhu, Xu, Huang, Li, Zheng, Zhu, Ma, Tang, Zha, Yan, Ren, Ren, Sha, Fu, Xu, Xie, Zhang,
  Hao, Ma, Yan, Wu, Gu, Zhu, Liu, Li, Xie, Song, Pan, Huang, Xu, Zhang, and Zhang]{deepseekai2025deepseekr1incentivizingreasoningcapability}
DeepSeek-AI, Guo, D., Yang, D., Zhang, H., Song, J., Zhang, R., Xu, R., Zhu, Q., Ma, S., Wang, P., Bi, X., Zhang, X., Yu, X., Wu, Y., Wu, Z.~F., Gou, Z., Shao, Z., Li, Z., Gao, Z., Liu, A., Xue, B., Wang, B., Wu, B., Feng, B., Lu, C., Zhao, C., Deng, C., Zhang, C., Ruan, C., Dai, D., Chen, D., Ji, D., Li, E., Lin, F., Dai, F., Luo, F., Hao, G., Chen, G., Li, G., Zhang, H., Bao, H., Xu, H., Wang, H., Ding, H., Xin, H., Gao, H., Qu, H., Li, H., Guo, J., Li, J., Wang, J., Chen, J., Yuan, J., Qiu, J., Li, J., Cai, J.~L., Ni, J., Liang, J., Chen, J., Dong, K., Hu, K., Gao, K., Guan, K., Huang, K., Yu, K., Wang, L., Zhang, L., Zhao, L., Wang, L., Zhang, L., Xu, L., Xia, L., Zhang, M., Zhang, M., Tang, M., Li, M., Wang, M., Li, M., Tian, N., Huang, P., Zhang, P., Wang, Q., Chen, Q., Du, Q., Ge, R., Zhang, R., Pan, R., Wang, R., Chen, R.~J., Jin, R.~L., Chen, R., Lu, S., Zhou, S., Chen, S., Ye, S., Wang, S., Yu, S., Zhou, S., Pan, S., Li, S.~S., Zhou, S., Wu, S., Ye, S., Yun, T., Pei, T., Sun, T., Wang, T., Zeng, W.,
  Zhao, W., Liu, W., Liang, W., Gao, W., Yu, W., Zhang, W., Xiao, W.~L., An, W., Liu, X., Wang, X., Chen, X., Nie, X., Cheng, X., Liu, X., Xie, X., Liu, X., Yang, X., Li, X., Su, X., Lin, X., Li, X.~Q., Jin, X., Shen, X., Chen, X., Sun, X., Wang, X., Song, X., Zhou, X., Wang, X., Shan, X., Li, Y.~K., Wang, Y.~Q., Wei, Y.~X., Zhang, Y., Xu, Y., Li, Y., Zhao, Y., Sun, Y., Wang, Y., Yu, Y., Zhang, Y., Shi, Y., Xiong, Y., He, Y., Piao, Y., Wang, Y., Tan, Y., Ma, Y., Liu, Y., Guo, Y., Ou, Y., Wang, Y., Gong, Y., Zou, Y., He, Y., Xiong, Y., Luo, Y., You, Y., Liu, Y., Zhou, Y., Zhu, Y.~X., Xu, Y., Huang, Y., Li, Y., Zheng, Y., Zhu, Y., Ma, Y., Tang, Y., Zha, Y., Yan, Y., Ren, Z.~Z., Ren, Z., Sha, Z., Fu, Z., Xu, Z., Xie, Z., Zhang, Z., Hao, Z., Ma, Z., Yan, Z., Wu, Z., Gu, Z., Zhu, Z., Liu, Z., Li, Z., Xie, Z., Song, Z., Pan, Z., Huang, Z., Xu, Z., Zhang, Z., and Zhang, Z.
\newblock Deepseek-r1: Incentivizing reasoning capability in llms via reinforcement learning, 2025.
\newblock URL \url{https://arxiv.org/abs/2501.12948}.

\bibitem[Dubey et~al.(2024)Dubey, Jauhri, Pandey, Kadian, Al-Dahle, Letman, Mathur, Schelten, Yang, Fan, Goyal, Hartshorn, Yang, Mitra, Sravankumar, Korenev, Hinsvark, Rao, Zhang, Rodriguez, Gregerson, Spataru, Roziere, Biron, Tang, Chern, Caucheteux, Nayak, Bi, Marra, McConnell, Keller, Touret, Wu, Wong, Ferrer, Nikolaidis, Allonsius, Song, Pintz, Livshits, Esiobu, Choudhary, Mahajan, Garcia-Olano, Perino, Hupkes, Lakomkin, AlBadawy, Lobanova, Dinan, Smith, Radenovic, Zhang, Synnaeve, Lee, Anderson, Nail, Mialon, Pang, Cucurell, Nguyen, Korevaar, Xu, Touvron, Zarov, Ibarra, Kloumann, Misra, Evtimov, Copet, Lee, Geffert, Vranes, Park, Mahadeokar, Shah, van~der Linde, Billock, Hong, Lee, Fu, Chi, Huang, Liu, Wang, Yu, Bitton, Spisak, Park, Rocca, Johnstun, Saxe, Jia, Alwala, Upasani, Plawiak, Li, Heafield, Stone, El-Arini, Iyer, Malik, Chiu, Bhalla, Rantala-Yeary, van~der Maaten, Chen, Tan, Jenkins, Martin, Madaan, Malo, Blecher, Landzaat, de~Oliveira, Muzzi, Pasupuleti, Singh, Paluri, Kardas, Oldham, Rita,
  Pavlova, Kambadur, Lewis, Si, Singh, Hassan, Goyal, Torabi, Bashlykov, Bogoychev, Chatterji, Duchenne, Çelebi, Alrassy, Zhang, Li, Vasic, Weng, Bhargava, Dubal, Krishnan, Koura, Xu, He, Dong, Srinivasan, Ganapathy, Calderer, Cabral, Stojnic, Raileanu, Girdhar, Patel, Sauvestre, Polidoro, Sumbaly, Taylor, Silva, Hou, Wang, Hosseini, Chennabasappa, Singh, Bell, Kim, Edunov, Nie, Narang, Raparthy, Shen, Wan, Bhosale, Zhang, Vandenhende, Batra, Whitman, Sootla, Collot, Gururangan, Borodinsky, Herman, Fowler, Sheasha, Georgiou, Scialom, Speckbacher, Mihaylov, Xiao, Karn, Goswami, Gupta, Ramanathan, Kerkez, Gonguet, Do, Vogeti, Petrovic, Chu, Xiong, Fu, Meers, Martinet, Wang, Tan, Xie, Jia, Wang, Goldschlag, Gaur, Babaei, Wen, Song, Zhang, Li, Mao, Coudert, Yan, Chen, Papakipos, Singh, Grattafiori, Jain, Kelsey, Shajnfeld, Gangidi, Victoria, Goldstand, Menon, Sharma, Boesenberg, Vaughan, Baevski, Feinstein, Kallet, Sangani, Yunus, Lupu, Alvarado, Caples, Gu, Ho, Poulton, Ryan, Ramchandani, Franco, Saraf,
  Chowdhury, Gabriel, Bharambe, Eisenman, Yazdan, James, Maurer, Leonhardi, Huang, Loyd, Paola, Paranjape, Liu, Wu, Ni, Hancock, Wasti, Spence, Stojkovic, Gamido, Montalvo, Parker, Burton, Mejia, Wang, Kim, Zhou, Hu, Chu, Cai, Tindal, Feichtenhofer, Civin, Beaty, Kreymer, Li, Wyatt, Adkins, Xu, Testuggine, David, Parikh, Liskovich, Foss, Wang, Le, Holland, Dowling, Jamil, Montgomery, Presani, Hahn, Wood, Brinkman, Arcaute, Dunbar, Smothers, Sun, Kreuk, Tian, Ozgenel, Caggioni, Guzmán, Kanayet, Seide, Florez, Schwarz, Badeer, Swee, Halpern, Thattai, Herman, Sizov, Guangyi, Zhang, Lakshminarayanan, Shojanazeri, Zou, Wang, Zha, Habeeb, Rudolph, Suk, Aspegren, Goldman, Molybog, Tufanov, Veliche, Gat, Weissman, Geboski, Kohli, Asher, Gaya, Marcus, Tang, Chan, Zhen, Reizenstein, Teboul, Zhong, Jin, Yang, Cummings, Carvill, Shepard, McPhie, Torres, Ginsburg, Wang, Wu, U, Saxena, Prasad, Khandelwal, Zand, Matosich, Veeraraghavan, Michelena, Li, Huang, Chawla, Lakhotia, Huang, Chen, Garg, A, Silva, Bell, Zhang, Guo,
  Yu, Moshkovich, Wehrstedt, Khabsa, Avalani, Bhatt, Tsimpoukelli, Mankus, Hasson, Lennie, Reso, Groshev, Naumov, Lathi, Keneally, Seltzer, Valko, Restrepo, Patel, Vyatskov, Samvelyan, Clark, Macey, Wang, Hermoso, Metanat, Rastegari, Bansal, Santhanam, Parks, White, Bawa, Singhal, Egebo, Usunier, Laptev, Dong, Zhang, Cheng, Chernoguz, Hart, Salpekar, Kalinli, Kent, Parekh, Saab, Balaji, Rittner, Bontrager, Roux, Dollar, Zvyagina, Ratanchandani, Yuvraj, Liang, Alao, Rodriguez, Ayub, Murthy, Nayani, Mitra, Li, Hogan, Battey, Wang, Maheswari, Howes, Rinott, Bondu, Datta, Chugh, Hunt, Dhillon, Sidorov, Pan, Verma, Yamamoto, Ramaswamy, Lindsay, Lindsay, Feng, Lin, Zha, Shankar, Zhang, Zhang, Wang, Agarwal, Sajuyigbe, Chintala, Max, Chen, Kehoe, Satterfield, Govindaprasad, Gupta, Cho, Virk, Subramanian, Choudhury, Goldman, Remez, Glaser, Best, Kohler, Robinson, Li, Zhang, Matthews, Chou, Shaked, Vontimitta, Ajayi, Montanez, Mohan, Kumar, Mangla, Ionescu, Poenaru, Mihailescu, Ivanov, Li, Wang, Jiang, Bouaziz,
  Constable, Tang, Wang, Wu, Wang, Xia, Wu, Gao, Chen, Hu, Jia, Qi, Li, Zhang, Zhang, Adi, Nam, Yu, Wang, Hao, Qian, He, Rait, DeVito, Rosnbrick, Wen, Yang, and Zhao]{dubey2024llama3herdmodels}
Dubey, A., Jauhri, A., Pandey, A., Kadian, A., Al-Dahle, A., Letman, A., Mathur, A., Schelten, A., Yang, A., Fan, A., Goyal, A., Hartshorn, A., Yang, A., Mitra, A., Sravankumar, A., Korenev, A., Hinsvark, A., Rao, A., Zhang, A., Rodriguez, A., Gregerson, A., Spataru, A., Roziere, B., Biron, B., Tang, B., Chern, B., Caucheteux, C., Nayak, C., Bi, C., Marra, C., McConnell, C., Keller, C., Touret, C., Wu, C., Wong, C., Ferrer, C.~C., Nikolaidis, C., Allonsius, D., Song, D., Pintz, D., Livshits, D., Esiobu, D., Choudhary, D., Mahajan, D., Garcia-Olano, D., Perino, D., Hupkes, D., Lakomkin, E., AlBadawy, E., Lobanova, E., Dinan, E., Smith, E.~M., Radenovic, F., Zhang, F., Synnaeve, G., Lee, G., Anderson, G.~L., Nail, G., Mialon, G., Pang, G., Cucurell, G., Nguyen, H., Korevaar, H., Xu, H., Touvron, H., Zarov, I., Ibarra, I.~A., Kloumann, I., Misra, I., Evtimov, I., Copet, J., Lee, J., Geffert, J., Vranes, J., Park, J., Mahadeokar, J., Shah, J., van~der Linde, J., Billock, J., Hong, J., Lee, J., Fu, J., Chi, J.,
  Huang, J., Liu, J., Wang, J., Yu, J., Bitton, J., Spisak, J., Park, J., Rocca, J., Johnstun, J., Saxe, J., Jia, J., Alwala, K.~V., Upasani, K., Plawiak, K., Li, K., Heafield, K., Stone, K., El-Arini, K., Iyer, K., Malik, K., Chiu, K., Bhalla, K., Rantala-Yeary, L., van~der Maaten, L., Chen, L., Tan, L., Jenkins, L., Martin, L., Madaan, L., Malo, L., Blecher, L., Landzaat, L., de~Oliveira, L., Muzzi, M., Pasupuleti, M., Singh, M., Paluri, M., Kardas, M., Oldham, M., Rita, M., Pavlova, M., Kambadur, M., Lewis, M., Si, M., Singh, M.~K., Hassan, M., Goyal, N., Torabi, N., Bashlykov, N., Bogoychev, N., Chatterji, N., Duchenne, O., Çelebi, O., Alrassy, P., Zhang, P., Li, P., Vasic, P., Weng, P., Bhargava, P., Dubal, P., Krishnan, P., Koura, P.~S., Xu, P., He, Q., Dong, Q., Srinivasan, R., Ganapathy, R., Calderer, R., Cabral, R.~S., Stojnic, R., Raileanu, R., Girdhar, R., Patel, R., Sauvestre, R., Polidoro, R., Sumbaly, R., Taylor, R., Silva, R., Hou, R., Wang, R., Hosseini, S., Chennabasappa, S., Singh, S.,
  Bell, S., Kim, S.~S., Edunov, S., Nie, S., Narang, S., Raparthy, S., Shen, S., Wan, S., Bhosale, S., Zhang, S., Vandenhende, S., Batra, S., Whitman, S., Sootla, S., Collot, S., Gururangan, S., Borodinsky, S., Herman, T., Fowler, T., Sheasha, T., Georgiou, T., Scialom, T., Speckbacher, T., Mihaylov, T., Xiao, T., Karn, U., Goswami, V., Gupta, V., Ramanathan, V., Kerkez, V., Gonguet, V., Do, V., Vogeti, V., Petrovic, V., Chu, W., Xiong, W., Fu, W., Meers, W., Martinet, X., Wang, X., Tan, X.~E., Xie, X., Jia, X., Wang, X., Goldschlag, Y., Gaur, Y., Babaei, Y., Wen, Y., Song, Y., Zhang, Y., Li, Y., Mao, Y., Coudert, Z.~D., Yan, Z., Chen, Z., Papakipos, Z., Singh, A., Grattafiori, A., Jain, A., Kelsey, A., Shajnfeld, A., Gangidi, A., Victoria, A., Goldstand, A., Menon, A., Sharma, A., Boesenberg, A., Vaughan, A., Baevski, A., Feinstein, A., Kallet, A., Sangani, A., Yunus, A., Lupu, A., Alvarado, A., Caples, A., Gu, A., Ho, A., Poulton, A., Ryan, A., Ramchandani, A., Franco, A., Saraf, A., Chowdhury, A., Gabriel,
  A., Bharambe, A., Eisenman, A., Yazdan, A., James, B., Maurer, B., Leonhardi, B., Huang, B., Loyd, B., Paola, B.~D., Paranjape, B., Liu, B., Wu, B., Ni, B., Hancock, B., Wasti, B., Spence, B., Stojkovic, B., Gamido, B., Montalvo, B., Parker, C., Burton, C., Mejia, C., Wang, C., Kim, C., Zhou, C., Hu, C., Chu, C.-H., Cai, C., Tindal, C., Feichtenhofer, C., Civin, D., Beaty, D., Kreymer, D., Li, D., Wyatt, D., Adkins, D., Xu, D., Testuggine, D., David, D., Parikh, D., Liskovich, D., Foss, D., Wang, D., Le, D., Holland, D., Dowling, E., Jamil, E., Montgomery, E., Presani, E., Hahn, E., Wood, E., Brinkman, E., Arcaute, E., Dunbar, E., Smothers, E., Sun, F., Kreuk, F., Tian, F., Ozgenel, F., Caggioni, F., Guzmán, F., Kanayet, F., Seide, F., Florez, G.~M., Schwarz, G., Badeer, G., Swee, G., Halpern, G., Thattai, G., Herman, G., Sizov, G., Guangyi, Zhang, Lakshminarayanan, G., Shojanazeri, H., Zou, H., Wang, H., Zha, H., Habeeb, H., Rudolph, H., Suk, H., Aspegren, H., Goldman, H., Molybog, I., Tufanov, I.,
  Veliche, I.-E., Gat, I., Weissman, J., Geboski, J., Kohli, J., Asher, J., Gaya, J.-B., Marcus, J., Tang, J., Chan, J., Zhen, J., Reizenstein, J., Teboul, J., Zhong, J., Jin, J., Yang, J., Cummings, J., Carvill, J., Shepard, J., McPhie, J., Torres, J., Ginsburg, J., Wang, J., Wu, K., U, K.~H., Saxena, K., Prasad, K., Khandelwal, K., Zand, K., Matosich, K., Veeraraghavan, K., Michelena, K., Li, K., Huang, K., Chawla, K., Lakhotia, K., Huang, K., Chen, L., Garg, L., A, L., Silva, L., Bell, L., Zhang, L., Guo, L., Yu, L., Moshkovich, L., Wehrstedt, L., Khabsa, M., Avalani, M., Bhatt, M., Tsimpoukelli, M., Mankus, M., Hasson, M., Lennie, M., Reso, M., Groshev, M., Naumov, M., Lathi, M., Keneally, M., Seltzer, M.~L., Valko, M., Restrepo, M., Patel, M., Vyatskov, M., Samvelyan, M., Clark, M., Macey, M., Wang, M., Hermoso, M.~J., Metanat, M., Rastegari, M., Bansal, M., Santhanam, N., Parks, N., White, N., Bawa, N., Singhal, N., Egebo, N., Usunier, N., Laptev, N.~P., Dong, N., Zhang, N., Cheng, N., Chernoguz, O.,
  Hart, O., Salpekar, O., Kalinli, O., Kent, P., Parekh, P., Saab, P., Balaji, P., Rittner, P., Bontrager, P., Roux, P., Dollar, P., Zvyagina, P., Ratanchandani, P., Yuvraj, P., Liang, Q., Alao, R., Rodriguez, R., Ayub, R., Murthy, R., Nayani, R., Mitra, R., Li, R., Hogan, R., Battey, R., Wang, R., Maheswari, R., Howes, R., Rinott, R., Bondu, S.~J., Datta, S., Chugh, S., Hunt, S., Dhillon, S., Sidorov, S., Pan, S., Verma, S., Yamamoto, S., Ramaswamy, S., Lindsay, S., Lindsay, S., Feng, S., Lin, S., Zha, S.~C., Shankar, S., Zhang, S., Zhang, S., Wang, S., Agarwal, S., Sajuyigbe, S., Chintala, S., Max, S., Chen, S., Kehoe, S., Satterfield, S., Govindaprasad, S., Gupta, S., Cho, S., Virk, S., Subramanian, S., Choudhury, S., Goldman, S., Remez, T., Glaser, T., Best, T., Kohler, T., Robinson, T., Li, T., Zhang, T., Matthews, T., Chou, T., Shaked, T., Vontimitta, V., Ajayi, V., Montanez, V., Mohan, V., Kumar, V.~S., Mangla, V., Ionescu, V., Poenaru, V., Mihailescu, V.~T., Ivanov, V., Li, W., Wang, W., Jiang, W.,
  Bouaziz, W., Constable, W., Tang, X., Wang, X., Wu, X., Wang, X., Xia, X., Wu, X., Gao, X., Chen, Y., Hu, Y., Jia, Y., Qi, Y., Li, Y., Zhang, Y., Zhang, Y., Adi, Y., Nam, Y., Yu, Wang, Hao, Y., Qian, Y., He, Y., Rait, Z., DeVito, Z., Rosnbrick, Z., Wen, Z., Yang, Z., and Zhao, Z.
\newblock The llama 3 herd of models, 2024.
\newblock URL \url{https://arxiv.org/abs/2407.21783}.

\bibitem[Guo et~al.(2024)Guo, Zhu, Yang, Xie, Dong, Zhang, Chen, Bi, Wu, Li, Luo, Xiong, and Liang]{guo2024deepseekcoderlargelanguagemodel}
Guo, D., Zhu, Q., Yang, D., Xie, Z., Dong, K., Zhang, W., Chen, G., Bi, X., Wu, Y., Li, Y.~K., Luo, F., Xiong, Y., and Liang, W.
\newblock Deepseek-coder: When the large language model meets programming -- the rise of code intelligence, 2024.
\newblock URL \url{https://arxiv.org/abs/2401.14196}.

\bibitem[Hartford(2024)]{dolphin}
Hartford, E.
\newblock dolphin-2.2.1-mistral-7b.
\newblock January 2024.

\bibitem[Hendrycks et~al.(2021)Hendrycks, Burns, Kadavath, Arora, Basart, Tang, Song, and Steinhardt]{hendrycks2021measuring}
Hendrycks, D., Burns, C., Kadavath, S., Arora, A., Basart, S., Tang, E., Song, D., and Steinhardt, J.
\newblock Measuring mathematical problem solving with the math dataset.
\newblock \emph{arXiv preprint arXiv:2103.03874}, 2021.

\bibitem[Hinton et~al.(1992)]{hinton1992neural}
Hinton, G.~E. et~al.
\newblock \emph{How neural networks learn from experience}.
\newblock na, 1992.

\bibitem[Hu et~al.(2024)Hu, Lu, and Clune]{hu2024ADAS}
Hu, S., Lu, C., and Clune, J.
\newblock Automated design of agentic systems.
\newblock \emph{arXiv preprint arXiv:2408.08435}, 2024.

\bibitem[Jiang et~al.(2023{\natexlab{a}})Jiang, Sablayrolles, Mensch, Bamford, Chaplot, de~las Casas, Bressand, Lengyel, Lample, Saulnier, Lavaud, Lachaux, Stock, Scao, Lavril, Wang, Lacroix, and Sayed]{jiang2023mistral}
Jiang, A.~Q., Sablayrolles, A., Mensch, A., Bamford, C., Chaplot, D.~S., de~las Casas, D., Bressand, F., Lengyel, G., Lample, G., Saulnier, L., Lavaud, L.~R., Lachaux, M.-A., Stock, P., Scao, T.~L., Lavril, T., Wang, T., Lacroix, T., and Sayed, W.~E.
\newblock Mistral 7b, 2023{\natexlab{a}}.

\bibitem[Jiang et~al.(2024)Jiang, Sablayrolles, Roux, Mensch, Savary, Bamford, Chaplot, de~las Casas, Hanna, Bressand, Lengyel, Bour, Lample, Lavaud, Saulnier, Lachaux, Stock, Subramanian, Yang, Antoniak, Scao, Gervet, Lavril, Wang, Lacroix, and Sayed]{jiang2024mixtral}
Jiang, A.~Q., Sablayrolles, A., Roux, A., Mensch, A., Savary, B., Bamford, C., Chaplot, D.~S., de~las Casas, D., Hanna, E.~B., Bressand, F., Lengyel, G., Bour, G., Lample, G., Lavaud, L.~R., Saulnier, L., Lachaux, M.-A., Stock, P., Subramanian, S., Yang, S., Antoniak, S., Scao, T.~L., Gervet, T., Lavril, T., Wang, T., Lacroix, T., and Sayed, W.~E.
\newblock Mixtral of experts, 2024.

\bibitem[Jiang et~al.(2023{\natexlab{b}})Jiang, Ren, and Lin]{llm-blender-2023}
Jiang, D., Ren, X., and Lin, B.~Y.
\newblock Llm-blender: Ensembling large language models with pairwise comparison and generative fusion.
\newblock In \emph{Proceedings of the 61th Annual Meeting of the Association for Computational Linguistics (ACL 2023)}, 2023{\natexlab{b}}.

\bibitem[Khattab et~al.(2023)Khattab, Singhvi, Maheshwari, Zhang, Santhanam, Vardhamanan, Haq, Sharma, Joshi, Moazam, Miller, Zaharia, and Potts]{khattab2023dspy}
Khattab, O., Singhvi, A., Maheshwari, P., Zhang, Z., Santhanam, K., Vardhamanan, S., Haq, S., Sharma, A., Joshi, T.~T., Moazam, H., Miller, H., Zaharia, M., and Potts, C.
\newblock Dspy: Compiling declarative language model calls into self-improving pipelines.
\newblock \emph{arXiv preprint arXiv:2310.03714}, 2023.

\bibitem[Li et~al.(2024{\natexlab{a}})Li, Zhang, Yu, Fu, and Ye]{li2024agentsneed}
Li, J., Zhang, Q., Yu, Y., Fu, Q., and Ye, D.
\newblock More agents is all you need, 2024{\natexlab{a}}.
\newblock URL \url{https://arxiv.org/abs/2402.05120}.

\bibitem[Li et~al.(2024{\natexlab{b}})Li, Chiang, Frick, Dunlap, Banghua~Zhu, and Stoica]{arenahard2024}
Li, T., Chiang, W.-L., Frick, E., Dunlap, L., Banghua~Zhu, J. E.~G., and Stoica, I.
\newblock From live data to high-quality benchmarks: The arena-hard pipeline, April 2024{\natexlab{b}}.
\newblock URL \url{https://lmsys.org/blog/2024-04-19-arena-hard/}.

\bibitem[Li et~al.(2023)Li, Zhang, Dubois, Taori, Gulrajani, Guestrin, Liang, and Hashimoto]{alpaca_eval}
Li, X., Zhang, T., Dubois, Y., Taori, R., Gulrajani, I., Guestrin, C., Liang, P., and Hashimoto, T.~B.
\newblock Alpacaeval: An automatic evaluator of instruction-following models.
\newblock \url{https://github.com/tatsu-lab/alpaca_eval}, 2023.

\bibitem[Li et~al.(2022)Li, Choi, Chung, Kushman, Schrittwieser, Leblond, Eccles, Keeling, Gimeno, Dal~Lago, et~al.]{li2022competition}
Li, Y., Choi, D., Chung, J., Kushman, N., Schrittwieser, J., Leblond, R., Eccles, T., Keeling, J., Gimeno, F., Dal~Lago, A., et~al.
\newblock Competition-level code generation with alphacode.
\newblock \emph{Science}, 378\penalty0 (6624):\penalty0 1092--1097, 2022.

\bibitem[Meng et~al.(2024)Meng, Xia, and Chen]{meng2024simpo}
Meng, Y., Xia, M., and Chen, D.
\newblock {SimPO}: Simple preference optimization with a reference-free reward.
\newblock \emph{ArXiv}, 2024.

\bibitem[Mialon et~al.(2023)Mialon, Fourrier, Swift, Wolf, LeCun, and Scialom]{mialon2023gaiabenchmarkgeneralai}
Mialon, G., Fourrier, C., Swift, C., Wolf, T., LeCun, Y., and Scialom, T.
\newblock Gaia: a benchmark for general ai assistants, 2023.
\newblock URL \url{https://arxiv.org/abs/2311.12983}.

\bibitem[Nardi et~al.(2019)Nardi, Souza, Koeplinger, and Olukotun]{9124618}
Nardi, L., Souza, A., Koeplinger, D., and Olukotun, K.
\newblock Hypermapper: a practical design space exploration framework.
\newblock In \emph{2019 IEEE 27th International Symposium on Modeling, Analysis, and Simulation of Computer and Telecommunication Systems (MASCOTS)}, pp.\  425--426, 2019.
\newblock \doi{10.1109/MASCOTS.2019.00053}.

\bibitem[Ni et~al.(2024)Ni, Xue, Yue, Deng, Shah, Jain, Neubig, and You]{ni2024mixevalderivingwisdomcrowd}
Ni, J., Xue, F., Yue, X., Deng, Y., Shah, M., Jain, K., Neubig, G., and You, Y.
\newblock Mixeval: Deriving wisdom of the crowd from llm benchmark mixtures, 2024.
\newblock URL \url{https://arxiv.org/abs/2406.06565}.

\bibitem[{OpenAI}(2024)]{openai2024o1}
{OpenAI}.
\newblock Learning to reason with {LLMs}.
\newblock \url{https://openai.com/research/learning-to-reason-with-llms}, September 2024.
\newblock Accessed November 13, 2024.

\bibitem[OpenAI et~al.(2024)OpenAI, Achiam, Adler, Agarwal, Ahmad, Akkaya, Aleman, Almeida, Altenschmidt, Altman, Anadkat, Avila, Babuschkin, Balaji, Balcom, Baltescu, Bao, Bavarian, Belgum, Bello, Berdine, Bernadett-Shapiro, Berner, Bogdonoff, Boiko, Boyd, Brakman, Brockman, Brooks, Brundage, Button, Cai, Campbell, Cann, Carey, Carlson, Carmichael, Chan, Chang, Chantzis, Chen, Chen, Chen, Chen, Chen, Chess, Cho, Chu, Chung, Cummings, Currier, Dai, Decareaux, Degry, Deutsch, Deville, Dhar, Dohan, Dowling, Dunning, Ecoffet, Eleti, Eloundou, Farhi, Fedus, Felix, Fishman, Forte, Fulford, Gao, Georges, Gibson, Goel, Gogineni, Goh, Gontijo-Lopes, Gordon, Grafstein, Gray, Greene, Gross, Gu, Guo, Hallacy, Han, Harris, He, Heaton, Heidecke, Hesse, Hickey, Hickey, Hoeschele, Houghton, Hsu, Hu, Hu, Huizinga, Jain, Jain, Jang, Jiang, Jiang, Jin, Jin, Jomoto, Jonn, Jun, Kaftan, Łukasz Kaiser, Kamali, Kanitscheider, Keskar, Khan, Kilpatrick, Kim, Kim, Kim, Kirchner, Kiros, Knight, Kokotajlo, Łukasz Kondraciuk, Kondrich,
  Konstantinidis, Kosic, Krueger, Kuo, Lampe, Lan, Lee, Leike, Leung, Levy, Li, Lim, Lin, Lin, Litwin, Lopez, Lowe, Lue, Makanju, Malfacini, Manning, Markov, Markovski, Martin, Mayer, Mayne, McGrew, McKinney, McLeavey, McMillan, McNeil, Medina, Mehta, Menick, Metz, Mishchenko, Mishkin, Monaco, Morikawa, Mossing, Mu, Murati, Murk, Mély, Nair, Nakano, Nayak, Neelakantan, Ngo, Noh, Ouyang, O'Keefe, Pachocki, Paino, Palermo, Pantuliano, Parascandolo, Parish, Parparita, Passos, Pavlov, Peng, Perelman, de~Avila Belbute~Peres, Petrov, de~Oliveira~Pinto, Michael, Pokorny, Pokrass, Pong, Powell, Power, Power, Proehl, Puri, Radford, Rae, Ramesh, Raymond, Real, Rimbach, Ross, Rotsted, Roussez, Ryder, Saltarelli, Sanders, Santurkar, Sastry, Schmidt, Schnurr, Schulman, Selsam, Sheppard, Sherbakov, Shieh, Shoker, Shyam, Sidor, Sigler, Simens, Sitkin, Slama, Sohl, Sokolowsky, Song, Staudacher, Such, Summers, Sutskever, Tang, Tezak, Thompson, Tillet, Tootoonchian, Tseng, Tuggle, Turley, Tworek, Uribe, Vallone, Vijayvergiya,
  Voss, Wainwright, Wang, Wang, Wang, Ward, Wei, Weinmann, Welihinda, Welinder, Weng, Weng, Wiethoff, Willner, Winter, Wolrich, Wong, Workman, Wu, Wu, Wu, Xiao, Xu, Yoo, Yu, Yuan, Zaremba, Zellers, Zhang, Zhang, Zhao, Zheng, Zhuang, Zhuk, and Zoph]{openai2024gpt4technicalreport}
OpenAI, Achiam, J., Adler, S., Agarwal, S., Ahmad, L., Akkaya, I., Aleman, F.~L., Almeida, D., Altenschmidt, J., Altman, S., Anadkat, S., Avila, R., Babuschkin, I., Balaji, S., Balcom, V., Baltescu, P., Bao, H., Bavarian, M., Belgum, J., Bello, I., Berdine, J., Bernadett-Shapiro, G., Berner, C., Bogdonoff, L., Boiko, O., Boyd, M., Brakman, A.-L., Brockman, G., Brooks, T., Brundage, M., Button, K., Cai, T., Campbell, R., Cann, A., Carey, B., Carlson, C., Carmichael, R., Chan, B., Chang, C., Chantzis, F., Chen, D., Chen, S., Chen, R., Chen, J., Chen, M., Chess, B., Cho, C., Chu, C., Chung, H.~W., Cummings, D., Currier, J., Dai, Y., Decareaux, C., Degry, T., Deutsch, N., Deville, D., Dhar, A., Dohan, D., Dowling, S., Dunning, S., Ecoffet, A., Eleti, A., Eloundou, T., Farhi, D., Fedus, L., Felix, N., Fishman, S.~P., Forte, J., Fulford, I., Gao, L., Georges, E., Gibson, C., Goel, V., Gogineni, T., Goh, G., Gontijo-Lopes, R., Gordon, J., Grafstein, M., Gray, S., Greene, R., Gross, J., Gu, S.~S., Guo, Y., Hallacy,
  C., Han, J., Harris, J., He, Y., Heaton, M., Heidecke, J., Hesse, C., Hickey, A., Hickey, W., Hoeschele, P., Houghton, B., Hsu, K., Hu, S., Hu, X., Huizinga, J., Jain, S., Jain, S., Jang, J., Jiang, A., Jiang, R., Jin, H., Jin, D., Jomoto, S., Jonn, B., Jun, H., Kaftan, T., Łukasz Kaiser, Kamali, A., Kanitscheider, I., Keskar, N.~S., Khan, T., Kilpatrick, L., Kim, J.~W., Kim, C., Kim, Y., Kirchner, J.~H., Kiros, J., Knight, M., Kokotajlo, D., Łukasz Kondraciuk, Kondrich, A., Konstantinidis, A., Kosic, K., Krueger, G., Kuo, V., Lampe, M., Lan, I., Lee, T., Leike, J., Leung, J., Levy, D., Li, C.~M., Lim, R., Lin, M., Lin, S., Litwin, M., Lopez, T., Lowe, R., Lue, P., Makanju, A., Malfacini, K., Manning, S., Markov, T., Markovski, Y., Martin, B., Mayer, K., Mayne, A., McGrew, B., McKinney, S.~M., McLeavey, C., McMillan, P., McNeil, J., Medina, D., Mehta, A., Menick, J., Metz, L., Mishchenko, A., Mishkin, P., Monaco, V., Morikawa, E., Mossing, D., Mu, T., Murati, M., Murk, O., Mély, D., Nair, A., Nakano, R.,
  Nayak, R., Neelakantan, A., Ngo, R., Noh, H., Ouyang, L., O'Keefe, C., Pachocki, J., Paino, A., Palermo, J., Pantuliano, A., Parascandolo, G., Parish, J., Parparita, E., Passos, A., Pavlov, M., Peng, A., Perelman, A., de~Avila Belbute~Peres, F., Petrov, M., de~Oliveira~Pinto, H.~P., Michael, Pokorny, Pokrass, M., Pong, V.~H., Powell, T., Power, A., Power, B., Proehl, E., Puri, R., Radford, A., Rae, J., Ramesh, A., Raymond, C., Real, F., Rimbach, K., Ross, C., Rotsted, B., Roussez, H., Ryder, N., Saltarelli, M., Sanders, T., Santurkar, S., Sastry, G., Schmidt, H., Schnurr, D., Schulman, J., Selsam, D., Sheppard, K., Sherbakov, T., Shieh, J., Shoker, S., Shyam, P., Sidor, S., Sigler, E., Simens, M., Sitkin, J., Slama, K., Sohl, I., Sokolowsky, B., Song, Y., Staudacher, N., Such, F.~P., Summers, N., Sutskever, I., Tang, J., Tezak, N., Thompson, M.~B., Tillet, P., Tootoonchian, A., Tseng, E., Tuggle, P., Turley, N., Tworek, J., Uribe, J. F.~C., Vallone, A., Vijayvergiya, A., Voss, C., Wainwright, C., Wang,
  J.~J., Wang, A., Wang, B., Ward, J., Wei, J., Weinmann, C., Welihinda, A., Welinder, P., Weng, J., Weng, L., Wiethoff, M., Willner, D., Winter, C., Wolrich, S., Wong, H., Workman, L., Wu, S., Wu, J., Wu, M., Xiao, K., Xu, T., Yoo, S., Yu, K., Yuan, Q., Zaremba, W., Zellers, R., Zhang, C., Zhang, M., Zhao, S., Zheng, T., Zhuang, J., Zhuk, W., and Zoph, B.
\newblock Gpt-4 technical report, 2024.
\newblock URL \url{https://arxiv.org/abs/2303.08774}.

\bibitem[Qwen(2024)]{qwen2}
Qwen.
\newblock Qwen2 technical report.
\newblock 2024.

\bibitem[Rein et~al.(2023)Rein, Hou, Stickland, Petty, Pang, Dirani, Michael, and Bowman]{rein2023gpqagraduatelevelgoogleproofqa}
Rein, D., Hou, B.~L., Stickland, A.~C., Petty, J., Pang, R.~Y., Dirani, J., Michael, J., and Bowman, S.~R.
\newblock Gpqa: A graduate-level google-proof qa benchmark, 2023.
\newblock URL \url{https://arxiv.org/abs/2311.12022}.

\bibitem[Rein et~al.(2024)Rein, Hou, Stickland, Petty, Pang, Dirani, Michael, and Bowman]{rein2024gpqa}
Rein, D., Hou, B.~L., Stickland, A.~C., Petty, J., Pang, R.~Y., Dirani, J., Michael, J., and Bowman, S.~R.
\newblock {GPQA}: A graduate-level google-proof qa benchmark.
\newblock In \emph{First Conference on Language Modeling}, 2024.
\newblock URL \url{https://openreview.net/forum?id=Ti67584b98}.

\bibitem[Ren et~al.(2021)Ren, Xiao, Chang, Huang, Li, Chen, and Wang]{ren2021comprehensive}
Ren, P., Xiao, Y., Chang, X., Huang, P.-Y., Li, Z., Chen, X., and Wang, X.
\newblock A comprehensive survey of neural architecture search: Challenges and solutions.
\newblock \emph{ACM Computing Surveys (CSUR)}, 54\penalty0 (4):\penalty0 1--34, 2021.

\bibitem[Snoek et~al.(2012)Snoek, Larochelle, and Adams]{snoek2012practicalbayesianoptimizationmachine}
Snoek, J., Larochelle, H., and Adams, R.~P.
\newblock Practical bayesian optimization of machine learning algorithms, 2012.
\newblock URL \url{https://arxiv.org/abs/1206.2944}.

\bibitem[Sreenivas et~al.(2024)Sreenivas, Muralidharan, Joshi, Chochowski, Patwary, Shoeybi, Catanzaro, Kautz, and Molchanov]{sreenivas2024llmpruningdistillationpractice}
Sreenivas, S.~T., Muralidharan, S., Joshi, R., Chochowski, M., Patwary, M., Shoeybi, M., Catanzaro, B., Kautz, J., and Molchanov, P.
\newblock Llm pruning and distillation in practice: The minitron approach, 2024.
\newblock URL \url{https://arxiv.org/abs/2408.11796}.

\bibitem[Team(2025)]{sky_t1_2025}
Team, N.
\newblock Sky-t1: Train your own o1 preview model within 450.
\newblock https://novasky-ai.github.io/posts/sky-t1, 2025.
\newblock Accessed: 2025-01-09.

\bibitem[Team(2024)]{qwq-32b-preview}
Team, Q.
\newblock Qwq: Reflect deeply on the boundaries of the unknown, November 2024.
\newblock URL \url{https://qwenlm.github.io/blog/qwq-32b-preview/}.

\bibitem[Tran et~al.(2023)Tran, Glaze, and Hancock]{viethoangtranduong}
Tran, H., Glaze, C., and Hancock, B.
\newblock Iterative dpo alignment.
\newblock Technical report, Snorkel AI, 2023.

\bibitem[Tunstall et~al.(2023)Tunstall, Beeching, Lambert, Rajani, Rasul, Belkada, Huang, von Werra, Fourrier, Habib, et~al.]{tunstall2023zephyr}
Tunstall, L., Beeching, E., Lambert, N., Rajani, N., Rasul, K., Belkada, Y., Huang, S., von Werra, L., Fourrier, C., Habib, N., et~al.
\newblock Zephyr: Direct distillation of lm alignment.
\newblock \emph{arXiv preprint arXiv:2310.16944}, 2023.

\bibitem[Wang et~al.(2024{\natexlab{a}})Wang, Wang, Athiwaratkun, Zhang, and Zou]{wang2024mixtureofagentsenhanceslargelanguage}
Wang, J., Wang, J., Athiwaratkun, B., Zhang, C., and Zou, J.
\newblock Mixture-of-agents enhances large language model capabilities, 2024{\natexlab{a}}.
\newblock URL \url{https://arxiv.org/abs/2406.04692}.

\bibitem[Wang et~al.(2024{\natexlab{b}})Wang, Ma, Zhang, Ni, Chandra, Guo, Ren, Arulraj, He, Jiang, et~al.]{wang2024mmlu}
Wang, Y., Ma, X., Zhang, G., Ni, Y., Chandra, A., Guo, S., Ren, W., Arulraj, A., He, X., Jiang, Z., et~al.
\newblock Mmlu-pro: A more robust and challenging multi-task language understanding benchmark.
\newblock \emph{arXiv preprint arXiv:2406.01574}, 2024{\natexlab{b}}.

\bibitem[Xu et~al.(2024)Xu, Sun, Zheng, Geng, Zhao, Feng, Tao, Lin, and Jiang]{xu2024wizardlm}
Xu, C., Sun, Q., Zheng, K., Geng, X., Zhao, P., Feng, J., Tao, C., Lin, Q., and Jiang, D.
\newblock Wizard{LM}: Empowering large pre-trained language models to follow complex instructions.
\newblock In \emph{The Twelfth International Conference on Learning Representations}, 2024.
\newblock URL \url{https://openreview.net/forum?id=CfXh93NDgH}.

\bibitem[Yuksekgonul et~al.(2024)Yuksekgonul, Bianchi, Boen, Liu, Huang, Guestrin, and Zou]{yuksekgonul2024textgrad}
Yuksekgonul, M., Bianchi, F., Boen, J., Liu, S., Huang, Z., Guestrin, C., and Zou, J.
\newblock Textgrad: Automatic "differentiation" via text.
\newblock 2024.

\bibitem[Zhang et~al.(2024)Zhang, Xiang, Yu, Teng, Chen, Chen, Zhuge, Cheng, Hong, Wang, Zheng, Liu, Luo, and Wu]{zhang2024aflowautomatingagenticworkflow}
Zhang, J., Xiang, J., Yu, Z., Teng, F., Chen, X., Chen, J., Zhuge, M., Cheng, X., Hong, S., Wang, J., Zheng, B., Liu, B., Luo, Y., and Wu, C.
\newblock Aflow: Automating agentic workflow generation, 2024.
\newblock URL \url{https://arxiv.org/abs/2410.10762}.

\bibitem[Zheng et~al.(2023)Zheng, Chiang, Sheng, Zhuang, Wu, Zhuang, Lin, Li, Li, Xing, Zhang, Gonzalez, and Stoica]{zheng2023judging}
Zheng, L., Chiang, W.-L., Sheng, Y., Zhuang, S., Wu, Z., Zhuang, Y., Lin, Z., Li, Z., Li, D., Xing, E.~P., Zhang, H., Gonzalez, J.~E., and Stoica, I.
\newblock Judging llm-as-a-judge with mt-bench and chatbot arena, 2023.

\bibitem[Zoph \& Le(2017)Zoph and Le]{zoph2017neuralarchitecturesearchreinforcement}
Zoph, B. and Le, Q.~V.
\newblock Neural architecture search with reinforcement learning, 2017.
\newblock URL \url{https://arxiv.org/abs/1611.01578}.

\end{thebibliography}
\bibliographystyle{icml2025}

\newpage
\appendix
\onecolumn

\clearpage
\appendix
\section{Appendix}

\subsection{Table of Contents}

\begin{enumerate}
\item {\textbf{\archon{} LLM Components}: Outline of LM components used in \archon{} and the rules for combining them} 
\item {\textbf{Utilities and Interactions of LLM Components}: Analysis of the effectiveness of individual components and their synergistic effects when combined}
\item {\textbf{Bayesian Optimization for \archon{}}: Description of search space and optimization methodology for architecture discovery}
\item {\textbf{Bayes Optimization vs. Alternative Approaches}: Comparative analysis of search techniques for finding optimal configurations}
\item {\textbf{\archon{} Architecture Algorithms Comparisons}: Evaluation of different optimization strategies across various inference budgets}
\item {\textbf{\archon{} Benchmarks and Results}: Comprehensive evaluation results across instruction-following, reasoning, and coding tasks}
\item {\textbf{\archon{} LLM Analysis}: Details of models tested with parameter counts and sequence lengths}
\item {\textbf{\archon{} Architectures}: Diagrams and descriptions of optimized architectures for different task categories}
\item {\textbf{\archon{} by Inference Compute Budget, Model Size, and Cost}: Analysis of performance across different compute constraints and model sizes}
\end{enumerate}

\subsection{\archon{} LLM Components}
\label{sec:lm_construction_and_prompts}

\begin{table}[H]
\centering
\scriptsize
\setlength{\tabcolsep}{2pt}
\begin{tabular}{clcccc}
\toprule
\textbf{\begin{tabular}[c]{@{}c@{}}Inference-Time\\ Technique\end{tabular}} & \multicolumn{1}{c}{\textbf{Definition}}                                                              & \textbf{Input}                                                                     & \textbf{Output}                                                                       & \textbf{\begin{tabular}[c]{@{}c@{}}Inference\\ Cost\end{tabular}} & \textbf{\begin{tabular}[c]{@{}c@{}}Domains\end{tabular}} \\ \midrule
Generator                                                       & \begin{tabular}[c]{@{}l@{}}Generates a candidate response \\ from an instruction prompt\end{tabular} & Instruction Prompt                                                                 & Candidate Response(s)                                                                 & \begin{tabular}[c]{@{}c@{}}1 call \\ per cand.\end{tabular}      & \begin{tabular}[c]{@{}c@{}}All \\ Domains\end{tabular}                                                        \\ \midrule
Fuser                                                           & \begin{tabular}[c]{@{}l@{}}Merges multiple candidate\\ responses into a single response\end{tabular}   & \begin{tabular}[c]{@{}c@{}}Instruction Prompt +\\ Candidate Response(s)\end{tabular} & \begin{tabular}[c]{@{}c@{}}Fused Candidate\\ Response(s)\end{tabular}                 & \begin{tabular}[c]{@{}c@{}}1 call\\ per cand.\end{tabular}       & \begin{tabular}[c]{@{}c@{}}All \\ Domains\end{tabular}                                                        \\ \midrule
Critic                                                          & \begin{tabular}[c]{@{}l@{}}Generates strengths/weaknesses\\ for each candidate response\end{tabular} & \begin{tabular}[c]{@{}c@{}}Instruction Prompt +\\ Candidate Response(s)\end{tabular} & \begin{tabular}[c]{@{}c@{}}Candidate Response(s) \\ Strengths/Weaknesses\end{tabular} & 1 call                                                            & \begin{tabular}[c]{@{}c@{}}All \\ Domains\end{tabular}                                                        \\ \midrule
Ranker                                                          & \begin{tabular}[c]{@{}l@{}}Returns top-K \\ candidate responses\end{tabular}                         & \begin{tabular}[c]{@{}c@{}}Instruction Prompt +\\ Candidate Response(s)\end{tabular} & \begin{tabular}[c]{@{}c@{}}Ranked Candidate \\ Response(s)\end{tabular}               & 1 call                                                            & \begin{tabular}[c]{@{}c@{}}All \\ Domains\end{tabular}                                                        \\ \midrule
Verifier                                                        & \begin{tabular}[c]{@{}l@{}}Returns the candidate responses \\ with verified reasoning\end{tabular}   & \begin{tabular}[c]{@{}c@{}}Instruction Prompt +\\ Candidate Response(s)\end{tabular} & \begin{tabular}[c]{@{}c@{}}Verified Candidate \\ Response(s)\end{tabular}             & \begin{tabular}[c]{@{}c@{}}2 calls\\per cand.\end{tabular}   & \begin{tabular}[c]{@{}c@{}}Reasoning \\ Tasks\end{tabular}         \\ \midrule
\begin{tabular}[c]{@{}c@{}}Unit Test \\ Generator\end{tabular}  & \begin{tabular}[c]{@{}l@{}}Generates unit tests to evaluate \\ the candidate responses\end{tabular}    & Instruction Prompt                                                                 & \begin{tabular}[c]{@{}c@{}}Instruction Prompt\\ + Unit Tests\end{tabular}   & 1 call                                                            & \begin{tabular}[c]{@{}c@{}}Reasoning\\ Tasks\end{tabular} \\ \midrule
\begin{tabular}[c]{@{}c@{}}Unit Test \\ Evaluator\end{tabular}  & \begin{tabular}[c]{@{}l@{}}Uses generated unit tests to\\evaluate candidate response\end{tabular}    & \begin{tabular}[c]{@{}c@{}}Instruction Prompt +\\ Unit Tests +\\Candidate Response(s)\end{tabular}                                                                & \begin{tabular}[c]{@{}c@{}}Scored Candidate\\Response(s)\end{tabular}   & \begin{tabular}[c]{@{}c@{}}1 call\\per cand.\end{tabular}                                                            & \begin{tabular}[c]{@{}c@{}}Reasoning\\ Tasks\end{tabular} \\ \bottomrule         
\end{tabular}
\caption{\textbf{Overview of \archon{}'s Inference-time Techniques}:  Definitions, Inputs, Outputs, Costs, and Application Domains.
}
\label{tab:archon_layers_outputs_figure}
\end{table}

\begin{table}[H]
\centering
\scriptsize
\begin{tabular}{cccccc}
\toprule
\textbf{Module}                                                & \textbf{\begin{tabular}[c]{@{}c@{}}Initial Layer\\ Placement\end{tabular}} & \textbf{\begin{tabular}[c]{@{}c@{}}Placement after\\ Initial Layer\end{tabular}} & \textbf{\begin{tabular}[c]{@{}c@{}}\textgreater{}1 Module\\ in Layer\end{tabular}} & \textbf{\begin{tabular}[c]{@{}c@{}}Increase\\ Candidate\\ Responses\end{tabular}} & \textbf{\begin{tabular}[c]{@{}c@{}}Decrease\\ Candidate\\ Responses\end{tabular}} \\ \midrule
Generator                                                      & Yes                                                               & No                                                                      & Yes                                                                       & Yes                                                                      & No                                                                       \\ \midrule
Fuser                                                          & No                                                                & Yes                                                                     & Yes                                                                       & Yes                                                                      & Yes                                                                      \\ \midrule
Ranker                                                         & No                                                                & Yes                                                                     & No                                                                        & No                                                                       & Yes                                                                      \\ \midrule
Critic                                                         & No                                                                & Yes                                                                     & No                                                                        & No                                                                       & No                                                                       \\ \midrule
Verifier                                                       & No                                                                & Yes                                                                     & No                                                                        & No                                                                       & Yes                                                                      \\ \midrule
\begin{tabular}[c]{@{}c@{}}Unit Test \\ Generator\end{tabular} & No                                                              & Yes                                                                      & No                                                                        & No                                                                       & No  \\ \midrule
\begin{tabular}[c]{@{}c@{}}Unit Test \\ Evaluator\end{tabular} & No                                                              & Yes                                                                      & No                                                                        & No                                                                       & No  \\ \bottomrule                                                                    
\end{tabular}
\caption{\textbf{Rules of \archon{} Construction}: Allowed combinations of each LLM component from Section \ref{sec:compound_lm_modules}.}
\label{tab:rules_of_construction}
\end{table}

\subsection{Utilities and Interactions of LLM Components}
\label{sec:tradeoffs}

\begin{figure*}[t!]
   \centering
   \includegraphics[width=\linewidth]{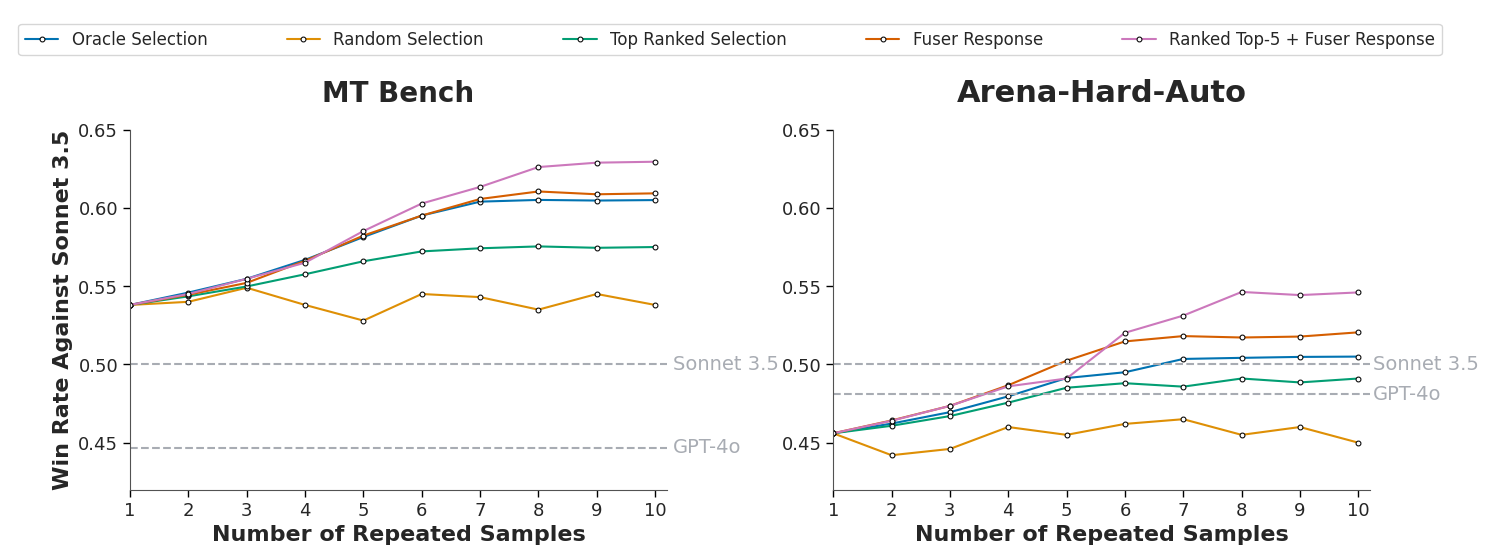}
   \caption{\textbf{Performance Gains from Applying Inference Time Techniques on a Single Model}: 
   We repeatedly sample more responses for each individual query.
   For each sample count, we choose the best response in 5 different ways: 
   \textbf{(1)} using an oracle (to get the upper bound for performance of best sample),
   \textbf{(2)} randomly,
   \textbf{(3)} using a ranker model,  
   \textbf{(4)} by fusion, in which a  model synthesizes a response based on all the samples, and 
   \textbf{(5)} by ranking the top-5 best answers and then fusing them. For both MT Bench and Arena-Hard-Auto, we find that fusion is an effective technique. In particular, ranking the candidates first, and then selecting the top-5 and fusing them scores the highest. The best open-source model for these tasks across all the 70B+ models we are considering is WizardLM-2-8x22B \citep{xu2024wizardlm}  (see \autoref{tab:generator_and_fusion_rankings} for details).
   For both ranking and fusion, we use Qwen2 72B Instruct \citep{qwen2}.}
   \label{fig:sampling_graphs}
\end{figure*}

In this subsection, we present our analysis of the effectiveness of each LLM component (i.e. the \textit{Utility}) and the relationships between each component (i.e. the \textit{Component Interactions}) by evaluating on \textit{instruction-following tasks} (MT Bench, AlpacaEval 2.0, Arena-Hard-Auto), \textit{reasoning tasks} (MixEval, MixEval-Hard, MATH) and \textit{coding tasks} (CodeContests) (Section \ref{sec:benchmarks_and_models}).
For our \archon{} models, we utilize a host of 70B+ open-source models (Section \ref{sec:benchmarks_and_models}; \autoref{tab:models_overview}).

\subsubsection{Generator}
\textbf{Utility}: 
For our Generator module, we find additional model sampling to significantly boost performance (\autoref{fig:sampling_graphs}), particularly for coding tasks (\autoref{tab:archon_main_table_results}).
In settings with a limited inference call budget, additional model samples lead to the largest marginal benefit.
We see a similar pattern for model ensembling, where sampling from additional models leads to continual performance increases (assuming the models are ordered from best to worst for the given task) (\autoref{fig:ensembling_graphs}).

\subsubsection{Fuser} 
\label{sec:fuser_analysis}
\textbf{Utility}: For every benchmark explored, we found that the Fuser module substantially improved performance (\autoref{fig:sampling_graphs}; \autoref{fig:ensembling_graphs}; \autoref{fig:compound_LM_ablations}).
For the single-generation 10-model ensemble of 70B+ models, the Fuser module improved downstream accuracy by 5.2 points, on average, compared to the single-generation best model (\autoref{fig:ensembling_graphs}).
When combined with the Ranker module for ranking the top-5 candidate responses, the Fuser improved downstream accuracy by 7.3 points and 3.6 points, on average, compared to the single-sample best model and the oracle best candidate response, respectively (\autoref{fig:ensembling_graphs}). 
Overall, we found that Fuser efficacy increased as more candidate responses were provided, demonstrating that additional candidate generations can continue to bolster inference-time architecture performance when combined with a Fuser.

In previous work like Mixture-of-Agents (MoA) \citep{wang2024mixtureofagentsenhanceslargelanguage}, multiple layers of Fusers was found to boost performance on some instruction-following tasks (i.e. MT Bench and Alpaca Eval 2.0).
Across all the benchmarks explored, we observed similar benefits in the \archon{} framework when adding multiple layers of Fusers (\autoref{fig:compound_LM_ablations}).
However, based on our results in \autoref{fig:fusion_layer_analysis}, the number of Fuser layers needed to improve performance varied by task, with some tasks receiving limited benefits from added layers (1-2 point increase in accuracy for MixEval) while others experienced significant benefits with 3-4 fusion layers and more (2 to 5 point increase in win rate for MT Bench and Alpaca Eval 2.0).
We attribute this distinction to the difference in task requirements, with chat and instruction following tasks benefiting more from multiple iterations of revisions through the multiple Fuser layers, leading to greater diversity in the final generation (\autoref{tab:jaccard_similarities}).

\begin{figure*}[t!]
   \centering
   \includegraphics[width=\linewidth]{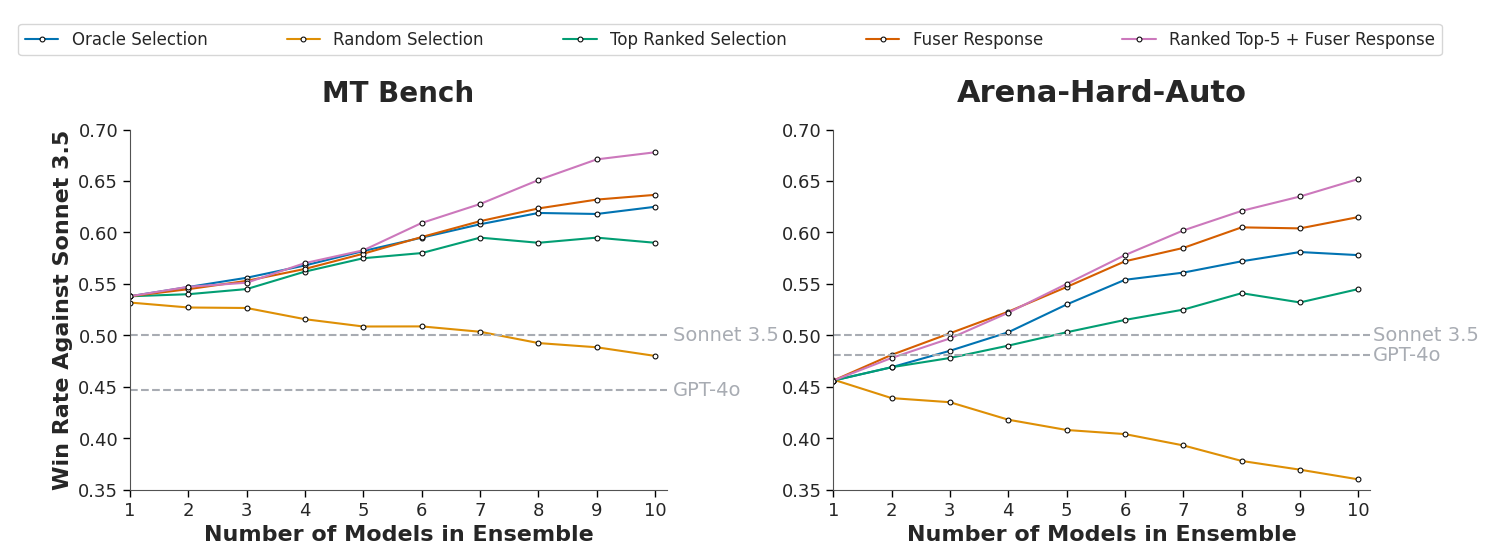}
   \caption{\textbf{Performance Gains from Applying Inference-Time Techniques on an Ensemble of Models}: 
  We incrementally add more models to the ensemble, which consists of open-source 70B+ models. The models are added to the pool based on their performance for each task, from best to worse (see \autoref{tab:generator_and_fusion_rankings} for details).
   For each ensemble size, we choose the best response in 5 different modes: 
   \textbf{(1)} using an oracle (to get the upper bound for performance of best individual response in the ensemble),
   \textbf{(2)} randomly,
   \textbf{(3)} using a ranker model,  
   \textbf{(4)} by fusion, in which one model synthesizes a response based on all the responses of the ensemble models, and 
   \textbf{(5)} ranking the top-5 best responses and then fusing them. For MT Bench and Arena-Hard-Auto, we find consistent performance improvements as we add more models to the ensemble. We find that fusion is beneficial across various ensemble sizes and in particular a fused candidate based on the top-5 ranked responses scores highest. 
 The ensemble approach scores higher than applying the same techniques on repeated samples from a single best-performing model (see \autoref{fig:sampling_graphs}). For both ranking and fusion, we use Qwen2 72B Instruct \citep{qwen2}.}
   \label{fig:ensembling_graphs}
\end{figure*}

\noindent \textbf{Component Interactions}: To better understand how the Fuser module works with the other LLM components, we took the single-sample 10-model ensemble of Generators with a Fuser and tried adding each of these components individually: a Critic, a Ranker, a Verifier, and a Unit Test Generator/Evaluator.
Across all of the benchmarks, the added candidate response analyses from the Critic improved the Fuser's ability to effectively merge the different candidate responses, increasing performance by an average of 3.1 percentage points (\autoref{fig:compound_LM_ablations}).
With the added Ranker, the \archon{} architecture improved the combined Ensemble + Critic + Fuser performance across all the benchmarks by 4.8 percentage points, on average (\autoref{fig:compound_LM_ablations}). 
The Ranker proved most effective for style-oriented tasks (e.g. MT Bench and AlpacaEval 2.0) since the examples mostly focus on improving the instruction-guidance towards the provided prompt.
With the added Verifier module (\autoref{fig:compound_LM_ablations}), the performance of the Ensemble + Critic + Fuser configuration improved marginally for the instruction-following tasks (1.2 percentage points, on average, for MT Bench, AlpacaEval 2.0, and Arena-Hard-Auto).
However, this configuration improved performance more on reasoning tasks (3.2 percentage points for MixEval and MixEval-Hard, on average), assisting generation by filtering out irrelevant or flawed answers before the final fusion step (\autoref{fig:compound_LM_ablations}).
The added Unit Test Generator and Evaluator was less effective for the instruction-following and reasoning tasks, only providing a 1.5 percentage points increase, on average, when added to the Ensemble + Critic + Fuser configuration (\autoref{tab:archon_component_ablations}).
However, for coding tasks, we found unit test generation and evaluation significantly improved performance, leading to a 10.7 percentage point increase (56\% performance increase comparatively) as we scale model sampling (\autoref{tab:archon_main_table_results}).

\subsubsection{Critic}

\textbf{Utility}: The Critic module proved effective for every task we explored in \autoref{fig:compound_LM_ablations} and \autoref{tab:archon_component_ablations}.
With our 10-model 70B+ Generator ensemble and Fuser configuration of \archon{}, the added Critic improved performance on average by 3.1 percentage points across the benchmarks explored.

\noindent \textbf{Component Interactions}: While useful for most \archon{} architectures, the added strengths and weaknesses from the Critic module are particularly useful when combined with the Fuser module, helping guide generation fusion for a single layer and even useful when placed between multiple fusion layers (on average 3.2 percentage point boost across benchmarks in \autoref{fig:compound_LM_ablations}).
The Critic module was also effective with the Ranker module, providing additional information for comparing candidate responses (\autoref{fig:sampling_graphs}) and leading to a 5.9 percentage point increase, on average (\autoref{tab:archon_component_ablations}).

\subsubsection{Ranker} 

\textbf{Utility}: From our results in \autoref{tab:archon_component_ablations}, \autoref{fig:sampling_graphs}, and \autoref{fig:ensembling_graphs}, we found the Ranker to be most effective for instruction-following tasks, where pair-wise comparisons of answers focus on style and adherence to the prompt.
To examine the candidate selection improvement provided by candidate ranking, we compare three approaches to the Ranker: \textbf{(1)} random selection of candidate generation, \textbf{(2)} oracle selection of candidate generation, and \textbf{(3)} the top-ranked candidate selected by our Ranker.
For MT Bench and Arena-Hard-Auto, we find that the ranker improves generation output quality by 3.8\% compared to random candidate selection and performs within 2.7\% of oracle selection (\autoref{fig:sampling_graphs}).

\noindent \textbf{Component Interactions}: Based on our benchmark results in \autoref{tab:archon_component_ablations}, the Ranker pairs well with the Critic module; the provided strengths and weaknesses helps guide ranking, particularly for instruction-following tasks, improving performance by 5.9 percentage points, on average.
Furthermore, the Ranker was also effective when paired with the Fuser; the filtered list of candidate responses helped improve the final condensed response produced by the Fuser by 3.8 percentage points, on average (\autoref{fig:ensembling_graphs}).
When paired with the Verifier and Unit Test Generator, the Ranker had neutral effects; performances changed marginally, either positively or negatively by 1-2 percentage points (\autoref{tab:archon_component_ablations}).

Overall, our findings demonstrate the value of added Rankers for instruction-following and reasoning tasks when paired with Fusers. 
We find that when Rankers are used alone with an ensemble of Generators, their performance lags behind the 10-sample best single model configuration by 3.0 percentage points, on average (\autoref{tab:archon_component_ablations}).
Additionally, our findings show the importance of building better rankers for more complex reasoning tasks, such as math and coding, which is a challenge also raised by \cite{brown2024largelanguagemonkeysscaling}.

\begin{table}[]
\centering
\tiny
\setlength{\tabcolsep}{1.5pt}
\begin{tabular}{clccccccccc}
\toprule
                           & & & \textbf{\begin{tabular}[c]{@{}c@{}}MT\\ Bench\end{tabular}} & \multicolumn{2}{c}{\begin{tabular}[c]{@{}c@{}}\textbf{AlpacaEval}\\\textbf{2.0}\end{tabular}} & \textbf{\begin{tabular}[c]{@{}c@{}}Arena\\ Hard Auto\end{tabular}} & \textbf{\begin{tabular}[c]{@{}c@{}}MixEval\\ Hard\end{tabular}} & \textbf{MixEval} & \textbf{MATH} & \textbf{\begin{tabular}[c]{@{}c@{}}Code\\Contests\end{tabular}} \\
                           \cmidrule(l{7pt}r{7pt}){1-11}
                           & \multicolumn{1}{c}{\textbf{Model / LLM System}} & \begin{tabular}[c]{@{}c@{}}\textbf{\# of}\\ \textbf{Infer.}\\ \textbf{Calls}\end{tabular} & W.R. & \begin{tabular}[c]{@{}c@{}}L.C.\\ W.R.\end{tabular} & \begin{tabular}[c]{@{}c@{}}Raw\\ W.R.\end{tabular} & W.R. & Acc. & Acc. & Acc. & Acc. \\ \midrule
 \multirow{3}{*}{\rotatebox{90}{\begin{tabular}[c]{@{}c@{}} \tiny \textbf{Control}\end{tabular}}} & Best Open-Source 70B+ Model, Sampled Once         & 1 & 55.0\% ±0.4 & 44.7\% ±0.5 & 37.1\% ±0.6 & 45.6\% ±0.5 & 58.7\% ±0.2 & 86.5\% ±0.3 & 84.5\% ±0.6 & {22.5\% ±0.3} \\ 
 & Ensemble \textit{+ Fuser} & 9 & 58.4\% ±0.6 & 57.5\% ±0.4 & 51.3\% ±0.5 & 54.3\% ±0.7 & 60.1\% ±0.5 & 87.3\% ±0.2 & 85.5\% ±0.3 & 23.1\% ±0.7  \\
 & Ensemble \textit{+ Critic + Fuser}        & 10 & 60.9\% ±0.3 & 58.7\% ±0.6 & \underline{65.8\% ±0.3} & 58.8\% ±0.4 & 61.7\% ±0.5 & 87.4\% ±0.3 & \underline{87.2\% ±0.5} & 24.9\% ±0.4 \\ 
 \midrule
 \multirow{8}{*}{\rotatebox{90}{\begin{tabular}[c]{@{}c@{}} \tiny\textbf{Ablations} \end{tabular}}} & Ensemble + \textit{Ranker}  & 9 & 52.5\% ±0.7 & 54.7\% ±0.5 & 47.6\% ±0.4 & 50.5\% ±0.6 & 58.7\% ±0.3 & 86.8\% ±0.4 & 80.4\% ±0.4 & 24.1\% ±0.4 \\
 & Ensemble + \textit{Verifier} & 24 & 53.2\% ±0.5 & 56.2\% ±0.3 & 50.2\% ±0.7 & 52.4\% ±0.3 & 55.9\% ±0.5 & 85.6\% ±0.2 & 85.2\% ±0.7 & 25.3\% ±0.5 \\
  & Ensemble + \textit{Unit Test Gen./Eval.}  & 18 & 51.5\% ±0.4 & 54.4\% ±0.6 & 49.4\% ±0.5 & 46.1\% ±0.8 & 55.2\% ±0.4 & 86.0\% ±0.3 & 85.2\% ±0.5 & 24.6\% ±0.6 \\ 
 & Ensemble + \textit{Ranker + Fuser}        & 10 & \underline{62.5\% ±0.8} & \underline{60.3\% ±0.4} & 63.6\% ±0.6 & 57.2\% ±0.5 & 59.7\% ±0.2 & 87.6\% ±0.3 & 85.3\% ±0.6 & 24.0\% ±0.2 \\
 & Ensemble + \textit{Verifier + Fuser}        & 25 & 60.5\% ±0.3 & 59.4\% ±0.7 & 58.7\% ±0.3 & 59.2\% ±0.4 & \textbf{68.3}\% ±0.3 & 87.5\% ±0.2 & {86.7\% ±0.4} & 26.3\% ±0.6 \\ 
  & Ensemble + \textit{Unit Test Gen./Eval. + Fuser}  & 17 & 61.4\% ±0.6 & 58.5\% ±0.5 & 55.1\% ±0.4 & 56.4\% ±0.7 & 63.9\% ±0.3 & 86.9\% ±0.3 & 86.4\% ±0.8 & \textbf{28.0\% ±0.6} \\
  & Ensemble + \textit{Critic + Verifier + Fuser}        & 25 & 61.3\% ±0.5 & 60.0\% ±0.3 & 61.0\% ±0.7 & \underline{59.5\% ±0.3} & {65.8\% ±0.4} & \underline{87.8\% ±0.4} & {86.1\% ±0.3} & \underline{26.8\% ±0.3} \\
  & Ensemble + \textit{Critic + Ranker + Fuser} & 11 & \textbf{64.7\% ±0.4} & \textbf{62.6\% ±0.6} & \textbf{72.4\% ±0.5} & \textbf{60.9\% ±0.6} & \underline{66.8\% ±0.4} & \textbf{88.3\% ±0.2} & \textbf{87.3\% ±0.5} & 25.5\% ±0.3 \\
 \bottomrule                                            
\end{tabular}
\caption{\textbf{Impact of Different Compositions of \archon{}'s Inference-Time Techniques}:  
We see increased task performances from adding new LLM components to \archon{}. 
For CodeContests, we find that there is a single model (Llama 3.1 405B Instruct) that performs considerably better than the rest of the LLMs studied, making it more effective leverage additional model sampling (\autoref{tab:archon_main_table_results}).
For our ensemble, we use the best 8 open-source 70B+ models for the task (\autoref{tab:generator_and_fusion_rankings}). 
For our fuser, critic, ranker, and verifier components, we use the best fuser model found for the task (\autoref{tab:generator_and_fusion_rankings}).
For each evaluation benchmark, we explain its configuration in \autoref{tab:benchmarks_overview} and Section \ref{sec:benchmarks_and_models}.
{The standard error numbers were calculated from 10 independent evaluation runs.}
}
\label{tab:archon_component_ablations}
\end{table}

\begin{table}[]
\centering
\scriptsize
\setlength{\tabcolsep}{1.5pt}

\begin{tabular}{clcccccccc}
\toprule
                           & & & \textbf{\begin{tabular}[c]{@{}c@{}}MT\\ Bench\end{tabular}} & \begin{tabular}[c]{@{}c@{}}\textbf{AlpacaEval}\\\textbf{2.0}\end{tabular} & \textbf{\begin{tabular}[c]{@{}c@{}}Arena\\ Hard Auto\end{tabular}} & \textbf{\begin{tabular}[c]{@{}c@{}}MixEval\\ Hard\end{tabular}} & \textbf{MixEval} & \textbf{MATH} & \textbf{\begin{tabular}[c]{@{}c@{}}Code\\Contests\end{tabular}} \\
                           \cmidrule(l{7pt}r{7pt}){1-10}
                           & \multicolumn{1}{c}{\textbf{Model / LLM System}} & \begin{tabular}[c]{@{}c@{}}\textbf{\# of}\\ \textbf{Infer.}\\ \textbf{Calls}\end{tabular} & W.R. & \begin{tabular}[c]{@{}c@{}}L.C.\\ W.R.\end{tabular} & W.R. & Acc. & Acc. & Acc. & Acc. \\ \midrule
 \multirow{3}{*}{\rotatebox{90}{\begin{tabular}[c]{@{}c@{}} \tiny \textbf{Control}\end{tabular}}} & Single Generation & 1 & 44.2\% ±0.6 & {57.8\% ±0.5} & 48.1\% ±0.7 & {63.4\% ±0.3} & \underline{87.5\% ±0.2} & 82.1\% ±0.4 & 17.9\% ±0.3 \\
 & Ensemble \textit{+ Fuser}         & 11 & 53.7\% ±0.3 & 59.5\% ±0.6 & 49.7\% ±0.5 & 65.5\% ±0.2 & 82.0\% ±0.3 & 81.0\% ±0.6 & 16.0\% ±0.4 \\ 
 & Ensemble \textit{+ Critic + Fuser}        & 12 & 56.1\% ±0.7 & 59.7\% ±0.4 & 53.9\% ±0.6 & 67.4\% ±0.4 & 82.0\% ±0.2 & 82.3\% ±0.5 & 18.9\% ±0.6 \\ 
 \midrule
 \multirow{8}{*}{\rotatebox{90}{\begin{tabular}[c]{@{}c@{}} \tiny\textbf{Ablations} \end{tabular}}} & Ensemble + \textit{Ranker}  & 11 & 47.6\% ±0.4 & 49.7\% ±0.5 & 45.5\% ±0.4 & 63.3\% ±0.3 & 81.6\% ±0.4 & 77.3\% ±0.7 & 17.9\% ±0.5 \\
 & Ensemble + \textit{Verifier} & 11 & 48.4\% ±0.5 & 51.2\% ±0.7 & 47.7\% ±0.8 & 61.4\% ±0.2 & 80.5\% ±0.3 & 75.5\% ±0.3 & 23.0\% ±0.4 \\
 & Ensemble + \textit{Unit Test Gen./Eval.}  & 21 & 46.8\% ±0.8 & 49.3\% ±0.3 & 41.2\% ±0.5 & 60.2\% ±0.4 & 80.7\% ±0.2 & 78.9\% ±0.8 & \underline{24.0\% ±0.7} \\ 
 & Ensemble + \textit{Ranker + Fuser}        & 12 & \underline{58.0\% ±0.2} & 60.1\% ±0.6 & 52.2\% ±0.3 & 65.0\% ±0.3 & 82.0\% ±0.4 & 82.1\% ±0.4 & 18.0\% ±0.3 \\
 & Ensemble + \textit{Verifier + Fuser}        & 12 & 55.8\% ±0.6 & 54.2\% ±0.4 & 60.3\% ±0.7 & 67.0\% ±0.2 & 82.5\% ±0.3 & 83.1\% ±0.6 & 22.4\% ±0.5 \\ 
 & Ensemble + \textit{Unit Test Gen./Eval. + Fuser}  & 22 & 56.5\% ±0.3 & 61.4\% ±0.5 & 51.6\% ±0.4 & 67.7\% ±0.4 & 81.7\% ±0.2 & 84.3\% ±0.5 & \textbf{25.4\% ±0.6} \\
 & Ensemble + \textit{Critic + Verifier + Fuser}        & 13 & 56.6\% ±0.7 & \underline{62.0\% ±0.3} & \underline{55.0\% ±0.6} & \underline{68.5\% ±0.3} & {82.7\% ±0.4} & \underline{85.7\% ±0.3} & 22.2\% ±0.4 \\
 & Ensemble + \textit{Critic + Ranker + Fuser} & 13 & \textbf{60.0\% ±0.4} & \textbf{62.8\% ±0.6} & \textbf{56.2\% ±0.5} & \textbf{69.4\% ±0.2} & \textbf{88.5\% ±0.3} & \textbf{87.0\% ±0.7} & 18.5\% ±0.5 \\
 \bottomrule                                            
\end{tabular}
\caption{ \textbf{\archon{} Component Compositions with GPT-4o}: The ensemble uses generates 10 samples for the given query.
The standard error numbers were calculated from 10 independent evaluation runs.
}
\label{tab:archon_component_ablations_with_gpt4o}
\end{table}

\begin{table}[]
\centering
\scriptsize
\setlength{\tabcolsep}{1.5pt}

\begin{tabular}{clcccccccc}
\toprule
                           & & & \textbf{\begin{tabular}[c]{@{}c@{}}MT\\ Bench\end{tabular}} & \begin{tabular}[c]{@{}c@{}}\textbf{AlpacaEval}\\\textbf{2.0}\end{tabular} & \textbf{\begin{tabular}[c]{@{}c@{}}Arena\\ Hard Auto\end{tabular}} & \textbf{\begin{tabular}[c]{@{}c@{}}MixEval\\ Hard\end{tabular}} & \textbf{MixEval} & \textbf{MATH} & \textbf{\begin{tabular}[c]{@{}c@{}}Code\\Contests\end{tabular}} \\
                           \cmidrule(l{7pt}r{7pt}){1-10}
                           & \multicolumn{1}{c}{\textbf{Model / LLM System}} & \begin{tabular}[c]{@{}c@{}}\textbf{\# of}\\ \textbf{Infer.}\\ \textbf{Calls}\end{tabular} & W.R. & \begin{tabular}[c]{@{}c@{}}L.C.\\ W.R.\end{tabular} & W.R. & Acc. & Acc. & Acc. & Acc. \\ \midrule
 \multirow{3}{*}{\rotatebox{90}{\begin{tabular}[c]{@{}c@{}} \tiny \textbf{Control}\end{tabular}}} & Single Generation & 1 & 32.1\% ±0.7 & 38.5\% ±0.5 & 30.4\% ±0.6 & {45.2\% ±0.3} & 69.5\% ±0.2 & {72.3\% ±0.5} & 10.5\% ±0.6 \\
 & Ensemble \textit{+ Fuser}         & 11 & 44.2\% ±0.3 & 43.0\% ±0.6 & 40.2\% ±0.4 & 46.0\% ±0.4 & 73.0\% ±0.3 & 70.5\% ±0.7 & 6.0\% ±0.4 \\ 
 & Ensemble \textit{+ Critic + Fuser}        & 12 & 46.6\% ±0.5 & 44.2\% ±0.4 & 44.4\% ±0.7 & 47.9\% ±0.2 & 73.0\% ±0.4 & 72.5\% ±0.3 & 8.4\% ±0.5 \\ 
 \midrule
 \multirow{8}{*}{\rotatebox{90}{\begin{tabular}[c]{@{}c@{}} \tiny\textbf{Ablations} \end{tabular}}} & Ensemble + \textit{Ranker}  & 11 & 38.1\% ±0.6 & 40.2\% ±0.7 & 36.0\% ±0.5 & 43.8\% ±0.3 & 72.1\% ±0.2 & 66.2\% ±0.6 & 7.5\% ±0.4 \\
 & Ensemble + \textit{Verifier} & 11 & 38.9\% ±0.4 & 41.7\% ±0.3 & 38.2\% ±0.8 & 41.9\% ±0.4 & 71.0\% ±0.3 & 68.5\% ±0.4 & {19.0\% ±0.7} \\
 & Ensemble + \textit{Unit Test Gen./Eval.}  & 21 & 37.3\% ±0.8 & 39.8\% ±0.6 & 31.7\% ±0.3 & 40.7\% ±0.2 & 71.2\% ±0.4 & 69.8\% ±0.8 & \underline{22.0\% ±0.3} \\ 
 & Ensemble + \textit{Ranker + Fuser}        & 12 & \underline{48.0\% ±0.2} & 45.6\% ±0.5 & 42.7\% ±0.6 & 45.0\% ±0.3 & 73.0\% ±0.2 & 70.1\% ±0.5 & 8.0\% ±0.6 \\
 & Ensemble + \textit{Verifier + Fuser}        & 12 & 46.3\% ±0.5 & 44.7\% ±0.4 & \underline{45.0\% ±0.4} & 50.5\% ±0.4 & 73.0\% ±0.3 & {71.3\% ±0.3} & 18.6\% ±0.5 \\ 
 & Ensemble + \textit{Unit Test Gen./Eval. + Fuser}  & 22 & 47.0\% ±0.3 & 43.9\% ±0.7 & 42.1\% ±0.7 & 48.2\% ±0.2 & 72.2\% ±0.4 & 73.1\% ±0.6 & \textbf{23.5\% ±0.4} \\
 & Ensemble + \textit{Critic + Verifier + Fuser}        & 13 & 47.1\% ±0.7 & \underline{46.0\% ±0.3} & \underline{45.0\% ±0.5} & \underline{52.4\% ±0.3} & \underline{73.2\% ±0.5} & \underline{74.1\% ±0.4} & 18.4\% ±0.7 \\
 & Ensemble + \textit{Critic + Ranker + Fuser} & 13 & \textbf{50.5\% ±0.4} & \textbf{48.3\% ±0.6} & \textbf{46.7\% ±0.3} & \textbf{55.1\% ±0.4} & \textbf{73.7\% ±0.3} & \textbf{76.4\% ±0.5} & 8.1\% ±0.5 \\
 \bottomrule                                            
\end{tabular}
\caption{ \textbf{\archon{} Component Compositions with GPT-4o-mini}: The ensemble uses generates 10 samples for the given query.
The standard error numbers were calculated from 10 independent evaluation runs.
}
\label{tab:archon_component_ablations_with_gpt4o_mini}
\end{table}

\begin{table}[]
\centering
\scriptsize
\setlength{\tabcolsep}{1.5pt}

\begin{tabular}{clcccccccc}
\toprule
                           & & & \textbf{\begin{tabular}[c]{@{}c@{}}MT\\ Bench\end{tabular}} & \begin{tabular}[c]{@{}c@{}}\textbf{AlpacaEval}\\\textbf{2.0}\end{tabular} & \textbf{\begin{tabular}[c]{@{}c@{}}Arena\\ Hard Auto\end{tabular}} & \textbf{\begin{tabular}[c]{@{}c@{}}MixEval\\ Hard\end{tabular}} & \textbf{MixEval} & \textbf{MATH} & \textbf{\begin{tabular}[c]{@{}c@{}}Code\\Contests\end{tabular}} \\
                           \cmidrule(l{7pt}r{7pt}){1-10}
                           & \multicolumn{1}{c}{\textbf{Model / LLM System}} & \begin{tabular}[c]{@{}c@{}}\textbf{\# of}\\ \textbf{Infer.}\\ \textbf{Calls}\end{tabular} & W.R. & \begin{tabular}[c]{@{}c@{}}L.C.\\ W.R.\end{tabular} & W.R. & Acc. & Acc. & Acc. & Acc. \\ \midrule
 \multirow{3}{*}{\rotatebox{90}{\begin{tabular}[c]{@{}c@{}} \tiny \textbf{Control}\end{tabular}}} & Single Generation & 1 & N/A & 52.7\% ±0.4 & 81.4\% ±0.6 & 68.7\% ±0.3 & 89.1\% ±0.2 & 83.5\% ±0.5 & 12.5\% ±0.3 \\
 & Ensemble \textit{+ Fuser}         & 11 & N/A & 53.0\% ±0.6 & 83.2\% ±0.4 & 69.5\% ±0.2 & 89.0\% ±0.3 & 81.8\% ±0.6 & 17.0\% ±0.4 \\ 
 & Ensemble \textit{+ Critic + Fuser}        & 12 & N/A & 54.2\% ±0.3 & \underline{85.4\% ±0.7} & \underline{70.9\% ±0.4} & 89.5\% ±0.2 & 82.6\% ±0.4 & 19.4\% ±0.6 \\ 
 \midrule
 \multirow{8}{*}{\rotatebox{90}{\begin{tabular}[c]{@{}c@{}} \tiny\textbf{Ablations} \end{tabular}}} & Ensemble + \textit{Ranker}  & 11 & N/A & 50.2\% ±0.5 & 85.7\% ±0.5 & 63.8\% ±0.3 & 82.1\% ±0.4 & 80.2\% ±0.7 & 18.5\% ±0.5 \\
 & Ensemble + \textit{Verifier} & 11 & N/A & 51.7\% ±0.7 & 78.2\% ±0.3 & 60.9\% ±0.2 & 81.0\% ±0.3 & 80.1\% ±0.3 & 21.0\% ±0.4 \\
 & Ensemble + \textit{Unit Test Gen./Eval.}  & 21 & N/A & 49.8\% ±0.4 & 71.7\% ±0.8 & 59.0\% ±0.2 & 81.2\% ±0.2 & 80.9\% ±0.8 & \underline{22.0\% ±0.7} \\ 
 & Ensemble + \textit{Ranker + Fuser}        & 12 & N/A & 55.6\% ±0.5 & 82.7\% ±0.4 & 65.0\% ±0.3 & 89.0\% ±0.4 & 82.4\% ±0.4 & 19.0\% ±0.3 \\
 & Ensemble + \textit{Verifier + Fuser}        & 12 & N/A & 54.7\% ±0.3 & 85.0\% ±0.6 & 70.5\% ±0.2 & \underline{89.3\% ±0.3} & \underline{84.1\% ±0.6} & 21.6\% ±0.5 \\ 
 & Ensemble + \textit{Unit Test Gen./Eval. + Fuser}  & 22 & N/A & 53.9\% ±0.6 & 82.1\% ±0.5 & 68.2\% ±0.4 & 89.2\% ±0.2 & 82.0\% ±0.5 & \textbf{23.5\% ±0.6} \\
 & Ensemble + \textit{Critic + Verifier + Fuser}        & 13 & N/A & \underline{56.0\% ±0.4} & 85.0\% ±0.3 & 71.0\% ±0.3 & 89.4\% ±0.4 & 83.1\% ±0.3 & 21.4\% ±0.4 \\
 & Ensemble + \textit{Critic + Ranker + Fuser} & 13 & N/A & \textbf{58.3\% ±0.5} & \textbf{86.7\% ±0.7} & \textbf{73.0\% ±0.2} & \textbf{89.7\% ±0.3} & \textbf{85.3\% ±0.7} & 19.1\% ±0.5 \\
 \bottomrule                                            
\end{tabular}
\caption{ \textbf{\archon{} Component Compositions with Claude 3.5 Sonnet}: The ensemble uses generates 10 samples for the given query.
The standard error numbers were calculated from 10 independent evaluation runs.
}
\label{tab:archon_component_ablations_with_sonnet}
\end{table}

\begin{table}[]
\centering
\scriptsize
\setlength{\tabcolsep}{1.5pt}

\begin{tabular}{clcccccccc}
\toprule
                           & & & \textbf{\begin{tabular}[c]{@{}c@{}}MT\\ Bench\end{tabular}} & \begin{tabular}[c]{@{}c@{}}\textbf{AlpacaEval}\\\textbf{2.0}\end{tabular} & \textbf{\begin{tabular}[c]{@{}c@{}}Arena\\ Hard Auto\end{tabular}} & \textbf{\begin{tabular}[c]{@{}c@{}}MixEval\\ Hard\end{tabular}} & \textbf{MixEval} & \textbf{MATH} & \textbf{\begin{tabular}[c]{@{}c@{}}Code\\Contests\end{tabular}} \\
                           \cmidrule(l{7pt}r{7pt}){1-10}
                           & \multicolumn{1}{c}{\textbf{Model / LLM System}} & \begin{tabular}[c]{@{}c@{}}\textbf{\# of}\\ \textbf{Infer.}\\ \textbf{Calls}\end{tabular} & W.R. & \begin{tabular}[c]{@{}c@{}}L.C.\\ W.R.\end{tabular} & W.R. & Acc. & Acc. & Acc. & Acc. \\ \midrule
 \multirow{3}{*}{\rotatebox{90}{\begin{tabular}[c]{@{}c@{}} \tiny \textbf{Control}\end{tabular}}} & Single Generation & 1 & 35.0\% ±0.5 & 42.0\% ±0.6 & 36.8\% ±0.7 & {64.6\% ±0.2} & 73.2\% ±0.3 & 74.3\% ±0.4 & 10.0\% ±0.5 \\
 & Ensemble \textit{+ Fuser}         & 11 & 48.2\% ±0.3 & 47.0\% ±0.4 & 44.2\% ±0.5 & 66.5\% ±0.3 & {77.0\% ±0.2} & 75.1\% ±0.7 & 10.8\% ±0.3 \\ 
 & Ensemble \textit{+ Critic + Fuser}        & 12 & 50.6\% ±0.7 & 48.2\% ±0.5 & 48.4\% ±0.3 & 68.1\% ±0.4 & {77.0\% ±0.4} & 76.3\% ±0.5 & 11.5\% ±0.6 \\ 
 \midrule
 \multirow{8}{*}{\rotatebox{90}{\begin{tabular}[c]{@{}c@{}} \tiny\textbf{Ablations} \end{tabular}}} & Ensemble + \textit{Ranker}  & 11 & 42.1\% ±0.4 & 44.2\% ±0.7 & 40.0\% ±0.6 & 58.8\% ±0.3 & 76.1\% ±0.2 & 71.8\% ±0.6 & 11.9\% ±0.4 \\
 & Ensemble + \textit{Verifier} & 11 & 42.9\% ±0.6 & 45.7\% ±0.3 & 42.2\% ±0.8 & 57.9\% ±0.2 & 75.0\% ±0.3 & 70.5\% ±0.4 & \underline{12.0\% ±0.7} \\
 & Ensemble + \textit{Unit Test Gen./Eval.}  & 21 & 41.3\% ±0.8 & 43.8\% ±0.6 & 35.7\% ±0.4 & 55.7\% ±0.4 & 75.2\% ±0.2 & 74.1\% ±0.8 & 13.0\% ±0.3 \\ 
 & Ensemble + \textit{Ranker + Fuser}        & 12 & \underline{52.0\% ±0.2} & 49.6\% ±0.5 & 46.7\% ±0.7 & 60.0\% ±0.3 & {77.0\% ±0.4} & 75.0\% ±0.5 & 12.0\% ±0.6 \\
 & Ensemble + \textit{Verifier + Fuser}        & 12 & 50.3\% ±0.5 & 48.7\% ±0.4 & {48.7\% ±0.5} & 67.5\% ±0.2 & {77.0\% ±0.3} & {77.4\% ±0.3} & 10.5\% ±0.5 \\ 
 & Ensemble + \textit{Unit Test Gen./Eval. + Fuser}  & 22 & 51.0\% ±0.3 & 47.9\% ±0.7 & 46.1\% ±0.6 & 64.2\% ±0.4 & 76.2\% ±0.2 & 78.3\% ±0.6 & \textbf{14.3\% ±0.4} \\
 & Ensemble + \textit{Critic + Verifier + Fuser}        & 13 & 51.1\% ±0.7 & \underline{50.0\% ±0.3} & \underline{49.0\% ±0.4} & \underline{68.0\% ±0.3} & \underline{77.2\% ±0.4} & \underline{77.8\%} ±0.3 & 10.0\% ±0.7 \\
 & Ensemble + \textit{Critic + Ranker + Fuser} & 13 & \textbf{54.5\% ±0.4} & \textbf{52.3\% ±0.6} & \textbf{50.7\% ±0.3} & \textbf{70.4\% ±0.2} & \textbf{77.7\% ±0.3} & \textbf{80.5\% ±0.5} & 11.5\% ±0.5 \\
 \bottomrule                                            
\end{tabular}
\caption{ \textbf{\archon{} Component Compositions with Claude-3-Haiku}: The ensemble uses generates 10 samples for the given query.
The standard error numbers were calculated from 10 independent evaluation runs.
}
\label{tab:archon_component_ablations_with_haiku}
\end{table}

\subsubsection{Verifier} 

\textbf{Utility}: The Verifier was most effective for the reasoning benchmarks explored in \autoref{tab:archon_component_ablations}.
When just using a 70B+ Generator ensemble with Verifier module after generation, the \archon{} configuration lagged behind the \archon{} ensemble and fuser configuration by 1.5 percentage points, on average, across all benchmarks explored.
This suggests that the Verifier is most effective when combined with other inference-time techniques.

\noindent \textbf{Component Interactions}: 
As noted in Section \ref{sec:fuser_analysis}, the Verifier augmented the performance of the Critic and Fuser on reasoning tasks (e.g. Arena-Hard-Auto, MixEval, MixEval-Hard), boosting performance by 3.7 percentage points, on average, when combined together with these modules.
Overall, the Verifier is most powerful when augmenting additional components for tasks requiring verification of intermediate steps and the final response (\autoref{tab:archon_component_ablations}).
Therefore, the Verifier was less helpful for instruction-following tasks (e.g. MT Bench and AlpacaEval) but more effective for reasoning tasks (e.g. Arena-Hard-Auto and MixEval).

\subsubsection{Unit Test Generator and Evaluator}

\textbf{Utility}: 
The Unit Test Generator and Evaluator were most effective on reasoning and coding tasks, improving performance on benchmarks that required more verification steps, such as Arena-Hard-Auto, MixEval, MixEval-Hard, MATH, and CodeContests (\autoref{tab:archon_component_ablations}). 
For the reasoning tasks, we found the unit test generator and evaluator to be most effective when combined with other components. 
When the 70B+ ensemble of Generators was only combined with unit tests, it was less effective for reasoning tasks like Arena-Hard-Auto and MixEval, lagging behind the ensemble and fuser configuration by 3.1 percentage points.
This inspired us to look into other inference-time techniques combinations for unit test generation, such as increased sampling and fusion.
When we increased generation sampling and added unit test generation/evaluation for CodeContests, we see a 56\% boost in Pass@1 performance (\autoref{tab:archon_main_table_results}), increasing from 17.9 to 29.3 Pass@1.

\noindent \textbf{Component Interactions}: 
When combined with the Fuser module, the Unit Test Generator and Evaluator improved performance by 2.1 percentage points across the benchmarks explored (\autoref{tab:archon_component_ablations}).
The combined ensemble, Unit Test Generator/Evaluator, and Fuser \archon{} configuration was most effective on the reasoning benchmarks, leading to a 2.5 percentage point boost, on average.
For coding, the unit test generator and evaluator was most effective when combined with the best performing Generator (using large sample counts) and a final Fuser (\autoref{sec:archon_results}).

\begin{figure*}
   \centering
   \includegraphics[width=1.0\linewidth]{figures/Archon_Budget_Analysis_v2.pdf}
   \caption{\textbf{\archon{}'s Performance Exceeds Baselines across Token Budgets}: 
   Across different token budgets (Section \ref{sec:search_algorithms}), we compare \archon{} architectures against top-performing inference-time system baselines.
   The MoA architecture and OpenAI's o1 are static so they use the same number of tokens across budgets.
   The results were averaged over 10 independent evaluation runs. $^*$MATH and CodeContests use a subset of their test sets for evaluation (Section \ref{sec:benchmarks_and_models}).
   }
   \label{fig:Archon_Budget_Analysis}
\end{figure*}

\begin{figure*}[t]
   \centering
   \includegraphics[width=\linewidth]{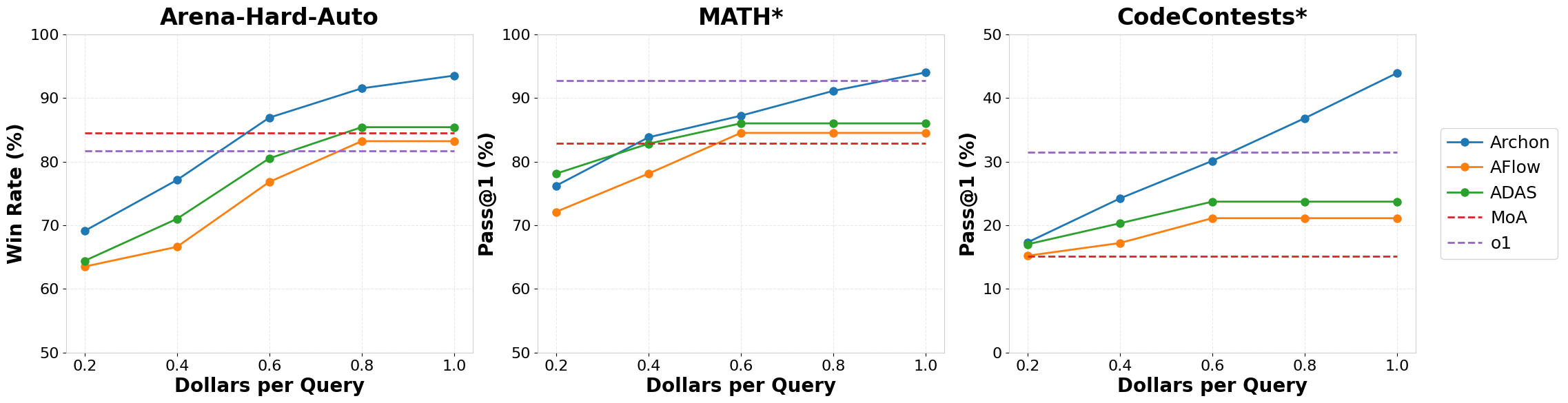}
   \caption{{\textbf{\archon{}'s Performance Exceeds Baselines across Dollar per Query Budgets}: 
   Across different dollar per query budgets (Section \ref{sec:search_algorithms}), we compare \archon{} architectures against top-performing inference-time system baselines.
   The MoA architecture and OpenAI's o1 are static so they use the same number of tokens across budgets.
   The results were averaged over 10 independent evaluation runs. $^*$MATH and CodeContests use a subset of their test sets for evaluation (Section \ref{sec:benchmarks_and_models}).
   }
   }
   \label{fig:Dollars_per_Query_vs_Performance}
\end{figure*}

\begin{table}[H]
    \scriptsize
    \centering
    \begin{tabular}{|m{0.9\textwidth}|}
        \hline
        \begin{minipage}{0.9\textwidth}
            \vspace{0.5cm}
            \texttt{<instruction here>}.
            \vspace{0.5cm}
        \end{minipage} \\
        \hline
    \end{tabular}
    \caption{\textbf{Generator Prompt}}
    \label{tab:generator_prompt}
\end{table}

\begin{table}[H]
    \centering
    \scriptsize
    \begin{minipage}{0.9\textwidth} %
        \centering
        \begin{tabular}{|m{\textwidth}|}
            \hline
            \begin{minipage}{\textwidth}
                \vspace{0.3cm}
                \setlength{\baselineskip}{0.9\baselineskip} %
                \setlength{\parskip}{0pt} %
                You have been provided with a set of responses with their individual critiques of strengths/weaknesses from various open-source models to the latest user query. Your task is to synthesize these responses into a single, high-quality response. It is crucial to critically evaluate the information provided in these responses and their provided critiques of strengths/weaknesses, recognizing that some of it may be biased or incorrect. Your response should not simply replicate the given answers but should offer a refined, accurate, and comprehensive reply to the instruction. Ensure your response is well-structured, coherent, and adheres to the highest standards of accuracy and reliability.
                \newline
                Responses from models:
                \newline
                \texttt{1.} \texttt{<response \#1>} \newline \texttt{Critique:} \texttt{<critique \#1>} \newline
                \texttt{2.} \texttt{<response \#2>} \newline
                \texttt{Critique:} \texttt{<critique \#2>}
                \newline
                \texttt{...} \newline
                \texttt{N.} \texttt{<response \#N>} \newline
                \texttt{Critique:} \texttt{<critique \#N>}
                \newline
                \texttt{<instruction here>}
                \vspace{0.3cm}
            \end{minipage} \\
            \hline
        \end{tabular}
        \caption{With Critiques}
        \label{tab:fuser_prompt_part2}
    \end{minipage}
    
    \vspace{0.3cm} %
    
    \begin{minipage}{0.9\textwidth} %
        \centering
        \scriptsize
        \begin{tabular}{|m{\textwidth}|}
            \hline
            \begin{minipage}{\textwidth}
                \vspace{0.3cm}
                \setlength{\baselineskip}{0.9\baselineskip} %
                \setlength{\parskip}{0pt} %
                You have been provided with a set of responses from various open-source models to the latest user query. Your task is to synthesize these responses into a single, high-quality response. It is crucial to critically evaluate the information provided in these responses, recognizing that some of it may be biased or incorrect. Your response should not simply replicate the given answers but should offer a refined, accurate, and comprehensive reply to the instruction. Ensure your response is well-structured, coherent, and adheres to the highest standards of accuracy and reliability.
                \newline
                \texttt{1.} \texttt{<response \#1>} \newline
                \texttt{2.} \texttt{<response \#2>} \newline
                \texttt{...} \newline
                \texttt{N.} \texttt{<response \#N>} \newline
                \texttt{<instruction here>}
                \vspace{0.3cm}
            \end{minipage} \\
            \hline
        \end{tabular}
        \caption{Without Critiques}
        \label{tab:fuser_prompt_part1}
    \end{minipage}
    
    \caption{\textbf{Fuser Prompt: Without and With Critiques}}
    \label{tab:fuser_prompt}
\end{table}

\begin{table}[H]
    \centering
    \scriptsize
    \begin{tabular}{|m{\textwidth}|}
        \hline
        \begin{minipage}{\textwidth}
            \vspace{0.3cm} %
            \setlength{\baselineskip}{0.9\baselineskip} %
            \setlength{\parskip}{0pt} %
            I will provide you with \texttt{N} responses, each indicated by a numerical identifier \texttt{[]}. Rank the responses based on their relevance to the instruction: \texttt{<instruction here>}. \newline
            \texttt{[1]} \texttt{<response \#1>} \newline
            \texttt{[2]} \texttt{<response \#2>} \newline
            \texttt{...} \newline
            \texttt{[N]} \texttt{<response \#N>} \newline
            Instruction: \texttt{<instruction here>}. \newline 
            Rank the \texttt{N} responses above based on their relevance to the instruction. All the responses should be included and listed using identifiers, in descending order of relevance to the instruction. The output format should be \texttt{[]} \texttt{>} \texttt{[]}, e.g., \texttt{[4] > [2]}. Only respond with the ranking results, do not say any word or explain.
            \vspace{0.3cm} %
        \end{minipage} \\
        \hline
    \end{tabular}
    \caption{\textbf{Decoder-Based Ranking Prompt}}
    \label{tab:decoder_ranker_prompt}
\end{table}

\begin{table}[H]
    \centering
    \scriptsize
    \begin{tabular}{|m{\textwidth}|}
        \hline
        \begin{minipage}{\textwidth}
            \vspace{0.2cm} %
            \setlength{\baselineskip}{0.9\baselineskip} %
            \setlength{\parskip}{0pt} %
            You are a helpful assistant. I will provide you with \texttt{N} responses, each indicated by a numerical identifier (e.g., [1], [2], etc.). Rank the responses based on their relevance to the instruction: \texttt{<instruction here>}. \newline
            \texttt{[1]} \texttt{<response \#1>} \newline
            \texttt{[2]} \texttt{<response \#2>} \newline
            \texttt{...} \newline
            \texttt{[N]} \texttt{<response \#N>} \newline
            Instruction: \texttt{<instruction here>}. \newline
            Evaluate the \texttt{N} responses above based on their relevance to the instruction. All the responses should be included and listed using identifiers. For each response, start the critique with the numerical identifier (e.g., [1]) followed by the strengths and weaknesses. You must include both strengths and weaknesses, even if there are more of one than the other. At the end of each response's analysis, include two new lines to separate the critiques. Do not include any preface or text after the critiques. Do not include any references to previous critiques within a critique. Start with the analysis for the first response and end with the analysis for the last response. All of the \texttt{N} responses should be included and evaluated using identifiers. Structure each response's analysis as follows:
            \newline 
            \texttt{Strengths:}
            \newline 
            \texttt{- <strength \#1>}
            \newline 
            \texttt{- <strength \#2>}
            \newline 
            \texttt{- <strength \#n>}
            \newline 
            \texttt{Weaknesses:}
            \newline 
            \texttt{- <weakness \#1>}
            \newline 
            \texttt{- <weakness \#2>}
            \newline 
            \texttt{- <weakness \#n>} 
            \vspace{0.2cm} %
        \end{minipage} \\
        \hline
    \end{tabular}
    \caption{\textbf{Critic Prompt}}
    \label{tab:critic_prompt}
\end{table}

\begin{table}[H]
    \centering
    \scriptsize
    \begin{tabular}{|m{\textwidth}|}
        \hline
        \begin{minipage}{\textwidth}
            \vspace{0.2cm} %
            \setlength{\baselineskip}{0.9\baselineskip} %
            \setlength{\parskip}{0pt} %
                I will provide you with a response indicated by the identifier 'Response'. Provide reasoning for why the response accurately and completely addresses the instruction: \texttt{<instruction here>}. \newline 
                Response: \texttt{<response>} \newline
                Instruction: \texttt{<instruction here>}. \newline 
                Provide the reasoning for the response above based on its relevance, completeness, and accuracy when compared to the instruction. Do not include any preface or text after the reasoning.
                \vspace{0.2cm} %
        \end{minipage} \\
        \hline
    \end{tabular}
    \caption{\textbf{Verifier Prompt}}
    \label{tab:verifier_prompt}
\end{table}

\begin{table}[H]
    \centering
    \scriptsize
    \begin{minipage}{0.9\textwidth} %
        \centering
        \scriptsize
        \begin{tabular}{|m{\textwidth}|}
            \hline
            \begin{minipage}{\textwidth}
                \vspace{0.3cm} %
                \setlength{\baselineskip}{0.85\baselineskip} %
                \setlength{\parskip}{0pt} %
                \setlength{\parindent}{0pt} %
                \textbf{Instruction Prompt:} Given the following query, generate a set of \texttt{N} unit tests that would evaluate the correctness of responses to this query. \newline
                - The unit tests should cover various aspects of the query and ensure comprehensive evaluation. \newline
                - Each unit test should be clearly stated and should include the expected outcome. \newline
                - The unit tests should be in the form of assertions that can be used to validate the correctness of responses to the query. \newline
                - The unit test should be formatted like 'The answer mentions...', 'The answer states...', 'The answer uses...', etc. followed by the expected outcome. \newline
                - Solely provide the unit tests for the question below. Do not provide any text before or after the list. Only output the unit tests as a list of strings (e.g., ['unit test \#1', 'unit test \#2', 'unit test \#3']). \newline
                Query: \texttt{<instruction here>} \newline
                \vspace{0.3cm} %
            \end{minipage} \\
            \hline
        \end{tabular}
        \caption{With Unit Test Cap}
        \label{tab:unit_test_generator_prompt_part1}
    \end{minipage}
    
    \vspace{0.3cm} %

    \begin{minipage}{0.9\textwidth} %
        \centering
        \scriptsize
        \begin{tabular}{|m{\textwidth}|}
            \hline
            \begin{minipage}{\textwidth}
                \vspace{0.3cm} %
                \setlength{\baselineskip}{0.85\baselineskip} %
                \setlength{\parskip}{0pt} %
                \setlength{\parindent}{0pt} %
                \textbf{Instruction Prompt:} Given the following query, generate a set of unit tests that would evaluate the correctness of responses to this query. \newline
                - The unit tests should cover various aspects of the query and ensure comprehensive evaluation. \newline
                - Each unit test should be clearly stated and should include the expected outcome. \newline
                - The unit tests should be in the form of assertions that can be used to validate the correctness of responses to the query. \newline
                - The unit test should be formatted like 'The answer mentions...', 'The answer states...', 'The answer uses...', etc. followed by the expected outcome. \newline
                - Solely provide the unit tests for the question below. Do not provide any text before or after the list. Only output the unit tests as a list of strings (e.g., ['unit test \#1', 'unit test \#2', 'unit test \#3']). \newline
                Query: \texttt{<instruction here>} \newline
                \vspace{0.3cm} %
            \end{minipage} \\
            \hline
        \end{tabular}
        \caption{Without Unit Test Cap}
        \label{tab:unit_test_generator_prompt_part2}
    \end{minipage}
    
    \caption{\textbf{Unit Test Generator Prompt: With and Without Unit Test Cap}}
    \label{tab:unit_test_generator_prompt}
\end{table}

\begin{table}[H]
    \centering
    \scriptsize
    \begin{minipage}{0.9\textwidth} %
        \centering
        \scriptsize
        \begin{tabular}{|m{\textwidth}|}
            \hline
            \begin{minipage}{\textwidth}
                \vspace{0.5cm}
                \textbf{Instruction Prompt:} Compose an engaging travel blog post about a recent trip to Hawaii, highlighting cultural experiences and must-see attractions.
                \begin{enumerate}
                    \item Unit Test \#1: The blog post mentions at least two cultural experiences specific to Hawaii.
                    \item Unit Test \#2: The blog post highlights at least three must-see attractions in Hawaii.
                    \item Unit Test \#3: The tone of the blog post is engaging and uses descriptive language that would appeal to readers interested in travel.
                    \item Unit Test \#4: The blog post includes factual information about Hawaii's culture, such as local customs, festivals, or historical facts.
                    \item Unit Test \#5: The blog post contains a clear narrative structure, including an introduction, main body, and a conclusion.
                    \vspace{0.5cm}
                \end{enumerate}
            \end{minipage} \\
            \hline
        \end{tabular}
        \caption{Instruction-Following Query}
    \end{minipage}
    
    \vspace{0.5cm} %

    \begin{minipage}{0.9\textwidth} %
        \centering
        \scriptsize
        \begin{tabular}{|m{\textwidth}|}
            \hline
            \begin{minipage}{\textwidth}
                \vspace{0.5cm}
                \textbf{Instruction Prompt:} Alice and Bob have two dice. They roll the dice together, note the sum of the two values shown, and repeat. For Alice to win, two consecutive turns (meaning, two consecutive sums) need to result in 7. For Bob to win, he needs to see an eight followed by a seven. Who do we expect to win this game?
                \begin{enumerate}
                    \item Unit Test \#1: The response correctly identifies the winning condition for Alice (two consecutive sums of 7).
                    \item Unit Test \#2: The response correctly identifies the winning condition for Bob (a sum of 8 followed by a sum of 7).
                    \item Unit Test \#3: The response explains the probability of achieving two consecutive 7s when rolling two dice.
                    \item Unit Test \#4: The response explains the probability of achieving an 8 followed by a 7 when rolling two dice.
                    \item Unit Test \#5: The response provides a conclusion on who is more likely to win based on the probability analysis.
                    \vspace{0.5cm}
                \end{enumerate}
            \end{minipage} \\
            \hline
        \end{tabular}
        \caption{Reasoning Query}
    \end{minipage}
    
    \caption{\textbf{Unit Test Examples}}
    \label{tab:unit_test_examples}
\end{table}

\begin{table}[H]
    \centering
    \scriptsize
    \begin{tabular}{|p{0.95\textwidth}|}
        \hline
        \begin{minipage}{0.95\textwidth}
            \vspace{0.2cm} %
            \setlength{\baselineskip}{0.9\baselineskip} %
            \setlength{\parskip}{0pt} %
            Given the following query, candidate response, and unit tests, evaluate whether or not the response passes each unit test. \newline 
            - In your evaluation, you should consider how the response aligns with the unit tests, retrieved documents, and query. \newline
            - Provide reasoning before you return your evaluation. \newline
            - At the end of your evaluation, you must finish with a list of verdicts corresponding to each unit test. 
            \newline
            - You must include a verdict with one of these formatted options: '[Passed]' or '[Failed]'. 
            \newline
            - Here is an example of the output format: 
            \newline
            Unit Test \#1: [Passed] \newline
            Unit Test \#2: [Failed] \newline
            Unit Test \#3: [Passed] \newline
            - Each verdict should be on a new line and correspond to the unit test in the same position. 
            \newline
            - Here is the query, response, and unit tests for your evaluation:
            \newline
            \newline
            Query: \texttt{<instruction here>}. 
            \newline 
            \newline
            Candidate Response: 
            \texttt{<response>}\newline\newline
            Unit Tests: \newline
            \texttt{Unit Test \#1:} \texttt{<Unit Test \#1>} \newline
            \texttt{Unit Test \#2:} \texttt{<Unit Test \#2>} \newline
            \texttt{...} \newline
            \texttt{Unit Test \#N:} \texttt{<Unit Test \#N>} \newline
            \vspace{0.2cm} %
        \end{minipage} \\
        \hline
    \end{tabular}
    \caption{\textbf{Unit Test Evaluator Prompt}}
    \label{tab:unit_test_evaluator_prompt}
\end{table}

\subsection{Bayesian Optimization for \archon{}}
\label{sec:bayesian_optimization}

\subsubsection{\archon{} Search Space and Objective}

The \archon{} configuration space can be defined as $\mathcal{X} = \{x_g, x_s, x_f, x_r, x_c, x_v\}$ where:

\begin{itemize}
    \item $x_g \in [1,10]$ : Number of generator models
    \item $x_s \in [1,5]$ : Samples per generator (extends to $[1, 1000]$ for CodeContests)
    \item $x_f \in [1,4]$ : Number of fusion layers, including the final fusion layer at the end
    \item $x_r \in [2,10]$ : Number of models per fusion layer, ranging from 2 to 10 increments of 2
    \item $x_c \in \{0,1\}$ : Whether to use critic and ranker layers before each fuser
    \item $x_v \in \{0,1\}$ : Whether to use verification layer before final fusion
\end{itemize}

The total search space initially contains 18,750 configurations ($10 \cdot 5 \cdot 5^{(4-1)} \cdot 3 = 18,750$), reduced to 9,576 after removing invalid configurations where: 1) initial generations exceed fuser context window (24 candidates); and 2) single fuser layer contains multiple fusers ($x_f = 1$ while $x_r \geq 2$).

Let $f(x)$ be the objective function evaluating an \archon{} configuration $x \in \mathcal{X}$, defined as:

\begin{equation}
    f(x) = \text{Performance}(x) - \lambda \cdot \text{Cost}(x)
\end{equation}

where Performance$(x)$ is the accuracy on a 20\% sample of target tasks and Cost$(x)$ represents inference compute usage.

\subsubsection{Optimization Process}

We describe the components of the Bayesian optimization search approach for \archon{}.

First, define $\mathcal{H}$ as a history of architectures and their resulting objective values, which we accumulate throughout the optimization process. We use Expected Improvement (EI) as the acquisition function for determining how to select the next architecture configuration to search:
\begin{equation}
    \text{EI}(x; \mathcal{H}) = \mathbb{E}[\max(0, f(x) - f(x^+))],
\end{equation}

where $f(x^+)$ is the best-observed value of our objective so far for $(x^+, f(x^+)) \in \mathcal{H}$. We use a Gaussian Process model as a surrogate model for approximating $f(x)$.

We now describe the Bayesian optimization process. We first initialize the observation history $\mathcal{H} = \emptyset$. For timestep $t = 1, \dots, T$, where $T$ is the maximum number of iterations parameter, we do the following:

\begin{enumerate}
    \item An architecture $x_t$ is selected using the acquisition function, $x_t = \text{argmax}_{x \in \mathcal{X}} \text{EI}(x; \mathcal{H})$.
    \item This architecture $x_t$ is evaluated, and we obtain $f(x_t)$. 
    \item We use $f(x_t)$ to update our acquisition function and the surrogate model using $H \leftarrow H \cup (x_t, f(x_t))$.
\end{enumerate}

The process continues until either:
\begin{itemize}
    \item A maximum number of iterations is reached, $T$
    \item Performance convergence: $|f(x_{n+1}) - f(x_n)| < \epsilon$
    \item Budget exhaustion: $\text{Cost}(x_1,...,x_n) > B$
\end{itemize}

For \archon{}'s implementation, we initialize with 230-240 random configurations, as this was found to be optimal through empirical testing. Additional samples beyond this point provide diminishing returns and are better allocated to configuration search. For our implementation, we utilize the \href{https://github.com/bayesian-optimization/BayesianOptimization}{Bayesian Optimization python package} for global optimization with Gaussian processes.

This formulation allows \archon{} to efficiently explore the configuration space, requiring 88.5\% fewer evaluations than greedy search and 90.4\% fewer than random search, with Bayesian optimization finding the best architectures in 96.0\% of iterations. Traditional greedy search methods may perform comparably for limited inference budgets (<20 calls), but Bayesian optimization becomes increasingly effective as the search space and compute budget grow.

\begin{figure}[h]
   \centering
   \includegraphics[width=1.0\linewidth]{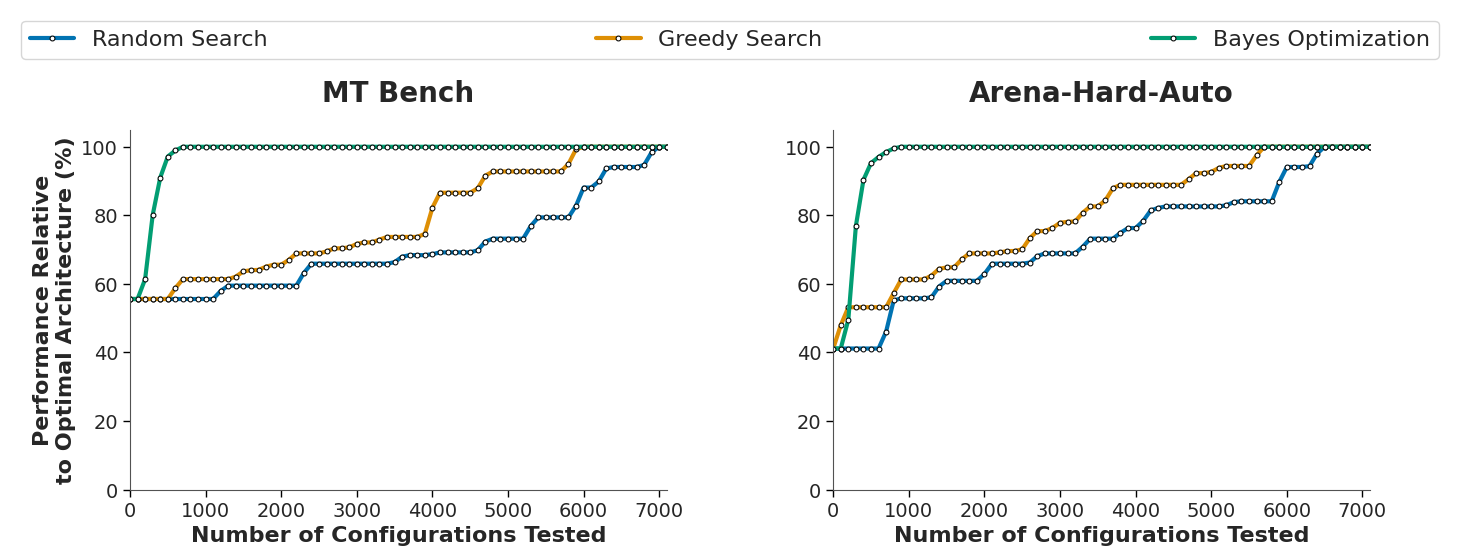}
   \caption{
   \textbf{Impact of Different Optimization Algorithms on \archon{}'s Architecture Search}:
   On the benchmarks MT Bench and Arena-Hard-Auto, we compare four approaches for finding the optimal inference-time architecture: random search, greedy search, and Bayes Optimization.
   Bayes Optimization finds the optimal architecture in 88.5\% less iterations compared to greedy search and 90.4\% less iterations compared to random search.
    }
   \label{fig:search_algorithms_comparison}
\end{figure}

\subsection{Bayes Optimization vs. Alternative Approaches}

\noindent \textbf{Search Techniques}: Within the hyperparameter space, we explored three search algorithms for automating the development of inference-time architectures:
\begin{enumerate}
    \item \textbf{Random Search}: Randomly selects a combination of hyperparameters for our \archon{} architecture.
    \item \textbf{Greedy Search}: Starting with a base \archon{} configuration, marginally changes each hyperparameter and test if it improves performance or not. If it does, incorporate the change. If not, move on to the next hyperparameter.
    \item \textbf{Bayesian Optimization}: Efficiently selects the most promising hyperparameter configurations for \archon{} by building a probabilistic surrogate model and leveraging an acquisition function for hyperparameter selection \citep{snoek2012practicalbayesianoptimizationmachine, 9124618} (Section \ref{sec:bayesian_optimization}).
\end{enumerate}

To get our model ranking for the benchmark, we calculate the model ranking by testing each model individually on a 20\% sample of each dataset benchmark in the first stage of the search.
To get our fusion model ranking for the benchmark, we use the same approach, testing each model's fusion performance with an ensemble of 10 randomly selected models from the available set. 
From our experiments, we found that the best generator and fusion models could vary widely dataset to dataset, making it beneficial to perform these rankings for new datasets (\autoref{tab:generator_and_fusion_rankings}).
For search, we use the same 20\% sample of each dataset that was used for evaluating generation and fusion, allowing us to guide architecture search with improved evaluation speed while getting meaningful development signal.

\noindent \textbf{Comparing Search Algorithms}: In \autoref{fig:search_algorithms_comparison}, we compare the effectiveness of each search algorithm on our explored benchmarks.
While random search guarantees the optimal \archon{} configuration, we found Bayesian optimization to be most effective in terms of tradeoff between finding the optimal configurations and minimizing the number of configurations tested.
For 96.0\% percent of the search iterations tested in \autoref{fig:search_algorithms_comparison}, we found that Bayesian optimization had the optimal configuration amongst the four explored search algorithms. 
We use 230 initial samples for our Bayes Optimization architecture search (Section \ref{sec:bayesian_optimization}).
Bayesian optimization also found the best architecture configuration in 88.5\% less evaluations than greedy search and 90.4\% less evaluations than random search.

\noindent \textbf{Bayesian Optimization Analysis}: 
In \autoref{tab:bayesian_optimization_comparisons}, we explore how the number of initial testing points, the number of exploration iterations, and the \archon{} inference call budget impacts the effectiveness of Bayesian optimization.
Additional initial testing points continue improving search efficacy up until 230-240 samples, where testing would be better delegated towards configuration search.
For lower inference call budgets with \archon{} (e.g. <20 inference calls), Bayesian optimization proved less effective, performing more similarly to greedy search or random search given the limited search space (\autoref{tab:search_algorithms_by_inference_call_budget}).
Therefore, Bayesian optimization is more effective for more open-ended \archon{} architecture search with larger inference call budgets (e.g. >20 inference calls) whereas traditional component engineering might be better for more limited inference call budgets.
\subsection{\archon{} Architecture Algorithms Comparisons}

\begin{table}[H]
\centering
\begin{tabular}{cc}
\begin{minipage}{0.45\textwidth}
\setlength{\tabcolsep}{1.8pt}
\centering
\scriptsize
\begin{tabular}{cccc}
\toprule
\textbf{\begin{tabular}[c]{@{}c@{}}\# of Init.\\ Points\end{tabular}} & \textbf{\begin{tabular}[c]{@{}c@{}}\% of Total\\ Configs\end{tabular}} & \textbf{\begin{tabular}[c]{@{}c@{}}Iter. till\\ Max. Config.\end{tabular}} & \textbf{\begin{tabular}[c]{@{}c@{}}Comb.\ Iter.\end{tabular}} \\
\midrule
200                          & 2.18\%               & 353                              & 553                    \\
210                          & 2.29\%            &  324                             & 534                    \\
220                          & 2.40\%               & 301                              & 521                    \\
230                          & 2.51\%                & 284                              & 514                    \\
240                          & 2.61\%                & \textbf{261}                              & \textbf{501}                    \\
250                          & 2.72\%                &  265                             & 515                    \\
260                          & 2.83\%                & 256                             & 516                    \\
270                          & 2.94\%                &   252                            & 522                    \\
\bottomrule
\end{tabular}
\caption{MT Bench}
\end{minipage}
\hspace{0.5cm} %
\begin{minipage}{0.45\textwidth}
\centering
\setlength{\tabcolsep}{1.8pt}
\scriptsize
\begin{tabular}{cccc}
\toprule
\textbf{\begin{tabular}[c]{@{}c@{}}\# of Init.\\ Points\end{tabular}} & \textbf{\begin{tabular}[c]{@{}c@{}}\% of Total\\ Configs\end{tabular}} & \textbf{\begin{tabular}[c]{@{}c@{}}Iter. till\\ Max. Config.\end{tabular}} & \textbf{\begin{tabular}[c]{@{}c@{}}Comb.\\ Iter.\end{tabular}} \\
\midrule
200                          & 2.18\%                &  478                             & 678                    \\
210                          & 2.29\%                & 431                              & 641                    \\
220                          & 2.40\%                & 415                              & 635                    \\
230                          & 2.51\%                & \textbf{382}                              & \textbf{612}                   \\
240                          & 2.61\%                & {389}                              & {629}                    \\
250                          & 2.72\%                & 385                               & 635                    \\
260                          & 2.83\%                & 372                              & 632                    \\
270                          & 2.94\%                & 368                              & 638                    \\
\bottomrule
\end{tabular}
\caption{Arena-Hard-Auto}
\end{minipage}
\end{tabular}
\caption{\textbf{Bayesian Optimization Hyperparameter Comparisons}: 
On MT Bench and Arena-Hard-Auto, we compare Bayesian optimization configurations for the number of initial sample points.
We find that 230 to 240 initial sample points minimizes the combined number of iterations (both initial sampling and exploring) to find the optimal configuration.
For the configurations explored, the total number of hyperparameter choices is 9,576.
}
\label{tab:bayesian_optimization_comparisons}
\end{table}

\begin{table}[H]
\centering
\scriptsize
\begin{tabular}{cccccc}
\toprule
                   & \multicolumn{5}{c}{\textbf{Iterations to Convergence}} \\ \midrule

\textbf{Inference Budget}   & 10       & 20       & 30        & 40        & 50       \\ \midrule
Random Selection   & 387       & 1152       & 2731       & 4359        & 5843      \\ 
Greedy Search      & 343       &  984      &  2153       & 3045        & 4895       \\
Bayes Optimization & \textbf{254}       & \textbf{386}       & \textbf{452}        & \textbf{515}        & \textbf{589}   \\ \bottomrule
\end{tabular}
\caption{\textbf{\archon{} Architecture Search Algorithms Comparison by Inference Call Budget}: For our comparison, we evaluate on MT Bench.}
\label{tab:search_algorithms_by_inference_call_budget}
\end{table}

\subsection{\archon{} Benchmarks and Results}

\begin{table}[!htbp]
\centering
\scriptsize
\setlength{\tabcolsep}{2.5pt}
\begin{tabular}{cccccc}
\toprule
\textbf{Benchmark} & \textbf{\begin{tabular}[c]{@{}c@{}}Example \\ Count\end{tabular}} & \textbf{\begin{tabular}[c]{@{}c@{}}Reference \\ Model\end{tabular}}     & \textbf{\begin{tabular}[c]{@{}c@{}}Judge \\ Model\end{tabular}}        & \textbf{Scoring Type}                                          & \textbf{Metric}                                                  \\ \midrule
AlpacaEval 2.0     & 805                                                               & GPT-4-Turbo                                                            & GPT-4-Turbo                                                            & \begin{tabular}[c]{@{}c@{}}Pairwise \\ Comparison\end{tabular} & \begin{tabular}[c]{@{}c@{}}L.C. \& Raw \\ Win Rates\end{tabular}  \\ \midrule
Arena-Hard-Auto    & 500                                                               & \begin{tabular}[c]{@{}c@{}}Claude-3.5-Sonnet\\GPT-4-0314\end{tabular}                                                             & GPT-4-Turbo                                                            & \begin{tabular}[c]{@{}c@{}}Pairwise \\ Comparison\end{tabular} & Win Rate                                                         \\ \midrule
MT-Bench           & 80                                                                & Claude-3.5-Sonnet                                                                  & GPT-4-0314                                                            & \begin{tabular}[c]{@{}c@{}}Pairwise \\ Comparison\end{tabular} & \begin{tabular}[c]{@{}c@{}}Adjusted \\ Win Rate\end{tabular}     \\ \midrule
MixEval            & 2000                                                              & N/A                                                                    & N/A                                                                    & Ground Truth                                                   & Accuracy                                                         \\ \midrule
MixEval-Hard       & 500                                                               & N/A                                                                    & N/A                                                                    & Ground Truth                                                   & Accuracy                                                         \\ \midrule
MATH       & \begin{tabular}[c]{@{}c@{}}200 \\ (sampled from 5000)\end{tabular}                                                               & N/A                                                                    & N/A                                                                    & Ground Truth                                                   & Pass@1                                                         \\ \midrule
CodeContests       & \begin{tabular}[c]{@{}c@{}}140 \\ (non-visual queries)\end{tabular}                                                               & N/A                                                                    & N/A                                                                    & Ground Truth                                                   & Pass@1                                                         \\ \bottomrule     %
\end{tabular}
\caption{\textbf{Benchmark Overview}: Evaluation configurations for AlpacaEval 2.0 \citep{alpaca_eval}, Arena-Hard-Auto \citep{arenahard2024}, MT-Bench \citep{zheng2023judging}, MixEval \citep{ni2024mixevalderivingwisdomcrowd}, MixEval Hard, MATH \citep{hendrycks2021measuring}, and CodeContests \citep{li2022competition}}.
\label{tab:benchmarks_overview}
\end{table}

\begin{figure*}[!htbp]%
   \centering
   \includegraphics[width=0.8\linewidth]{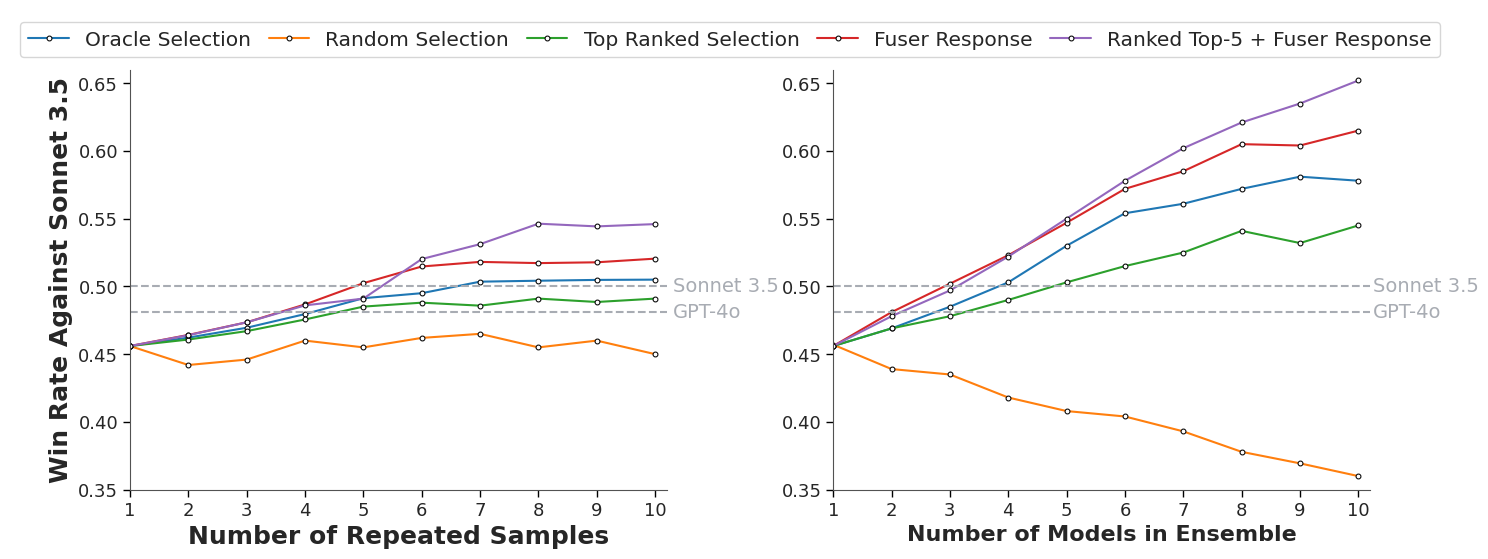}
   \caption{\textbf{Performance Gains from Repeated Sampling, Ensembling, Ranking, and Fusing on Arena-Hard-Auto}: 
   The \archon{} win-rate continues to grow significantly as we scale model sampling \textbf{(left)} or add additional models to the generator ensemble \textbf{(right)}, increasing by 9.3\% and 18.5\%, respectively. These best results are achieved by selecting the top-5 responses and fusing them. 
   The ensemble models are added based on their individual performance on this task, from best to worse (\autoref{tab:generator_and_fusion_rankings}).
   The oracle selection is the performance of picking the best answer generation out of all the generated samples from the ensemble.
   {The results were averaged over 10 independent evaluation runs.}
   }
   \label{fig:combined_sampling_and_ensembling_graphs}
\end{figure*}

\begin{table}[H]
\centering
\small
\begin{tabular}{cccc}
\toprule
\multicolumn{1}{l}{}           &                                                                               & \multicolumn{2}{c}{\textbf{Arena-Hard-Auto}} \\ \cmidrule(l{7pt}r{7pt}){3-4}
\multicolumn{1}{l}{}           & \textbf{Model / LLM System}                                                  & Score          & C.I.               \\ \midrule
\multicolumn{1}{l}{}           & Claude 3.5 Sonnet                                                             & N/A           & N/A        \\
\multicolumn{1}{l}{}           & GPT-4o                                                                        & 48.1\%           & (-2.3, 1.8)        \\
\multicolumn{1}{l}{}           & Llama 3.1 405B Instruct                                                       & 28.4\%           & (-2.7, 2.5)        \\ \midrule
\multirow{2}{*}{\rotatebox{90}{\textbf{\begin{tabular}[c]{@{}c@{}}Open\\ Source\end{tabular}}}}   & \begin{tabular}[c]{@{}c@{}}General-purpose\\ \archon{} Architecture\end{tabular} & 66.2\%         & (-2.4, 2.2)        \\
                               & \begin{tabular}[c]{@{}c@{}}Task-specific\\ \archon{} Architectures\end{tabular}  & 69.0\%         & (-2.8, 2.5)        \\ \midrule
\multirow{2}{*}{\rotatebox{90}{\textbf{\begin{tabular}[c]{@{}c@{}}Closed\\ Source\end{tabular}}}} & \begin{tabular}[c]{@{}c@{}}General-purpose\\ \archon{} Architecture\end{tabular} & 70.5\%         & (-2.5, 2.0)        \\
                               & \begin{tabular}[c]{@{}c@{}}Task-specific\\ \archon{} Architectures\end{tabular}  & 74.4\%         & (-2.3, 1.6)        \\ \midrule
\multirow{2}{*}{\rotatebox{90}{\textbf{\begin{tabular}[c]{@{}c@{}}All\\ Source\end{tabular}}}}     & \begin{tabular}[c]{@{}c@{}}General-purpose\\ \archon{} Architecture\end{tabular} & 72.5\%        & (-2.5, 1.8)        \\
                               & \begin{tabular}[c]{@{}c@{}}Task-specific\\ \archon{} Architectures\end{tabular}  & \textbf{76.1\%}         & (-1.8, 2.2)     \\ \bottomrule  
\end{tabular}
\caption{\textbf{\archon{} Results on Arena-Hard-Auto Results with Claude-3.5-Sonnet as Baseline Model}: The baseline model is Claude-3.5-Sonnet (default baseline model: GPT-4-0314) while the judge model is GPT-4-Turbo.}
\label{tab:arena_hard_auto_sonnet_baseline_results}
\end{table}

\begin{table}[H]
\scriptsize
\centering
\setlength{\tabcolsep}{3.0pt}
\begin{tabular}{ccccccccc}
\toprule
                                                                              & \multicolumn{1}{l}{}                                            & \multicolumn{6}{c}{\textbf{MixEval - Sub-Datasets}} & \multicolumn{1}{l}{}                                                       \\ \cmidrule(l{7pt}r{7pt}){3-8} 
\multicolumn{1}{c}{\textbf{Model / LLM System}}                                & \textbf{\begin{tabular}[c]{@{}c@{}}Infer.\\ Calls\end{tabular}} & GSM8K  & TriviaQA  & DROP  & MATH  & BBH  & AGIEval & \textbf{\begin{tabular}[c]{@{}c@{}}Average\end{tabular}} \\ \midrule
GPT-4o - 2024-05-13                                                           & 1                                                               & 94.9   & 89.1      & 88.2  & \textbf{98.5}  & 98.3 & 71.5    & 90.3                                                                       \\ \midrule
Claude 3.5 Sonnet                                                             & 1                                                               & 98.0   & 92.0      & 92.6  & 96    & 95.6 & 78.0    & 92.0                                                                       \\ \midrule
Llama 3.1 405B Instruct                                                       & 1     &    \textbf{98.2}  & 87.9  & 89.6  & 91.5     & 95.8   &  73.2         & 89.6                                                                    \\ \midrule
\begin{tabular}[c]{@{}c@{}}General-purpose\\ \archon{} Architecture\end{tabular} & 29                                                              & 98.3   & 94.8      & 94.6  & 98.1  & 97.3 & 82.1    & 94.2                                                                       \\ \midrule
\begin{tabular}[c]{@{}c@{}}Task-specific\\ \archon{} Architectures\end{tabular}  & 34                                                              & \textbf{98.2}   & \textbf{96.7}      & \textbf{95.6}  & \textbf{98.5}  & \textbf{98.8} & \textbf{84.2}    & \textbf{95.7}   \\ \bottomrule                                                                   
\end{tabular}
\caption{\textbf{MixEval Results by Sub-Dataset}: For the average computed, we do not introduce any weighting for each dataset.}
\label{tab:mixeval_subdataset_results}
\end{table}

\begin{table}[H]
\scriptsize
\centering
\setlength{\tabcolsep}{3.0pt}
\begin{tabular}{ccccccccc}
\toprule
\multicolumn{1}{l}{}                                                          & \multicolumn{1}{l}{}                                            & \multicolumn{6}{c}{\textbf{MixEval - Sub-Datasets}} & \multicolumn{1}{l}{}                                                       \\  \cmidrule(l{7pt}r{7pt}){3-8}
\textbf{Model / LLM System}                                                    & \textbf{\begin{tabular}[c]{@{}c@{}}Infer.\\ Calls\end{tabular}} & GSM8K  & TriviaQA  & DROP  & MATH  & BBH  & AGIEval & \textbf{\begin{tabular}[c]{@{}c@{}}Average\end{tabular}} \\ \midrule
GPT-4o - 2024-05-13                                                           & 1                                                               & 72.3   & 70.5      & 70.2  & 94.4  & 80.0 & 53.5    & 73.5                                                                       \\ \midrule
Claude 3.5 Sonnet                                                             & 1                                                               & 87.3   & 75.5      & 79.3  & 82.5  & 80.0 & 74.6    & 79.9                                                                       \\ \midrule
Llama 3.1 405B Instruct                                                       & 1                                                                       & 98.7 & 71.2 & 70.7 & 86.9 & 78.8 & 62.0 &  78.1                                                                      \\ \midrule
\begin{tabular}[c]{@{}c@{}}General-purpose\\ \archon{} Architecture\end{tabular} & 33                                                              & 96.7   & 82.7      & 83.2  & 93.4  & 82.0 & 76.7  & 85.8                                                                      \\ \midrule
\begin{tabular}[c]{@{}c@{}}Task-specific\\ \archon{} Architectures\end{tabular}  & 37                                                              & \textbf{98.9}   & \textbf{86.2}      & \textbf{85.2}  & \textbf{96.2}  & \textbf{86.0} & \textbf{80.1}    & \textbf{88.8} \\ \bottomrule                                                                     
\end{tabular}
\caption{\textbf{MixEval-Hard Results by Sub-Dataset}: For the average computed, we do not introduce any weighting for each dataset.}
\label{tab:mixeval_hard_subdataset_results}
\end{table}

\begin{table}[H]
\centering
\scriptsize
\begin{tabular}{ccccc}
\toprule
\textbf{}               & \textbf{GSM8K} & \textbf{\begin{tabular}[c]{@{}c@{}}MMLU\\ Math\end{tabular}} & \textbf{\begin{tabular}[c]{@{}c@{}}HumanEval \\ Python\end{tabular}} & \textbf{MBPP} \\ \midrule
\textbf{Model}          & Pass@1         & Pass@1                                                       & Pass@1                                                               & Pass@1        \\ \midrule
GPT-4o                  & 97.1\%         & 84.8\%                                                       & 89.0\%                                                               & 87.5\%        \\ \midrule
Claude 3.5 Sonnet       & 96.8\%         & 90.9\%                                                       & 90.2\%                                                               & 88.9\%        \\ \midrule
Llama 3.1 405B Instruct & 95.9\%         & 85.4\%                                                       & 90.2\%                                                               & 88.6\%  \\ \bottomrule     
\end{tabular}
\caption{\textbf{Additional Math and Code Benchmarks Explored}}
\label{tab:additional_code_benchmark_options}
\end{table}

\begin{table}[H]
\scriptsize
\setlength{\tabcolsep}{1.5pt}
\centering
\begin{tabular}{ccccccccccc}
\toprule
                           & & & & & \multicolumn{5}{c}{\textbf{Datasets}} \\ \cmidrule(l{7pt}r{7pt}){4-11}
                           & & & \begin{tabular}[c]{@{}c@{}}MT\\ Bench\end{tabular} & \multicolumn{2}{c}{\begin{tabular}[c]{@{}c@{}}Alpaca\\Eval 2.0\end{tabular}} & \begin{tabular}[c]{@{}c@{}}Arena\\ Hard Auto\end{tabular} & \begin{tabular}[c]{@{}c@{}}Arena\\ Hard Auto\end{tabular} & \begin{tabular}[c]{@{}c@{}}MixEval\\ Hard\end{tabular} & MixEval & MATH$^*$  
                           
\\ \midrule
                           & \textbf{Judge Model} & & \begin{tabular}[c]{@{}c@{}}GPT-4\\0314\end{tabular} & \multicolumn{2}{c}{\begin{tabular}[c]{@{}c@{}}GPT-4\\Turbo\end{tabular}} & \begin{tabular}[c]{@{}c@{}}GPT-4\\Turbo\end{tabular} & \begin{tabular}[c]{@{}c@{}}GPT-4\\Turbo\end{tabular}  & N/A & N/A & N/A                    
\\ \midrule
 & \textbf{Reference Model} & & \begin{tabular}[c]{@{}c@{}}Claude 3.5\\Sonnet\end{tabular} & \multicolumn{2}{c}{\begin{tabular}[c]{@{}c@{}}GPT-4\\Turbo\end{tabular}} & \begin{tabular}[c]{@{}c@{}}Claude 3.5\\Sonnet\end{tabular} & \begin{tabular}[c]{@{}c@{}}GPT-4\\Turbo\end{tabular}  & N/A & N/A & N/A
                           
\\

\cmidrule(l{7pt}r{7pt}){1-11}
& \textbf{Model / LLM System} & \begin{tabular}[c]{@{}c@{}} \textbf{Infer.}\\ \textbf{Calls}\end{tabular} & W.R. & \begin{tabular}[c]{@{}c@{}}L.C.\\ W.R.\end{tabular} & \begin{tabular}[c]{@{}c@{}}Raw\\ W.R.\end{tabular} & W.R. & W.R & Acc. & Acc. & \begin{tabular}[c]{@{}c@{}}Pass\\ @1\end{tabular} \\ \midrule

& GPT-4o - 2024-05-13        & 1 & 44.7\% & 57.5\% & 51.3\% & 48.1\% & 80.3\% & 63.6\% & 88.0\% & 84.5\% \\ 
& Claude 3.5 Sonnet          & 1 & N/A & 52.4\% & 40.6\% & N/A & 80.9\% & 68.9\% & 89.7\% & 85.0\% \\ 
& Llama 3.1 405B Instruct    & 1 & 44.7\% & 40.3\% & 37.7\% & 28.4\% & 64.1\% & 66.2\% & 88.9\% & 83.5\% \\ \midrule
& MoA                        & 19 & 51.6\% & 65.1\% & 59.8\% & 52.2\% & 84.2\% & 62.5\% & 87.3\% & 82.0\% \\ 
& MoA Lite                   & 7 & 45.6\% & 59.3\% & 57.0\% & 40.6\% & 87.8\% & 61.1\% & 87.1\% & 83.0\% \\ \midrule

\multirow{2}{*}{\rotatebox{90}{\begin{tabular}[c]{@{}c@{}} \textbf{Open} \\ \textbf{Source} \end{tabular}}} & \begin{tabular}[c]{@{}c@{}}  General-purpose \\ \archon{} Architecture \end{tabular}     & 35  & 67.5\% & 63.0\% & 68.3\% & 66.2\% & 85.1\% & 65.5\% & 86.9\% & 86.5\% \\ 
& \begin{tabular}[c]{@{}c@{}} Task-specific \\  \archon{} Architectures\end{tabular}        & 44  & 71.6\% & 66.7\% & 70.7\% & 69.0\% & 89.5\% & 67.5\% & 89.6\% & {90.5\%} \\ \midrule
\multirow{2}{*}{\rotatebox{90}{\begin{tabular}[c]{@{}c@{}} \textbf{Closed} \\ \textbf{Source} \end{tabular}}} & \begin{tabular}[c]{@{}c@{}} General-purpose \\ \archon{} Architecture \end{tabular}      & 32  & 73.1\%  & 63.5\% & 69.1\% & 70.5\% & 85.8\% & 67.7\% & 88.2\% & 88.0\%  \\ 
& \begin{tabular}[c]{@{}c@{}} Task-specific \\  \archon{} Architectures\end{tabular}        & 40  & 77.5\% & 68.4\% & 72.1\% & 74.4\% & 90.2\% & \textbf{72.9\%} & 90.4\% & 89.5\% \\ \midrule
\multirow{2}{*}{\rotatebox{90}{\begin{tabular}[c]{@{}c@{}} \textbf{All} \\ \textbf{Source} \end{tabular}}} & \begin{tabular}[c]{@{}c@{}} General-purpose \\ \archon{} Architecture \end{tabular}      & 35  & 76.8\% & 65.8\% & 70.2\% & 72.5\% & 89.3\% & 70.1\% & 88.1\% & 90.0\%  \\ 
& \begin{tabular}[c]{@{}c@{}} Task-specific \\ \archon{} Architectures\end{tabular}         & 39  & \textbf{80.4\%} & \textbf{67.6\%} & \textbf{73.3\%} & \textbf{76.1\%} & \textbf{92.1\%} & \textbf{72.9\%} & \textbf{90.6\%} & \textbf{93.5\%} \\ 

\bottomrule
\end{tabular}

\caption{
\textbf{\archon{}'s Strong Performance on the Complete Evaluation Datasets after \archon{} Architecture Optimization}:
We find that \archon{}'s inference-time architectures consistently outperform single-call state-of-the-art LLMs, both open-source and closed-source baselines, when evaluating on the complete benchmarks (\autoref{tab:benchmarks_overview}).
We explore two configurations: architecture search for building custom \archon{} configurations for each individual benchmark and architecture search for building a single general-purpose \archon{} configuration for all the benchmarks (Section \ref{sec:benchmarks_and_models}).
We find that a general \archon{} configuration lags behind the custom ones by only 3.2 percentage points, on average, across our all-source settings, which suggests the efficacy of general-purpose inference-time architectures created with our framework.  
For Arena-Hard-Auto, we also include a configuration with Claude 3.5 Sonnet as a stronger reference model for comparison against \archon{} inference-time architectures and to mitigate bias from GPT judges towards GPT generations.
For MT Bench, we use a GPT-4-0314 judge model instead of newer LLM judges to be consistent with previous results on this benchmark.
For our task-specific \archon{} architectures, we also provide the average inference calls across the given benchmarks.
For our full-list of models explored, please see \autoref{tab:models_overview}.
For MATH, we use a randomly sampled subset of size 200 for evaluation (Section \ref{sec:benchmarks_and_models}; \autoref{tab:benchmarks_overview}).
We include our \archon{} architecture results on the held-out 80\% subset of each evaluation benchmark in \autoref{tab:archon_main_table_results}.
}
\label{tab:archon_full_dataset_results}
\end{table}

\clearpage
\subsection{\archon{} LLM Analysis}

\begin{table}[H]
\centering
\scriptsize
\begin{tabular}{cccc}
\toprule
\textbf{Model}                                 & \textbf{Source Code} & \textbf{\begin{tabular}[c]{@{}c@{}}Parameter \\ Count\end{tabular}} & \textbf{\begin{tabular}[c]{@{}c@{}}Max Sequence\\ Length\end{tabular}} \\ \midrule
GPT-4o \citep{openai2024gpt4technicalreport}                                         & Closed-Source        & ---                                                                 & 128K                                                                   \\
GPT-4-Turbo \citep{openai2024gpt4technicalreport}                                    & Closed-Source        & ---                                                                 & 128K                                                                   \\
Claude-3-Opus \citep{claude3}                                & Closed-Source        & ---                                                                 & 200K                                                                   \\
Claude-3.5-Sonnet \citep{claude3}                               & Closed-Source        & ---                                                                 & 200K                                                                   \\
Claude-3-Haiku \citep{claude3}                                 & Closed-Source        & ---                                                                 & 200K                                                                   \\ \midrule
Llama-3.1-70B-Instruct \citep{dubey2024llama3herdmodels}                           & Open-Source          & 70B                                                                 & 8k                                                                     \\
Llama-3.1-405B-Instruct \citep{dubey2024llama3herdmodels}                           & Open-Source          & 70B                                                                 & 8k                                                                     \\
DeepSeek LLM 67B Chat \citep{guo2024deepseekcoderlargelanguagemodel}                            & Open-Source          & 67B                                                                 & 32k                                                                    \\
Qwen2 72B Instruct \citep{qwen2}                             & Open-Source          & 72B                                                                 & 32k                                                                    \\
Qwen1.5 110B Chat \citep{qwen}                              & Open-Source          & 110B                                                                & 32k                                                                    \\
Qwen1.5 72B Chat \citep{qwen}                               & Open-Source          & 72B                                                                 & 32k                                                                    \\
Mixtral 8x22B v0.1 \citep{jiang2024mixtral}                            & Open-Source          & 176B                                                                & 32k                                                                    \\
WizardLM 8x22B \citep{xu2024wizardlm}                                & Open-Source          & 176B                                                                & 32k                                                                    \\
dbrx-instruct \citep{dbrx}                                 & Open-Source          & 132B                                                                & 32k                                                                    \\ \midrule
princeton-nlp/Llama-3-Instruct-8B-SimPO \citep{meng2024simpo}       & Open-Source          & 8B                                                                  & 8k                                                                     \\
princeton-nlp/Llama-3-Instruct-8B-DPO \citep{meng2024simpo}         & Open-Source          & 8B                                                                  & 8k                                                                     \\
princeton-nlp/Llama-3-Instruct-8B-RDPO \citep{meng2024simpo}         & Open-Source          & 8B                                                                  & 8k                                                                     \\
princeton-nlp/Llama-3-Instruct-8B-IPO \citep{meng2024simpo}          & Open-Source          & 8B                                                                  & 8k                                                                     \\ \midrule
Llama-3.1-8B-Instruct \citep{dubey2024llama3herdmodels}            & Open-Source          & 8B                                                                  & 8k                                                                     \\
Qwen2-7B-Instruct \citep{qwen2}                              & Open-Source          & 7B                                                                  & 32k                                                                    \\
Qwen/Qwen1.5-7B-Chat \citep{qwen}                          & Open-Source          & 7B                                                                  & 32k                                                                    \\
mistralai/Mistral-7B-Instruct-v0.2 \citep{jiang2023mistral}             & Open-Source          & 7B                                                                  & 32k                                                                    \\
cognitivecomputations/dolphin-2.2.1-mistral-7b \citep{dolphin} & Open-Source          & 7B                                                                  & 32k                                                                    \\
microsoft/Phi-3-mini-4k-instruct \citep{abdin2024phi}               & Open-Source          & 4B                                                                  & 4k                                                                     \\
HuggingFaceH4/zephyr-7b-beta \citep{tunstall2023zephyr}                   & Open-Source          & 7B                                                                  & 32k                                                                    \\
microsoft/Phi-3-small-8k-instruct \citep{abdin2024phi}             & Open-Source          & 7B                                                                  & 8k                                                                     \\
snorkelai/Snorkel-Mistral-PairRM-DPO \citep{viethoangtranduong}          & Open-Source          & 7B                                                                  & 32k                                                                    \\
mistralai/Mistral-7B-Instruct-v0.3 \citep{jiang2023mistral}             & Open-Source          & 7B                                                                  & 32k                                                                    \\
\bottomrule                                                                 
\end{tabular}
\caption{\textbf{Models Tested with \archon{}}.}
\label{tab:models_overview}
\end{table}

\begin{table}
\centering
\scriptsize
\setlength{\tabcolsep}{2.0pt} %
\begin{tabular}{ccccccccccccccc}
\toprule
 & \multicolumn{2}{c}{\textbf{MT Bench}} & \multicolumn{2}{c}{\textbf{Alpaca Eval 2.0}} & \multicolumn{2}{c}{\textbf{Arena Hard Auto}} & \multicolumn{2}{c}{\textbf{MixEval}} & \multicolumn{2}{c}{\textbf{MixEval Hard}} & \multicolumn{2}{c}{\textbf{MATH}} & \multicolumn{2}{c}{\textbf{CodeContests}} \\ 
\cmidrule(l{5pt}r{5pt}){2-3} \cmidrule(l{5pt}r{5pt}){4-5} \cmidrule(l{5pt}r{5pt}){6-7} \cmidrule(l{5pt}r{5pt}){8-9} \cmidrule(l{5pt}r{5pt}){10-11} \cmidrule(l{5pt}r{5pt}){12-13} \cmidrule(l{5pt}r{5pt}){14-15}
 \textbf{Models} & \textbf{Gen} & \textbf{Fusion} & \textbf{Gen} & \textbf{Fusion} & \textbf{Gen} & \textbf{Fusion} & \textbf{Gen} & \textbf{Fusion} & \textbf{Gen} & \textbf{Fusion} & \textbf{Gen} & \textbf{Fusion} & \textbf{Gen} & \textbf{Fusion} \\ 
\midrule
GPT-4o & 44.7\% & 61.9\% & \textbf{57.5\%} & 64.5\%  & \textbf{48.1\%} & 69.2\% & 88.0\% & \textbf{89.4\%}  & 63.6\% & 65.4\% & 82.0\% & 81.0\% & 17.9\% & 19.4\% \\ 
\midrule
GPT-4-Turbo & 42.2\%  & \textbf{63.1\%} & 55.0\% & \textbf{65.8\%} & \textbf{48.1\%} & 61.9\% & 88.9\%  & 89.0\%  & 64.1\% & 64.4\% & 79.5\% & 73.5\% & 9.3\% & 14.2\% \\ 
\midrule
\begin{tabular}[c]{@{}c@{}}Claude 3 \\ Opus\end{tabular} & 30.9\%  & 57.2\% & 40.5\%  & N/A & 27.0\% & 47.9\% & 88.3\% & 88.2\%  & 63.6\% & 64.0\% & 74.5\% & 74.0\% & 10.0\% & 12.5\% \\ 
\midrule
\begin{tabular}[c]{@{}c@{}}Claude 3.5 \\ Sonnet\end{tabular} & N/A & 71.9\% & 52.37\%  & 63.6\% & N/A & \textbf{73.2\%} & \textbf{89.7\%} & 89.3\%  & \textbf{68.9\%} & \textbf{69.5\%} & \textbf{83.5\%} & \textbf{86.5\%} & 12.1\% & 15.5\% \\ 
\midrule
\begin{tabular}[c]{@{}c@{}}Qwen 2 \\ 72B Instruct \end{tabular} & 35.0\% & 59.7\% & 37.48\% & 56.0\% & 14.5\% & 49.5\% & 86.5\% & 87.5\%  &  58.7\% & 61.1\% & {81.0\%} & {78.5\%} & 3.6\% & 5.2\% \\ 
\midrule
\begin{tabular}[c]{@{}c@{}}DeepSeek LLM \\ 67B Instruct\end{tabular} & 18.4\% & 20.0\% & 17.8\% & 17.1\% & N/A & N/A & 79.2\% & N/A  & 42.5\% & N/A & 57.0\% & N/A & 5.7\% & N/A \\ 
\midrule
\begin{tabular}[c]{@{}c@{}}Qwen 1.5 \\ 72B Chat\end{tabular} & 24.7\% & 46.3\% & 36.6\% & 55.7\% & 14.4\% & 36.4\% & 84.5\% & 82.5\% & 50.3\% & 52.2\% & 71.5\% & 67.5\% & 15.0\% & 13.9\% \\ 
\midrule
\begin{tabular}[c]{@{}c@{}}Qwen 1.5 \\ 110B Chat\end{tabular} & 34.4\%  & 50.3\% & 43.6\% & 55.9\% & 21.9\% & 39.7\% & 85.3\% & 86.5\%  & 51.8\% & 55.6\% & 67.0\% & 75.5\% & 3.6\% & 7.8\% \\ 
\midrule
Wizard 8x22B & \textbf{53.8\%}  & 57.2\% & 44.7\% & 50.6\% & 45.6\% & 51.2\% & 83\% & 78.1\%  & 54.3\% & 50.4\% & 76.0\% & 60.5\% & 7.1\% & 10.4\% \\ 
\midrule
\begin{tabular}[c]{@{}c@{}}Llama 3.1 \\ 8B Instruct\end{tabular} & 33.1\% & 45.9\% & 25.6\% &  34.9\% & 11.9\% & 28.6\% & 75.0\% & 57.5\% & 41.3\% & 46.5\% & 65.5\% & 60.5\% & 8.6\% & 7.8\% \\ 
\midrule
\begin{tabular}[c]{@{}c@{}}Llama 3.1 \\ 70B Instruct\end{tabular}    & 45.0\% & 51.9\% & 35.6\% & 40.2\%  & 23.8\% & 37.2\%  & 85.7\% & 83.5\% & 61.1\% & 65.5\% & 74.0\% & 73.5\% & 20.7\% & \textbf{23.4\%} \\ 
\midrule
\begin{tabular}[c]{@{}c@{}}Llama 3.1 \\ 405B Instruct\end{tabular}   & 44.7\% & N/A  & 40.3\% & N/A & 28.4\% & N/A & 88.9\% & N/A  & 66.2\% & N/A & 78.0\% & N/A & \textbf{27.1\%} & N/A \\ 
\bottomrule
\end{tabular}
\caption{\textbf{\archon{} Generation and Fusion Performances for Single Models}: 
For Alpaca Eval 2.0, we use the length-controlled win rate (LC WR). For fusion, we gather one candidate from each of the top-10 generator models.}
\label{tab:generator_and_fusion_rankings}
\end{table}

\begin{table}[H]
\centering
\scriptsize
\setlength{\tabcolsep}{2.0pt}
\begin{tabular}{cccccccc}
\toprule
\multicolumn{1}{l}{}                                                                          & \multicolumn{5}{c}{\textbf{Jaccard Similarity (\%)}}                 \\ \cmidrule(l{5pt}r{5pt}){2-8}
\textbf{\begin{tabular}[c]{@{}c@{}}Inference-Time\\Architecture\end{tabular}}                                                                               & MT Bench & AlpacaEval 2.0 & \begin{tabular}[c]{@{}c@{}}Arena-Hard\\Auto\end{tabular} & MixEval & \begin{tabular}[c]{@{}c@{}}MixEval\\Hard\end{tabular} & MATH & \begin{tabular}[c]{@{}c@{}}Code\\Contests\end{tabular} \\ \midrule
\begin{tabular}[c]{@{}c@{}}Best Open-Source 70B+ Model,\\ Sampled 8 Times + Fuser\end{tabular} & 45.3\%   & 52.1\%         & 48.4\%          & 55.2\%  & 58.9\% & 65.2\% & 63.7\%     \\ \midrule
\begin{tabular}[c]{@{}c@{}}Ensemble (8 Top Models), \\ Sampled Once Each + Fuser\end{tabular}     & 31.6\%   & 34.1\%         & 28.9\%          & 38.6\%  & 40.9\% & 57.1\% & 53.4\% \\ \bottomrule    
\end{tabular}
\caption{\textbf{Jaccard Similarities between Candidates Responses and Fused Response by Benchmark}: For the fuser, we use the best-performing 70B+ model for each benchmark.
}
\label{tab:jaccard_similarities}
\end{table}

\begin{figure}[H]
   \centering
   \includegraphics[width=0.8\linewidth]{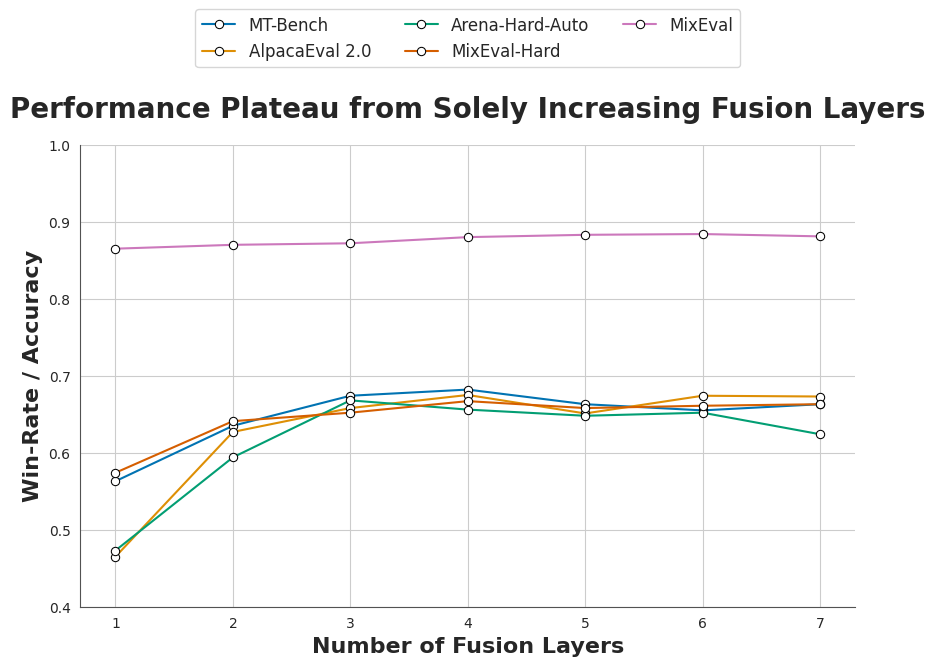}
   \caption{\textbf{Fusion Layer Efficacy by Benchmark}: 
   From solely scaling the fusion layers, we see limited benefits across the benchmarks explored but when we add other inference-time techniques, such as Critic and Ranker, we see increased downstream performance as we continue scaling inference-time compute (\autoref{fig:compound_LM_ablations}).
   We use an 8-model ensemble of the top Generator models for each benchmark (\autoref{tab:generator_and_fusion_rankings}). 
   For our Fuser layers, we use the best Fuser model for the final fuser layer (\autoref{tab:generator_and_fusion_rankings}). 
   For the intermediate layers, we use the top-8 Fuser models for each benchmark.
   }
   \label{fig:fusion_layer_analysis}
\end{figure}

\subsection{\archon{} Architectures}
\label{sec:archon_architectures}

\begin{figure}[H]
   \centering
   \includegraphics[width=\linewidth]{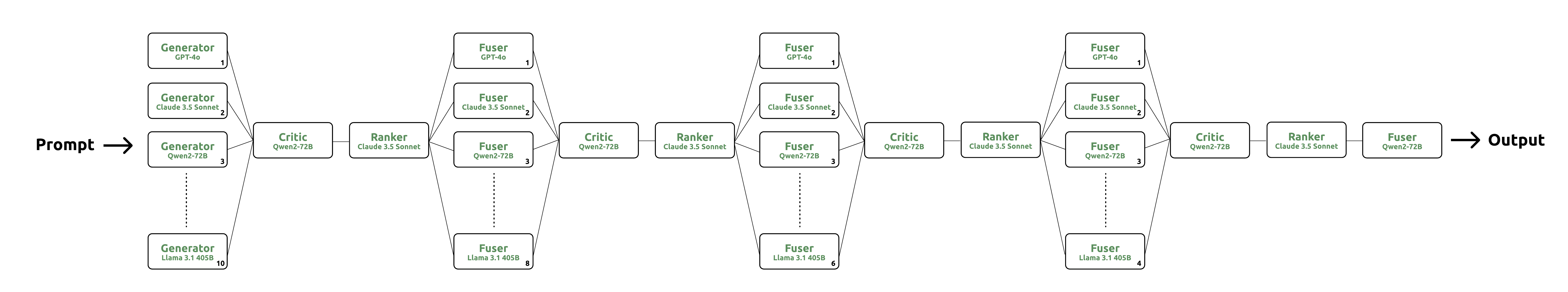}
   \caption{\textbf{All-Source Generalizable \archon{} Architecture}: Using \archon{}'s architecture search, we found this all-source \archon{} configuration to be effective across the benchmarks explored (except for CodeContests). 
   In the diagram above, we use 10 SOTA all-source LLMs to create multiple successive layers of critic, ranker, and fusers, with each successive fuser layer having less fusers to produce a "funneling" effect as the candidate generations are processed.
   The layers of critic, ranker, and fuser led to better candidate generations through iterative critique and rewriting.
   Each of the initial Generator models were sampled once.
   }
   \label{fig:funneling_architecture}
\end{figure}

\begin{figure}[H]
   \centering
   \includegraphics[width=\linewidth]{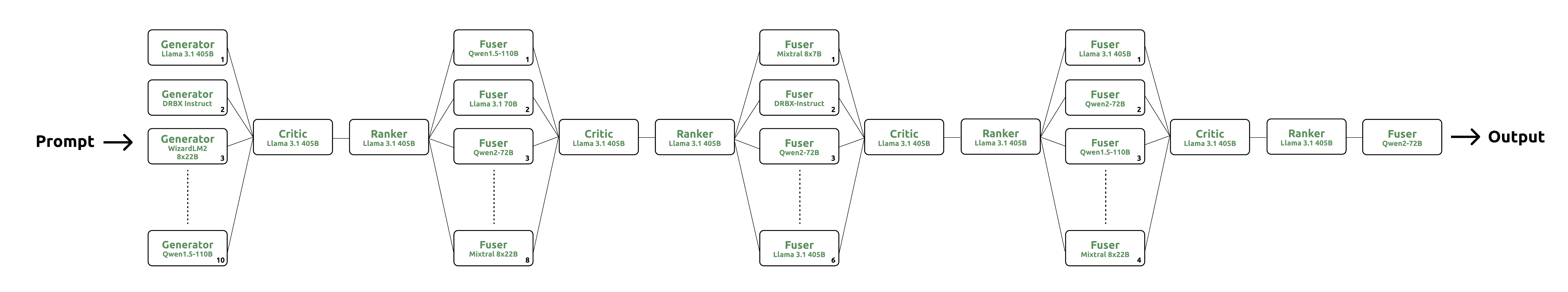}
   \caption{\textbf{All-Source \archon{} Architecture for Instruction-Following and Reasoning}: Using \archon{}'s architecture search, we found this all-source \archon{} configuration to be effective across the instruction-following benchmarks explored (MT Bench, AlpacaEval 2.0, ArenaHardAuto).
   }
   \label{fig:funneling_architecture_open-source}
\end{figure}

\begin{figure}[H]
   \centering
   \includegraphics[width=0.7\linewidth]{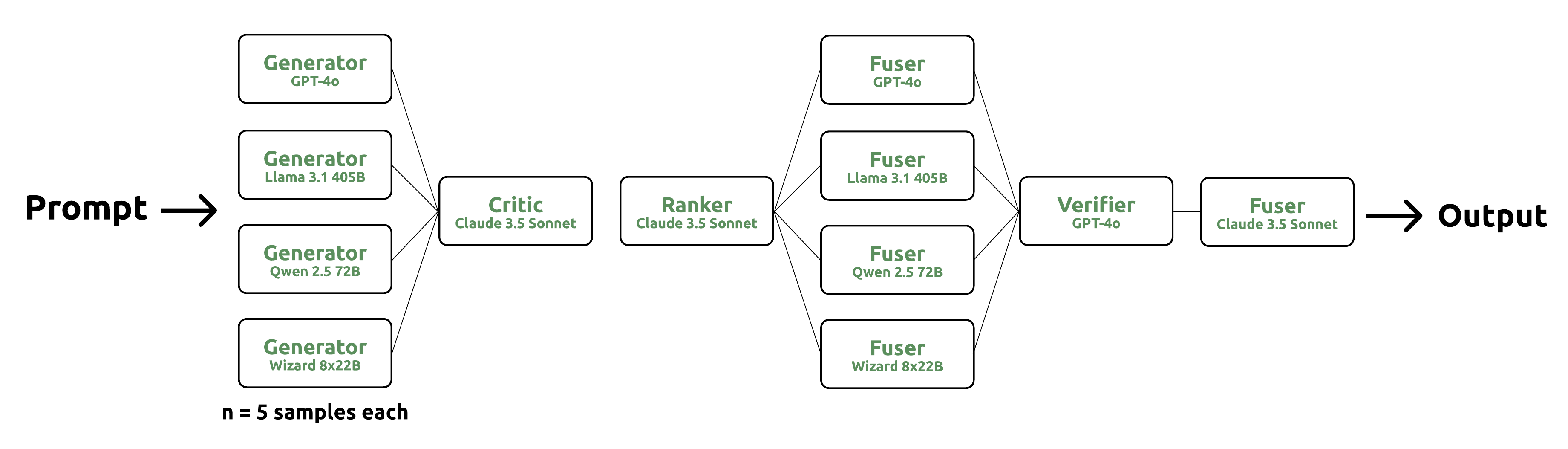}
   \caption{
   \textbf{All-Source \archon{} Architecture for Math}: Using \archon{}'s architecture search, we found this all-source \archon{} configuration to be effective across the math benchmarks explored (MATH).
   }
   \label{fig:claude_only}
\end{figure}

\begin{figure}[H]
   \centering
   \includegraphics[width=0.8\linewidth]{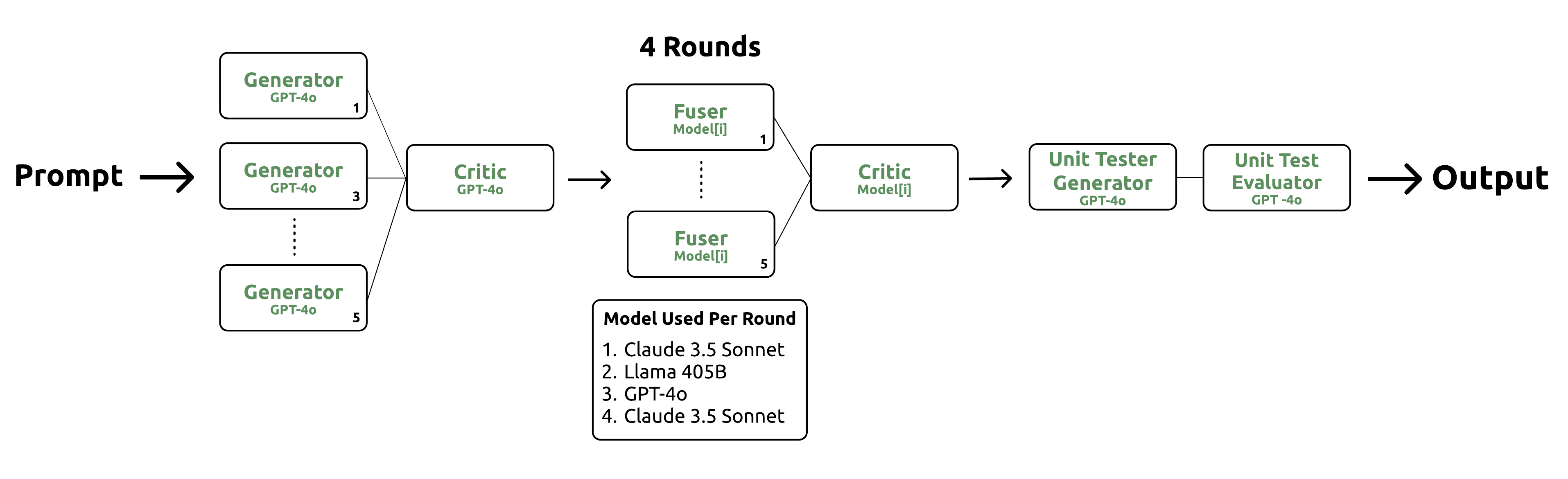}
   \caption{
   \textbf{All-Source \archon{} Architecture for Coding}: Using \archon{}'s architecture search, we found this all-source \archon{} configuration to be effective across the coding benchmarks explored (CodeContests).
   }
   \label{fig:llama_only}
\end{figure}

\subsection{\archon{} by Inference Compute Budget, Model Size, and Cost}
\begin{table}[H]
\centering
\scriptsize
\begin{tabular}{c|cccccc}
\toprule
& & \multicolumn{5}{c}{\textbf{Datasets}} \\ \cmidrule(l{7pt}r{7pt}){3-7}
 & \textbf{\begin{tabular}[c]{@{}c@{}}Number of \\ Inference Calls\end{tabular}} & \begin{tabular}[c]{@{}c@{}}MT\\ Bench\end{tabular} & \begin{tabular}[c]{@{}c@{}}Alpaca\\ Eval 2.0\end{tabular} & \begin{tabular}[c]{@{}c@{}}Arena\\ Hard Auto\end{tabular} & MixEval              & \begin{tabular}[c]{@{}c@{}}MixEval\\ Hard\end{tabular} \\ \cmidrule(l{7pt}r{7pt}){2-7} \multirow{6}{*}{\rotatebox{90}{\textbf{\begin{tabular}[c]{@{}c@{}}70B+\\ Models\end{tabular}}}}
& 1 & 55.0\% & 44.7\% & 45.6\% & 86.5\% & 61.1\% \\  %
& 10 & 52.5\% & 50.6\% & 45.6\% & 86.5\% & 63.9\% \\ 
& 20 & 65.3\% & 60.4\% & 59.4\% & 89.0\% & 65.0\% \\ 
& 30 & 69.2\% & 64.5\% & \textbf{69.0\%} & \textbf{89.5\%} & \textbf{67.5\%} \\ 
& 40 & 69.5\% & \textbf{66.7\%} & \textbf{69.0\%} & \textbf{89.5\%} & \textbf{67.5\%} \\ 
& 50 & \textbf{71.6\%} & \textbf{66.7\%} & \textbf{69.0\%} & \textbf{89.5\%} & \textbf{67.5\%} \\ \midrule
\multirow{6}{*}{\rotatebox{90}{\textbf{\begin{tabular}[c]{@{}c@{}}Closed\\ Models\end{tabular}}}} 
& 1 & 45.0\% & 57.5\% & 48.1\% & 88.9\% & 68.9\% \\  %
& 10 & 57.1\%  & 63.2\% & 68.4\% & 90.0\% & 70.1\% \\ 
& 20 & 59.4\% & 66.5\% & 75.5\% & \textbf{90.6\%} & 70.5\% \\ 
& 30 & 70.2\%  & \textbf{68.8\%} & \textbf{77.4\%}  & \textbf{90.6\%} & \textbf{72.9\%} \\ 
& 40 & 75.5\%  & \textbf{68.8\%} & \textbf{77.4\%}  & \textbf{90.6\%} & \textbf{72.9\%} \\ 
& 50 & \textbf{80.4\%} & \textbf{68.8\%} & \textbf{77.4\%} & \textbf{90.6\%} & \textbf{72.9\%} \\ \bottomrule
\end{tabular}
\caption{\textbf{\archon{} with Different Inference Budgets}:
For AlpacaEval 2.0, we use the length-controlled win rate (LC WR).}
\label{tab:archon_performance_vs_inference_budget}
\end{table}

\begin{table}[H]
\centering
\scriptsize
\begin{tabular}{cccccc}
\toprule
                                                             & \multicolumn{5}{c}{\textbf{Datasets}} \\ \cmidrule(l{7pt}r{7pt}){2-6}
                                                                                         \textbf{Models / LLM Systems}  & \begin{tabular}[c]{@{}c@{}}MT\\ Bench\end{tabular} & \begin{tabular}[c]{@{}c@{}}Alpaca\\ Eval 2.0\end{tabular} & \begin{tabular}[c]{@{}c@{}}Arena\\ Hard Auto\end{tabular} & MixEval & \begin{tabular}[c]{@{}c@{}}MixEval\\ Hard\end{tabular} \\ \midrule
Best 7B Model, 1-Sample                                                                      & 15.7\% &  41.0\%                                                  &   18.3\%                                                        &  76.2\%                                                                 &  46.1\%                                                      \\ \midrule
Best 7B Model - 10-Sample + Ranking                                                          & 16.5\%                                                   & 43.2\%                                                          & 18.9\%                                                          & 78.4\%        & 48.5\%                                                       \\ \midrule
10-Model, 1-Sample Ensemble + Ranking                                                     &  \textbf{22.4\%}                                                  &   \textbf{48.2\%}                                                       & \textbf{25.6\%}                                                          & \textbf{81.5\%}    & \textbf{52.9\%}                                                      \\ \midrule
10-Model, 1-Sample Ensemble + Fusion                                                      &  14.3\%                                                  &   39.4\%                                                        & 17.5\%                                                          & 73.2\%        &  45.2\%                                                      \\ \midrule
\begin{tabular}[c]{@{}c@{}}10-Model, 1-Sample Ensemble \\ + Top-5 Ranking + Fusion\end{tabular} &  15.9\%                                                  & 41.2\%                                                          &  18.0\%                                                         & 75.1\%        & 46.9\%                                                       \\ \midrule
\begin{tabular}[c]{@{}c@{}}10-Model, 1-Sample Ensemble \\ + Critic + Fusion\end{tabular}  &  10.5\%                                                 &  38.4\%                                                        &  16.5\%                                                         & 71.4\%        &  42.5\%                                                      \\ \bottomrule
\end{tabular}
\caption{\textbf{\archon{} with 7B Open-Source Models}: 
For AlpacaEval 2.0, we use the length-controlled win rate (LC WR).
We use open-source 7B models for testing from \autoref{tab:models_overview}.}
\label{tab:archon_with_7b}
\end{table}

\begin{table}[H]
\centering 
\scriptsize
\begin{tabular}{ccc}
\toprule
\textbf{Models}                      & \textbf{\begin{tabular}[c]{@{}c@{}}Cost (\$) per \\ Million Input Tokens\end{tabular}} & \textbf{\begin{tabular}[c]{@{}c@{}}Cost (\$) per\\ Million Output Tokens\end{tabular}} \\ \midrule
Claude 3.5 Sonnet                    & \$3                                                                               & \$15                                                                              \\ \midrule
Claude 3.0 Opus                      & \$15                                                                              & \$75                                                                              \\ \midrule
GPT-4o                               & \$5                                                                               & \$15                                                                              \\ \midrule
GPT-4-Turbo                          & \$10                                                                              & \$30                                                                              \\ \midrule
TogetherAI - Llama 3.1 405B Instruct & \$5                                                                               & \$5                                                                               \\ \midrule
TogetherAI - Llama 3.1 70B Instruct  & \$0.88                                                                            & \$0.88                                                                            \\ \midrule
TogetherAI - Other Models            & \$0.90                                                                            & \$0.90  \\ \bottomrule                                                                         
\end{tabular}
\caption{\textbf{Model API Costs as of November 2024}}
\label{tab:model_costs}
\end{table}

\begin{table}[H]
\centering
\scriptsize
\setlength{\tabcolsep}{2.5pt}
\begin{tabular}{cccccccc}
\toprule
\multicolumn{1}{l}{}                                                                              & \multicolumn{5}{c}{\textbf{Cost (\$) per Query for Benchmark}}       \\ \cmidrule(l{7pt}r{7pt}){2-8}
\textbf{\begin{tabular}[c]{@{}c@{}}Model /\\LLM System\end{tabular}}                                                                               & MT Bench & AlpacaEval 2.0 & \begin{tabular}[c]{@{}c@{}}Arena-Hard\\Auto\end{tabular} & MixEval & \begin{tabular}[c]{@{}c@{}}MixEval\\Hard\end{tabular} & MATH & \begin{tabular}[c]{@{}c@{}}Code\\Contests\end{tabular} \\ \midrule
Claude 3.5 Sonnet                                                                                 & 0.0305   & 0.0171         & 0.0212          & 0.0231  & 0.0226 & 0.0325 & 0.384       \\ \midrule
GPT-4o                                                                                            & 0.0481   & 0.0236         & 0.0324          & 0.0357  & 0.0361 & 0.514 & 0.562    \\ \midrule
Llama 3.1 405B Instruct                                                                           & 0.0281   & 0.0174         & 0.0185          & 0.0212  & 0.0205 & 0.305 & 0.372     \\ \midrule
\begin{tabular}[c]{@{}c@{}}General Purpose\\ \archon{} Architecture\end{tabular} & 0.364    & 0.189          & 0.195           & 0.284   & 0.252 & 0.375 & 0.461       \\ \midrule
\begin{tabular}[c]{@{}c@{}}Task Specific \\ \archon{} Architecture\end{tabular}  & 0.401    & 0.210          & 0.221           & 0.295   & 0.265 & 0.425 & 0.448  \\ \bottomrule  
\end{tabular}
\caption{\textbf{\archon{} Costs per Query by Benchmark}}
\label{tab:archon_costs}
\end{table}

\end{document}